\documentclass{article}
\usepackage[utf8]{inputenc}
\usepackage[english]{babel}

\usepackage{amssymb}   
\usepackage{amsthm}    
\usepackage{thmtools}  
\usepackage{mathtools} 
\usepackage{mathrsfs}  

\usepackage{graphicx}  
\usepackage{adjustbox}
\graphicspath{{figurer/}}
\usepackage{booktabs, multirow}  
\usepackage{babel}     
\usepackage{csquotes}  
\usepackage{textcomp}  
\usepackage{listings}  
\usepackage{accents}
\newcommand{\thickbar}{} 
\DeclareRobustCommand*\thickbar[1]{\accentset{\rule{.35em}{.65pt}}{#1}}

\usepackage{xcolor}
\newcommand*\dif{\mathop{}\!\mathrm{d}}
\usepackage[bottom]{footmisc}
\usepackage[width=1.14\textwidth, font=small,labelfont=bf]{caption}
\usepackage{makecell} 

\usepackage{siunitx}
\renewrobustcmd{\bfseries}{\fontseries{b}\selectfont}
\renewrobustcmd{\boldmath}{}
\newrobustcmd{\B}{\bfseries}

\usepackage{mathscinet}
\usepackage[backend    = biber,
            sortcites  = true,
            sorting    = nyt,
            giveninits = true,
            uniquename = init,
            doi        = false,
            isbn       = false,
            url        = false,
            maxcitenames = 2,
            style      = authoryear-comp]{biblatex}
\DeclareNameAlias{sortname}{family-given}
\DeclareNameAlias{default}{family-given}
\DeclareFieldFormat[article]{volume}{\bibstring{jourvol}\addnbspace#1}
\DeclareFieldFormat[article]{number}{\bibstring{number}\addnbspace#1}
\renewbibmacro*{volume+number+eid}
{
    \printfield{volume}
    \setunit{\addcomma\space}
    \printfield{number}
    \setunit{\addcomma\space}
    \printfield{eid}
}
\usepackage{appendix}

\usepackage{varioref}
\usepackage[pdfusetitle]{hyperref}
\usepackage[nameinlink, noabbrev]{cleveref}
\Crefname{chapter}{Chapter}{Chapters}
\Crefname{section}{Section}{Sections}

\newcommand{\sqbr}[1]{\left\lbrack #1 \right\rbrack}%

\newcommand{\set}[1]{\left\lbrace #1 \right\rbrace}
\newcommand{\diff}{\mathop{}\!\mathrm{d}}

\DeclarePairedDelimiterX{\infdivx}[2]{(}{)}{%
  #1\;\delimsize\|\;#2%
}

\newcommand{\R}{\mathbb{R}}   
\newcommand{\E}{\mathbb{E}}   

\renewcommand{\v}{\boldsymbol{v}}
\newcommand{\w}{\boldsymbol{w}}
\newcommand{\x}{\boldsymbol{x}}

\newcommand{\z}{\boldsymbol{z}}

\newcommand{\xsb}{{\boldsymbol{x}_{\thickbar{\mathcal{S}}}}}
\newcommand{\xs}{{\boldsymbol{x}_{\mathcal{S}}}}
\newcommand{\xss}{{\boldsymbol{x}_{\mathcal{S}}^*}}
\newcommand{\sbb}{{\thickbar{\mathcal{S}}}}
\newcommand{\s}{{\mathcal{S}}}
\newcommand{\M}{{\mathcal{M}}}
\newcommand{\pow}{{\mathcal{P}}}
\newcommand{\btheta}{{\boldsymbol{\theta}}}
\newcommand{\bpsi}{{\boldsymbol{\psi}}}
\newcommand{\bphi}{{\boldsymbol{\varphi}}}

\usepackage{booktabs, multirow}  
\usepackage{adjustbox}

\usepackage{amssymb}   
\usepackage{mathtools} 
\usepackage{mathrsfs}  
\usepackage{enumitem}  

\usepackage{varioref}
\usepackage[nameinlink, noabbrev]{cleveref}
\Crefname{chapter}{Chapter}{Chapters}
\Crefname{section}{Section}{Sections}

\usepackage[margin=1.40in]{geometry}

\usepackage{authblk}

\title{Using Shapley Values and Variational Autoencoders to \\ Explain Predictive Models with Dependent Mixed Features}

\author[1]{Lars Henry Berge Olsen\thanks{E-mail addresses: \href{mailto:lholsen@math.uio.no}{\texttt{lholsen@math.uio.no}}, \href{mailto:glad@math.uio.no}{\texttt{glad@math.uio.no}}, \href{mailto:jullum@nr.no}{\texttt{jullum@nr.no}}, and \href{mailto:kjersti@nr.no}{\texttt{kjersti@nr.no}}. }}
\author[1]{Ingrid Kristine Glad}
\author[2]{\\Martin Jullum}
\author[2]{Kjersti Aas}
\affil[1]{Department of Mathematics, University of Oslo}
\affil[2]{Norwegian Computing Center}
\date{\today}

\begin{document}

\maketitle

\begin{abstract}
    Shapley values are today extensively used as a model-agnostic explanation framework to explain complex predictive machine learning models. Shapley values have desirable theoretical properties and a sound mathematical foundation in the field of cooperative game theory. Precise Shapley value estimates for dependent data rely on accurate modeling of the dependencies between all feature combinations. In this paper, we use a variational autoencoder with arbitrary conditioning (\texttt{VAEAC}) to model all feature dependencies simultaneously. We demonstrate through comprehensive simulation studies that our \texttt{VAEAC} approach to Shapley value estimation outperforms the state-of-the-art methods for a wide range of settings for both continuous and mixed dependent features. For high-dimensional settings, our \texttt{VAEAC} approach with a non-uniform masking scheme significantly outperforms competing methods. Finally, we apply our \texttt{VAEAC} approach to estimate Shapley value explanations for the Abalone data set from the UCI Machine Learning Repository.
\end{abstract}

\section{Introduction}
\label{sec:Introduction}

Explainable artificial intelligence (XAI) and interpretable machine learning (IML) have become active research fields in recent years \parencite{adadi2018peeking, molnar2019, covert2021explaining}. This is a natural consequence as complex machine learning (ML) models are now applied to solve supervised learning problems in many high-risk areas: cancer prognosis \parencite{KOUROU20158}, credit scoring \parencite{KVAMME2018207}, and money laundering detection \parencite{jullum2020detecting}. The high prediction accuracy of complex ML models may come at the expense of model interpretability, as discussed by \textcite{johansson2011trade, Guo2019AnIM, luo2019balancing}, while \textcite{rudin2019stop} conjectures that equally accurate but interpretable models exist across domains even though they might be hard to find. As the goal of science is to gain knowledge from the collected data, the use of black-box models hinders the understanding of the underlying relationship between the features and the response, and thereby curtails scientific discoveries. 

Model explanation frameworks from the XAI field extract the hidden knowledge about the underlying data structure captured by a black-box model, making the model's decision-making process transparent. This is crucial for, e.g., medical researchers who apply an ML model to obtain well-performing predictions but who simultaneously also strive to discover important risk factors. Another driving factor is the \textit{Right to Explanation} legislation in the European Union's General Data Protection Regulation (GDPR) which gives a data subject the right to obtain an explanation of decisions reached by algorithms that significantly affect the individual \parencite{regulation2016}.

A promising explanation methodology, with a strong mathematical foundation and unique theoretical properties within the field of cooperative game theory, is \textit{Shapley values} \parencite{shapley1953value}. Within XAI, Shapley values are most commonly used as a \textit{model-agnostic} explanation framework for individual predictions, i.e., \textit{local explanations}. The methodology has also been used to provide \textit{global explanations}, see \textcite{owen2014sobol, covert2020understanding, frye_shapley-based_2020,  giudici2021shapley}. Model-agnostic means that Shapley values do not rely on model internals and can be used to compare and explain different ML models trained on the same supervised learning problem. Examples of ML models are neural networks \parencite{gurney2018introduction_neuralnetworks}, random forest \parencite{breiman2001random}, and XGBoost \parencite{chen2015xgboost}. Local explanation means that Shapley values do not explain the global model behavior across all data instances but rather locally for an individual observation. See \textcite{molnar2019} for an overview and detailed introduction to other explanation frameworks.

Shapley values originated in cooperative game theory but have been re-introduced as a framework for model explanation by \textcite{strumbelj2010efficient, strumbelj2014explaining, lundberg2017unified, lundberg2017consistent}. Originally, Shapley values described a possible solution concept of how to fairly distribute the payout of a game to the players based on their contribution to the overall cooperation/payout. The solution concept is based on several desirable axioms, for which Shapley values are the unique solution. When applying Shapley values as an explanation framework, we treat the features as the ``players", the machine learning model as the ``game", and the corresponding prediction as the ``payout". In this framework, the difference between the prediction of a specified individual and the global average prediction is fairly distributed among the associated features.

\textcite{lundberg2017unified} were a major driving force in the popularization of Shapley values for model explanation with their Python library \texttt{SHAP}. In their approach, they implicitly assumed independence between the features, as this partially simplifies the Shapley value estimation. However, the estimation is still fundamentally computationally expensive as the number of terms involved in the Shapley value formula, elaborated in \Cref{sec:ShapleyValues}, grows exponentially with the number of features. Furthermore, in observational studies, the features are rarely statistically independent and the \texttt{independence} approach can therefore compute incorrect explanations, see, for example, \textcite{merrick_explanation_2020, frye_shapley-based_2020, aas2019explaining}. 

\textcite{aas2019explaining} extend the ideas of \textcite{lundberg2017unified} and propose three methods for modeling the dependence between the features: the \texttt{Gaussian}, \texttt{copula}, and \texttt{empirical} approaches. The same paper demonstrates through simulation studies that these conditional approaches obtain more precise Shapley value estimates, in terms of the mean absolute error, than the \texttt{independence} approach for moderate to strong dependence between some or all of the features. \textcite{aas2021explaining} further improve the explanation precision by using non-parametric vine copulas \parencite{CopulaVine}. \textcite{redelmeier:2020} extend the methodology to mixed data, that is, continuous and categorical features, by modeling the feature dependencies using conditional inference trees \parencite{hothorn2006unbiased}. However, these improvements further increase the computational complexity of Shapley values, as we in addition need to estimate the feature dependencies between all feature combinations. \textcite{jullum2021groupshapley} propose to explain groups of related features instead of individual features to decrease the fundamental computational problem for Shapley values.

The above papers use methods from the statistical community to model the feature dependencies, but machine learning methods can also be applied. Two promising ML methodologies for constructing generative models for conditional distributions are \textit{generative adversarial network} (\texttt{GAN}) \parencite{goodfellow2014generative} and \textit{variational autoencoder} (\texttt{VAE}) \parencite{kingma2014autoencoding, pmlr-v32-rezende14}. The \texttt{VAE} can be seen as an extension of the EM algorithm \parencite{Kingma2019AnIT, DING2021} and as the natural evolution of robust principal component analysis (PCA) models \parencite{VAE_and_PCA}. Both the \texttt{VAE} and \texttt{GAN} frameworks consist of neural networks (NNs) but with different model setups. They have been used in a wide variety of generative modeling tasks; human image synthesis \parencite{styleGAN2}, simulating gravitational lensing for dark matter research \parencite{cosmoGAN}, blending and exploring musical scores \parencite{musicVAE, musicVAE2}, and generating coherent novel sentences \parencite{bowman2016generating}. 

Both frameworks have methods that could potentially be used to estimate Shapley values, namely \textit{generative adversarial imputation nets} (\texttt{GAIN}) \parencite{yoon_gain_2018} and \textit{conditional variational autoencoders} (\texttt{CVAE}) \parencite{CVAE_Sohn}, respectively. In this paper, we use the \textit{variational autoencoder with arbitrary conditioning} (\texttt{VAEAC}) by \textcite{ivanov_variational_2018}, which is a generalization of \texttt{CVAE}. \texttt{VAEAC} estimates the dependencies between all features simultaneously with a \emph{single} variational autoencoder and it supports mixed data. 

That \texttt{VAEAC} uses a single model is a great advantage compared to the methods of \textcite{redelmeier:2020, aas2019explaining, aas2021explaining}, which fit a new model for each feature combination. As the number of feature combinations grows exponentially with the number of features, the \texttt{VAEAC} methodology can reduce the computational burden of Shapley values as a model explanation framework. An abstract visualisation of how the \texttt{VAEAC} approach estimates feature dependencies is presented in \Cref{fig:VAEAC_abstract_latent}. The figure illustrates how \texttt{VAEAC} uses two probabilistic \textit{encoders} to learn latent representations of the data and a single probabilistic \textit{decoder} to map the latent representation back to the feature space while taking the feature dependencies into account.

In this article, we focus on how we can use the \texttt{VAEAC} methodology to compute \textit{local} Shapley values in a regression setting. A related approach has been used by \textcite{frye_shapley-based_2020} for obtaining \textit{global} explanations by aggregating local Shapley values in a classification setting. The degree of similarity with our \texttt{VAEAC} approach to Shapley value estimation is not straightforward to quantify, as they provide little details on their method. There are at least three main differences as far as we are concerned. First, we use a novel non-uniform \textit{masking scheme} which greatly increases the performance of the \texttt{VAEAC} methodology in high-dimensional simulation studies. Second, our approach handles \textit{missing feature values}, and third, it is resilient to \textit{vanishing gradients} as we use skip-layer connections and LeakyReLU. Additionally, we conduct thorough simulation studies where our \texttt{VAEAC} approach to Shapley value estimation outperforms the current state-of-the-art methods \parencite{lundberg2017unified, redelmeier:2020, aas2019explaining} implemented in the \textsc{R}-package \texttt{shapr} \parencite{shapr}.

The rest of the paper is organized as follows. In \Cref{sec:ShapleyValues}, we introduce the concept of Shapley values in the game-theoretical setting and as a local model-agnostic explanation framework with individual explanations. The variational autoencoder with arbitrary conditioning (\texttt{VAEAC}) is introduced in \Cref{sec:VAEAC}, where we also show how the methodology is used to estimate precise Shapley values for dependent features. In \Cref{sec:SimulationStudy}, we illustrate the excellent performance of our \texttt{VAEAC} approach compared to the state-of-the-art methods through exhaustive simulations studies. Then, in \Cref{sec:RealDataExample}, we use our \texttt{VAEAC} approach to generate Shapley value explanations for the Abalone data set \parencite{nash1994population} from the UCI Machine Learning Repository. \Cref{sec:Conclusion} ends the paper with some concluding remarks and potential further work. In the \nameref{Appendix}, we provide full mathematical derivations, implementations details, and additional simulation studies.

\section{Shapley Values}
\label{sec:ShapleyValues}

In this section, we first describe Shapley values in cooperative game theory before we discuss how they can be used for model explanation.

\subsection{Shapley Values in Cooperative Game Theory}
\label{subsec:ShapleyValuesGameTheory}

Shapley values are a unique solution concept of how to divide the payout of a cooperative game $v:\mathcal{P}(\M) \mapsto \R$ based on four axioms \parencite{shapley1953value}. The game is played by $M$ players where $\M = \{1,2,\dots,M\}$ denotes the set of all players and $\mathcal{P}(\M)$ is the power set, that is, the set of all subsets of $\M$. We call $v(\s)$ for the \textit{contribution function} and it maps a subset of players $\s \in \pow(\M)$, also called a coalition, to a real number representing their contribution in the game $v$. The Shapley values $\phi_j = \phi_j(v)$ assigned to each player $j$, for $j = 1, \dots, M$, satisfy the following properties:
\begin{enumerate}[align=left, itemsep=2pt, topsep=4pt] 
    \item [\textbf{Efficiency}:] They sum to the value of the grand coalition $\M$ over the empty set $\emptyset$, that is, $\sum_{j=1}^M \phi_j = v(\M) - v(\emptyset)$.
    \item [\textbf{Symmetry}:] Two equally contributing players $j$ and $k$, that is,  $v(\s \cup \{j\}) = v(\s \cup \{k\})$ for all $\s$, receive equal payouts $\phi_j = \phi_k$.
    \item [\textbf{Dummy}:] A non-contributing player $j$, that is, $v(\s) = v(\s \cup \{j\})$ for all $\s$, receives $\phi_j = 0$.
    \item [\textbf{Linearity}:] A linear combination of $n$ games $\{v_1, \dots, v_n\}$, that is, $v(\s) = \sum_{k=1}^nc_kv_k(\s)$, has Shapley values given by $\phi_j(v) = \sum_{k=1}^nc_k\phi_j(v_k)$.
\end{enumerate}
    
\textcite{shapley1953value} showed that the values $\phi_j$ which satisfy these axioms are given by
\begin{align}
    \label{eq:ShapleyValuesDef}
    \phi_j = \sum_{\mathcal{S} \in \pow(\mathcal{M}) \backslash \{j\}} \frac{|\mathcal{S}|!(M-|\mathcal{S}|-1)!}{M!}\left(v(\mathcal{S} \cup \{j\}) - v(S) \right),
\end{align}
where $|\mathcal{S}|$ is the number of players in coalition $\s$. The number of terms in \eqref{eq:ShapleyValuesDef} is $2^{M}$, hence, the complexity grows exponentially with the number of players $M$. Each Shapley value is a weighted average of the player’s marginal contribution to each coalition $\mathcal{S}$. The proposed payouts, that is, the Shapley values, are said to be \textit{fair} as they satisfy the four axioms, which are discussed in more detail in, for example, \textcite{shapley1953value, roth1988shapley, hart1989shapley}.

\subsection{Shapley Values in Model Explanation}
\label{subsec:ShapleyValuesExplainability}

Assume that we are in a supervised learning setting where we want to explain a predictive model $f(\boldsymbol{x})$ trained on $\{\boldsymbol{x}^{[i]}, y^{[i]}\}_{i = 1, \dots, N_\text{train}}$, where $\boldsymbol{x}^{[i]}$ is an $M$-dimensional feature vector, $y^{[i]}$ is a univariate response, and $N_\text{train}$ is the number of training observations.

Shapley values as a model-agnostic explanation framework \parencite{strumbelj2010efficient, strumbelj2014explaining, lundberg2017unified} enable us to fairly explain the prediction $\hat{y} = f(\boldsymbol{x})$ for a specific feature vector $\boldsymbol{x} = \boldsymbol{x}^*$. The fairness aspect of Shapley values in the model explanation setting is discussed in, for example, \textcite{chen2020true, Fryer2021ShapleyVF, aas2019explaining}. 

In the Shapley value framework, the predictive model $f$ replaces the cooperative game and the $M$-dimensional feature vector replaces the $M$ players. The Shapley value $\phi_j$ describes the importance of the $j$th feature in the prediction $f(\boldsymbol{x}^*) = \phi_0 + \sum_{j=1}^M\phi_j$, where $\phi_0 = \E \left[f(\boldsymbol{x})\right]$. That is, the sum of the Shapley values explains the difference between the prediction $f(\boldsymbol{x}^*)$ and the global average prediction. 

To calculate \eqref{eq:ShapleyValuesDef}, we need to define an appropriate contribution function $v(\mathcal{S}) = v(\mathcal{S}, \x)$ which should resemble the value of $f(\boldsymbol{x})$ when only the features in coalition $\mathcal{S}$ are known. We use the contribution function proposed by \textcite{lundberg2017unified}, namely the expected outcome of $f(\boldsymbol{x})$ conditioned on the features in $\mathcal{S}$ taking on the values $\boldsymbol{x}_\mathcal{S}^*$. That is, 
\begin{align}
    \label{eq:ContributionFunc}
    \begin{split}
        v(\mathcal{S}) 
        &=
        \E\left[ f(\boldsymbol{x}) | \boldsymbol{x}_{\mathcal{S}} = \boldsymbol{x}_{\mathcal{S}}^* \right] 
        =
        \E\left[ f(\boldsymbol{x}_{\thickbar{\mathcal{S}}}, \boldsymbol{x}_{\mathcal{S}}) | \boldsymbol{x}_{\mathcal{S}} = \boldsymbol{x}_{\mathcal{S}}^* \right] 
        = 
        \int f(\boldsymbol{x}_{\thickbar{\mathcal{S}}}, \boldsymbol{x}_{\mathcal{S}}^*) p(\boldsymbol{x}_{\thickbar{\mathcal{S}}} | \boldsymbol{x}_{\mathcal{S}} = \boldsymbol{x}_{\mathcal{S}}^*) \diff \boldsymbol{x}_{\thickbar{\mathcal{S}}},
    \end{split}
\end{align}
where $\boldsymbol{x}_{\mathcal{S}} = \{x_j:j \in \mathcal{S}\}$ denotes the features in subset $\mathcal{S}$, $\boldsymbol{x}_{\thickbar{\mathcal{S}}} = \{x_j:j \in \thickbar{\mathcal{S}}\}$ denotes the features outside $\mathcal{S}$, that is, $\thickbar{\mathcal{S}} = \mathcal{M}\backslash\mathcal{S}$, and $p(\boldsymbol{x}_{\thickbar{\mathcal{S}}} | \boldsymbol{x}_{\mathcal{S}} = \boldsymbol{x}_{\mathcal{S}}^*)$ is the conditional density of $\xsb$ given $\xs = \xss$. The conditional expectation summarises the whole probability distribution, it is the most common estimator in prediction applications, and it is also the minimizer of the commonly used squared error loss function. Note that the last equality of \eqref{eq:ContributionFunc} only holds for continuous features. That is, the integral should be replaced by sums for discrete or categorical features in $\boldsymbol{x}_{\thickbar{\mathcal{S}}}$, if there are any, and $p(\boldsymbol{x}_{\thickbar{\mathcal{S}}} | \boldsymbol{x}_{\mathcal{S}} = \boldsymbol{x}_{\mathcal{S}}^*)$ is then no longer continuous.

The contribution function in \eqref{eq:ContributionFunc} is also used by, for example, \textcite{chen2020true, covert2020understanding, redelmeier:2020, frye_shapley-based_2020, aas2019explaining, aas2021explaining}. \textcite{covert2021explaining} argue that the conditional approach in \eqref{eq:ContributionFunc} is the only approach that is consistent with standard probability axioms. In practice, the contribution function $v(\mathcal{S})$ needs to be empirically approximated by, e.g., Monte Carlo integration for all $\s \in \pow(\mathcal{M})$. That is,
\begin{align}
    \label{eq:KerSHAPConditionalFunction}
    v(\mathcal{S})  
    =
    \E\left[ f(\boldsymbol{x}_{\thickbar{\mathcal{S}}}, \boldsymbol{x}_{\mathcal{S}}) | \boldsymbol{x}_{\mathcal{S}} = \boldsymbol{x}_{\mathcal{S}}^* \right] 
    \approx
    \hat{v}(\mathcal{S}) 
    =
    \frac{1}{K} \sum_{k=1}^K f(\boldsymbol{x}_{\thickbar{\mathcal{S}}}^{(k)}, \boldsymbol{x}_{\mathcal{S}}^*),
\end{align}
where $f$ is the machine learning model, $\boldsymbol{x}_{\thickbar{\mathcal{S}}}^{(k)} \sim p(\boldsymbol{x}_{\thickbar{\mathcal{S}}} | \boldsymbol{x}_{\mathcal{S}} = \boldsymbol{x}_{\mathcal{S}}^*)$, for $k=1,2,\dots,K$, and $K$ is the number of Monte Carlo samples.

\textcite{lundberg2017unified} assume feature independence, i.e., $p(\boldsymbol{x}_{\thickbar{\mathcal{S}}} | \boldsymbol{x}_{\mathcal{S}} = \boldsymbol{x}_{\mathcal{S}}^*) = p(\boldsymbol{x}_{\thickbar{\mathcal{S}}})$, which means that $\boldsymbol{x}_{\thickbar{\mathcal{S}}}^{(k)}$ can be randomly sampled from the training set. However, in observational studies, the features are rarely statistically independent. Thus, the \texttt{independence} approach may lead to incorrect explanations for real data, as discussed in, for example, \textcite{merrick_explanation_2020, aas2019explaining, frye_shapley-based_2020}. \textcite{redelmeier:2020, aas2019explaining, aas2021explaining} use different methods from the statistical community to model the feature dependencies and generate the conditional samples $\boldsymbol{x}_{\thickbar{\mathcal{S}}}^{(k)} \sim p(\boldsymbol{x}_{\thickbar{\mathcal{S}}} | \boldsymbol{x}_{\mathcal{S}} = \boldsymbol{x}_{\mathcal{S}}^*)$.

The aforementioned approaches need to estimate the conditional distributions for all feature combinations, i.e., $p(\boldsymbol{x}_{\thickbar{\mathcal{S}}} | \boldsymbol{x}_{\mathcal{S}} = \boldsymbol{x}_{\mathcal{S}}^*)$ for all $\mathcal{S} \in \pow(\mathcal{M})$. The number of unique feature combinations is $2^M$, hence, the number of conditional distributions to model increases exponentially with the number of features. Thus, the methods above have to train $2^M$ different models, which eventually becomes computationally intractable. The method we propose in \Cref{sec:VAEAC} comes from the machine learning community and can model all the $2^M$ different conditional distributions simultaneously with a single variational autoencoder.

\section{Variational Autoencoder with Arbitrary Conditioning}
\label{sec:VAEAC}

The methodology of the \textit{variational autoencoder with arbitrary conditioning} (\texttt{VAEAC}) \parencite{ivanov_variational_2018} is an extension of the \textit{conditional variational autoencoder} (\texttt{CVAE}) \parencite{CVAE_Sohn}, which is in turn a special type of a \textit{variational autoencoder} (\texttt{VAE}) \parencite{kingma2014autoencoding, pmlr-v32-rezende14}. If the concept of \texttt{VAE}s is unfamiliar to the reader, we recommend reading \textcite[Ch.\ 1 \& 2]{Kingma2019AnIT}. These chapters give a motivational introduction to \texttt{VAE}s and lay a solid technical foundation which is necessary to understand the methodology of \texttt{VAEAC}. 

The goal of the \texttt{VAE} is to give a probabilistic representation of the true unknown distribution $p(\boldsymbol{x})$. Briefly stated, the \texttt{VAE} assumes a \textit{latent variable model} $p_{\boldsymbol{\xi}}(\x, \z) = p_{\boldsymbol{\xi}}(\z)p_{\boldsymbol{\xi}}(\x|\z)$, where $\boldsymbol{\xi}$ are the model parameters and $\z$ are the latent variables which conceptually represent an encoding of $\x$. Marginalizing over $\z$ gives $p_{\boldsymbol{\xi}}(\boldsymbol{x}) = \int p_{\boldsymbol{\xi}}(\x, \z) \dif{\z}$, which is an estimate of $p(\boldsymbol{x})$. The \texttt{CVAE} does the same for the conditional distribution $p(\xsb | \xs = \xss)$, for a specific subset $\mathcal{S}\in \pow(\mathcal{M})$ of features. Here $\xsb$ is the complement to $\xs$, that is, the features of $\x$ not in $\xs$. The \texttt{VAEAC} generalizes this methodology to all conditional distributions $p(\xsb | \xs = \xss)$, for all possible feature subsets $\mathcal{S} \in \pow(\mathcal{M})$ simultaneously.

\begin{figure}[!t]
    \centering
    \includegraphics[width = 0.975\textwidth]{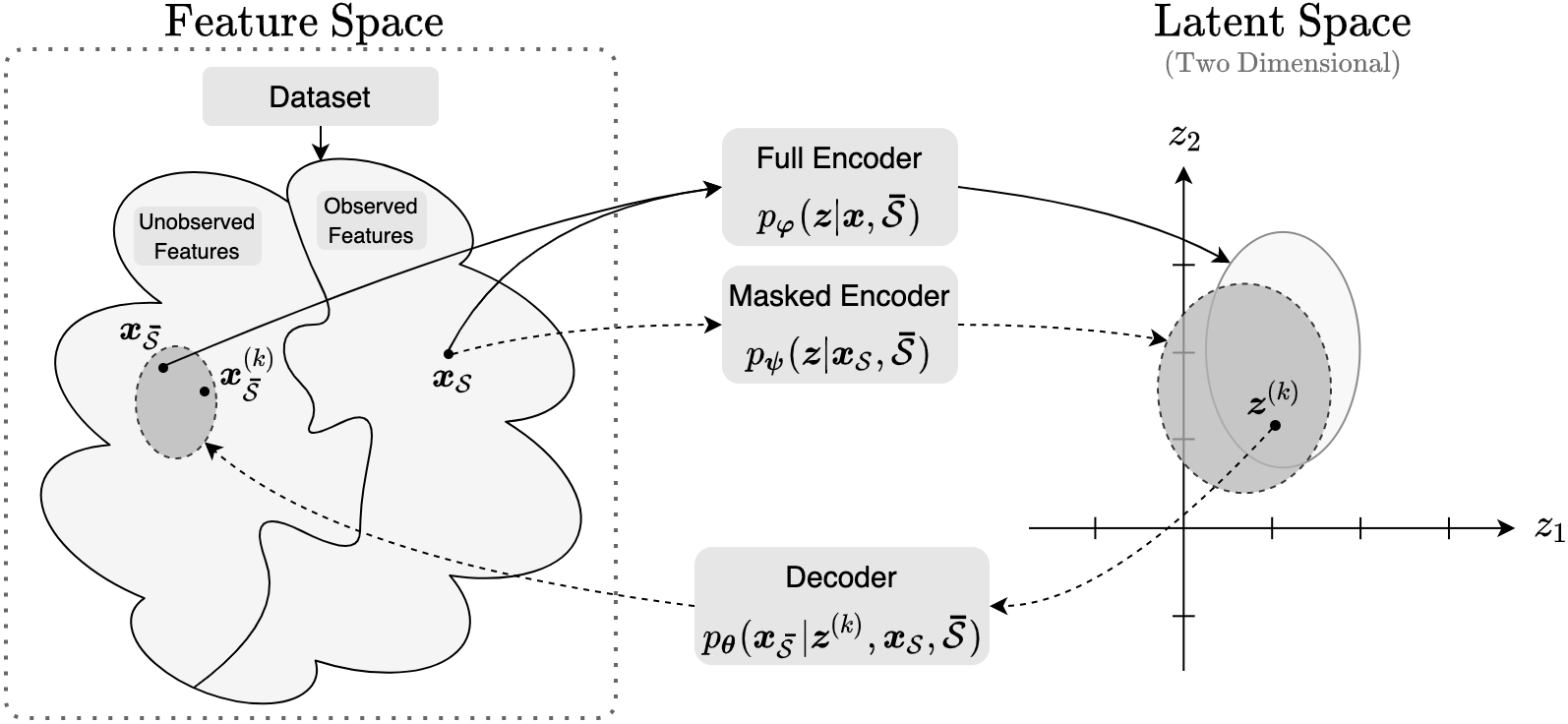}
    \caption{\small{An abstract representation of how \texttt{VAEAC} generates the conditional samples $\x_{\sbb}^{(k)}$ conditioned on the observed features $\xs$, for $k=1,2,\dots,K$, where $K$ denotes the number of Monte Carlo samples in \eqref{eq:KerSHAPConditionalFunction}. The feature space can, e.g., be the positive real numbers, as in \Cref{subsection:simstudy:cont}. Conceptually, the full encoder guides the masked encoder in the \textit{training phase} of the \texttt{VAEAC} method, while only the masked encoder and decoder are used in the \textit{employment phase} when generating the conditional samples.}}
    \label{fig:VAEAC_abstract_latent}
\end{figure}

In this paper, we present the \texttt{VAEAC} methodology for the Shapley value setting in the conventional Shapley value notation used in \Cref{sec:ShapleyValues}. The \texttt{VAEAC} consists of three fully connected feedforward neural networks (NNs): the \textit{full encoder}\footnote{In \textit{variational inference} notation, the variational distributions are often denoted by $q$ instead of $p$.} $p_\bphi$, \textit{masked encoder} $p_\bpsi$, and \textit{decoder} $p_\btheta$.\footnote{\textcite{ivanov_variational_2018} call them the \textit{proposal network}, \textit{prior network}, and \textit{generative network}, respectively.} The roles of the different entities are abstractly presented in \Cref{fig:VAEAC_abstract_latent}, which is an extension of Figure 2.1 in \textcite{Kingma2019AnIT}, and they will be explained in greater detail in the following sections.

In \Cref{subsec:VAEAC:Overview}, we give an overview of the \texttt{VAEAC} methodology. \Cref{subsub:ModelDescriptionVAEAC,subsub:Training} explain the \textit{employment phase} and \textit{training phase} of the \texttt{VAEAC} method, respectively, where the former is when we generate the conditional samples $\x_{\sbb}^{(k)}$. Our novel non-uniform \textit{masking scheme} is introduced in \Cref{subseq:MaskingSchemes}. Finally, \texttt{VAEAC}'s handling of \textit{missing data} and use of \textit{skip connections} is discussed in \Cref{subseq:MissingFeatures,subsec:SkipConnections}, respectively.



\subsection{Overview of the \texttt{VAEAC} Method}
\label{subsec:VAEAC:Overview}

Let $p(\x)$ denote the true, but unknown distribution of the data, where the input $\x$ is an $M$-dimensional vector of continuous and/or categorical features. The true conditional distribution $p(\xsb|\xs)$ is in general unknown, except for some distributions $p(\x)$, such as the multivariate Gaussian and Burr distributions. Here $\xs$ and $\xsb$ denote the \textit{observed features} (that we condition on) and the \textit{unobserved features} of the \textit{full} sample $\x$, respectively. \textcite{ivanov_variational_2018} say that the unobserved features $\xsb$ are \textit{masked} by the \textit{mask} $\thickbar{\mathcal{S}}$. We will use mask $\sbb$ and coalition $\s$ interchangeably as they are uniquely related by $\sbb  = \M \backslash \s$. 

The \texttt{VAEAC} model gives a probabilistic representation $p_{\bpsi, \btheta}(\xsb | \xs, \sbb)$ of the true unknown conditional distribution $p(\xsb|\xs)$, for a given $\xs = \xss$ and an arbitrary coalition $\mathcal{S}\in \pow(\mathcal{M})$. Here, $\bpsi$ and $\btheta$ denote the learned model parameters of the masked encoder $p_\bpsi$ and decoder $p_\btheta$, respectively. The masked encoder is an extension of the \texttt{VAE} methodology, which solely consists of a decoder and a full encoder $p_\bphi$, where the latter is parameterized by $\bphi$. Both encoders are stochastic mappings from parts of the feature space to a latent space. The full encoder $p_\bphi$ stochastically maps the \textit{full} observation $\x = \{\xsb, \xs\}$ to a latent representation $\z$, hence, the name full encoder. The masked encoder $p_\bpsi$ does the same but only conditioned on the observed features $\xs$. The unobserved features $\xsb$ of $\x$ have been \textit{masked}, hence, the name masked encoder. The decoder stochastically maps latent representations $\z^{(k)}$ back to the unobserved part of the feature space, from which we can generate on-distribution samples $\x_{\sbb}^{(k)}$. These samples are then used to estimate the contribution function $v(\s)$ in \eqref{eq:KerSHAPConditionalFunction}.

The full encoder is only used during the \textit{training phase} of the \texttt{VAEAC} method when we have access to the full training observation $\x$. Conceptually, the purpose of the full encoder is to guide the masked encoder in finding proper latent representations. In the training phase, we artificially mask out coalitions of features from $\x$ and train \texttt{VAEAC} to generate likely values of the artificially masked/unobserved features $\xsb$ conditioned on the observed features $\xs$. In the \textit{employment phase}, the \texttt{VAEAC} method only uses the masked encoder and decoder to generate the conditional samples $\x_{\sbb}^{(k)} \sim p_{\bpsi, \btheta}(\xsb | \xs, \sbb)$ used in \eqref{eq:KerSHAPConditionalFunction}. This can be seen by following the arrows in \Cref{fig:VAEAC}, where we give a detailed visual representation of a \texttt{VAEAC} model and its associated NNs, which should make the explanations of the \texttt{VAEAC} approach easier to follow. We will go through an example based on \Cref{fig:VAEAC} once all notation and terminology is introduced.

Since \texttt{VAEAC} estimates all conditional distributions simultaneously, the trained conditional model $p_{\bpsi, \btheta}(\xsb | \xs, \sbb)$ may be more precise for some $\sbb$ and less precise for others. The precision can be controlled by introducing a \textit{masking scheme}, which is a distribution $p(\sbb)$ over different masks/feature subsets $\sbb$. The mask distribution $p(\sbb)$ can be arbitrary, or defined based on the problem at the hand. However, full support over $\pow(\mathcal{M})$ is crucial to ensure that $p_{\bpsi, \btheta}(\xsb | \xs, \sbb)$ can evaluate arbitrary conditioning. Masking schemes and possible applications are further discussed in \Cref{subseq:MaskingSchemes,sec:Conclusion}.

The log-likelihood objective function of the \texttt{VAEAC} model is
\begin{align}
    \label{eq:loglikVAEAC}
    \E_{p(\x)}\E_{p(\sbb)} \big[ \log p_{\bpsi, \btheta}(\xsb | \xs, \sbb) \big],
\end{align}
which is to be maximized by tuning $\bpsi$ and $\btheta$. When all features are unobserved, i.e., $\sbb = \M$, \texttt{VAEAC} estimates the logarithm of $p(\xsb | \xs, \sbb) = p(\x_\M | \x_\emptyset, \M) = p(\x)$, that is, the full joint distribution, which corresponds to goal of the original \texttt{VAE} \parencite{kingma2014autoencoding, pmlr-v32-rezende14}. Furthermore, if $p(\sbb) = 1$ for some fixed $\sbb$ (that is, one specific mask/distribution) then \eqref{eq:loglikVAEAC} corresponds to the objective function of \texttt{CVAE} \parencite{CVAE_Sohn}. Thus, \texttt{VAEAC} generalizes both the \texttt{VAE} and \texttt{CVAE}.

\subsection{The Employment Phase of the \texttt{VAEAC} Method}
\label{subsub:ModelDescriptionVAEAC}
\begin{figure}[!t]
    \centering
    \includegraphics[width=1\textwidth]{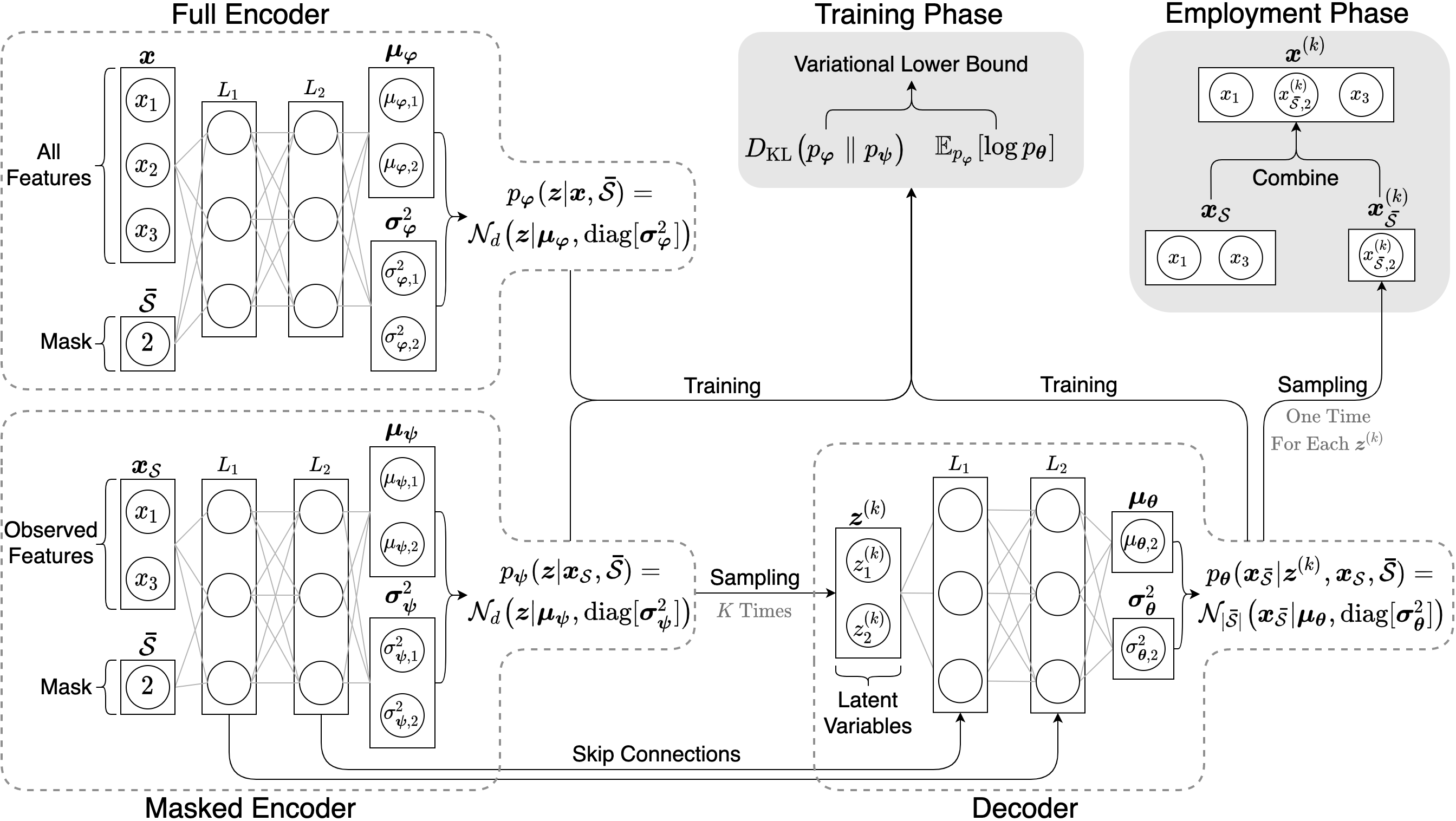}
    \caption{{\small A visual representation of the \texttt{VAEAC} model used in the example given at the end of \Cref{subsub:ModelDescriptionVAEAC}. The model has parameters $\texttt{depth} = 2$, $\texttt{width} = 3$, $d = \texttt{latent\_dim} = 2$ for an $M=3$-dimensional continuous input $\x = \{x_1, x_2, x_3\}$ with coalition $\mathcal{S} = \{1,3\}$ and mask $\thickbar{\mathcal{S}} = \{2\}$. All networks are used during the training phase, but only the masked encoder and decoder are used in the employment phase to generate the conditional samples $\x_{\thickbar{\mathcal{S}}}^{(k)} \sim p_{\bpsi, \btheta}(\boldsymbol{x}_{\thickbar{\mathcal{S}}} | \boldsymbol{x}_{\mathcal{S}} = \boldsymbol{x}_{\mathcal{S}}^*)$, for $k = 1,...,K$, where $K$ is the number of Monte Carlo samples in \eqref{eq:KerSHAPConditionalFunction}.}}
    \label{fig:VAEAC}
\end{figure}

The generative process of the \texttt{VAEAC} method is a two-step procedure, similar to that of \texttt{VAE} and \texttt{CVAE}, and constitutes the \textit{employment phase} where we generate the conditional samples $\x_{\sbb}$ in \eqref{eq:KerSHAPConditionalFunction}. First, we generate a latent representation $\z \sim p_\bpsi(\z|\xs, \sbb) \in \R^d$ using the masked encoder, where the dimension of the latent space, $d$, is specified by the user of \texttt{VAEAC}. Second, we sample the unobserved features from $\xsb \sim p_\btheta(\xsb | \z, \xs, \sbb)$ using the decoder, as illustrated in \Cref{fig:VAEAC_abstract_latent} and \Cref{fig:VAEAC}. This process marginalizes over the latent variables, resulting in the following distribution over the unobserved features:
\begin{align}
\label{eq:VAEAC:marginal_dist}
    p_{\bpsi, \btheta}(\xsb | \xs, \sbb) 
    =
    \E_{\z \sim p_\bpsi(\z|\xs, \sbb)} \big[{p_{\btheta}(\xsb |\z, \xs, \sbb)}\big],
\end{align}
which is similar to the procedures in \texttt{VAE}s and \texttt{CVAE}s. 

The continuous latent variables $\z$ of the masked encoder are assumed to be distributed according to a multivariate Gaussian distribution with a diagonal covariance matrix, that is, $p_{\boldsymbol{\psi}}(\boldsymbol{z} | \boldsymbol{x}_{\mathcal{S}}, \thickbar{\mathcal{S}}) = \mathcal{N}_d\big(\boldsymbol{z} | \boldsymbol{\mu}_{\boldsymbol{\psi}}(\boldsymbol{x}_\mathcal{S}, \thickbar{\mathcal{S}}), \operatorname{diag}[\boldsymbol{\sigma}_{\boldsymbol{\psi}}^2(\boldsymbol{x}_\mathcal{S}, \thickbar{\mathcal{S}})] \big)$. Here $\boldsymbol{\mu}_\bpsi$ and $\boldsymbol{\sigma}_\bpsi$ are the $d$-dimensional outputs of the NN for the masked encoder $p_\bpsi$. As the output nodes of NNs are unbounded, we apply the soft-plus function\footnote{$\operatorname{softplus}(x) = \log(1 + \exp(x))$.} to $\boldsymbol{\sigma}_\bpsi$ to ensure that the standard deviations are positive. We directly send any continuous features of $\xs$ to the masked encoder, while any categorical features are one-hot encoded first, see \Cref{app:subsec:VAEAC_implementaion_length}. One-hot encoding is the most widely used coding scheme for categorical features in NNs \parencite{onehotencoding}. Note that the components of the latent representation $\z$ are conditionally independent given $\xs$ and $\sbb$.

The full encoder is also assumed to be a multivariate Gaussian distribution with a diagonal covariance matrix, that is, $p_{\boldsymbol{\phi}}(\boldsymbol{z} | \boldsymbol{x}, \thickbar{\mathcal{S}}) = \mathcal{N}_d\big(\boldsymbol{z} | \boldsymbol{\mu}_{\boldsymbol{\phi}}(\boldsymbol{x}, \thickbar{\mathcal{S}}), \operatorname{diag}[\boldsymbol{\sigma}_{\boldsymbol{\phi}}^2(\boldsymbol{x}, \thickbar{\mathcal{S}})] \big)$. The full encoder is conditioned on the full observation $\x$, in contrast to the masked encoder which is solely conditioned on the observed features $\xs$. Furthermore, it also uses one-hot encoding to handle any categorical features. The full encoder is only used during the training phase of the \texttt{VAEAC} method, which will become clear in \Cref{subsub:Training}.

The decoder $p_{\boldsymbol{\theta}}(\boldsymbol{x}_{\thickbar{\mathcal{S}}} | \boldsymbol{z}, \boldsymbol{x}_{{\mathcal{S}}}, \thickbar{\mathcal{S}})$ is also assumed to follow a multivariate Gaussian structure with a diagonal covariance matrix for the continuous features of $\boldsymbol{x}_{\thickbar{\mathcal{S}}}$, that is, $p_\btheta(\xsb | \z, \xs, \sbb) = \mathcal{N}\big(\boldsymbol{x}_{\thickbar{\mathcal{S}}} | \boldsymbol{\mu}_{\boldsymbol{\theta}}(\z, \xs, \sbb), \operatorname{diag}[\boldsymbol{\sigma}_{\boldsymbol{\theta}}^2(\z, \xs, \sbb)] \big)$. We discuss this assumption in \Cref{sec:Conclusion}. As the latent variables $\z$ are continuous and $p_{\btheta}(\xsb |\z, \xs, \sbb)$ is in this case Gaussian, the distribution $p_{\bpsi, \btheta}(\xsb | \xs, \sbb)$ in \eqref{eq:VAEAC:marginal_dist} can be seen as an infinite mixture of Gaussian distributions \parencite[p.\ 12]{Kingma2019AnIT}, which is a universal approximator of densities, in the sense that any smooth density can be arbitrarily well approximated \parencite[p.\ 65]{Goodfellow-et-al-2016}. We illustrate this property in \Cref{Appendix:sec:UniversalApproximator}.

The Gaussian structure is incompatible with categorical features. Assume for now that $x_j$ is categorical with $L$ categories. When working with multi-class features and NNs, it is common practice to let the outputs $\w_{\btheta, j}(\z, \xs, \sbb)$ of the NN be the logits of the probabilities for each category. That is, $\w_{\btheta, j}(\z, \xs, \sbb) = \operatorname{logit}(\boldsymbol{p}_j) = \log [{\boldsymbol{p}_j}/{(1-\boldsymbol{p}_j)}]$ is an $L$-dimensional vector of the logits for each of the $L$ categories for the categorical $j$th feature. As logits are unbounded, we use the softmax function\footnote{$\operatorname{softmax}(\w_{\btheta, j})_i = {\exp\{\w_{\btheta, j, i}\}}\big/{\sum_{l=1}^L \exp\{\w_{\btheta,j,l}\}}$, for $i = 1, \dots,L$.} to map the logits to probabilities in $(0,1)$. Finally, we use the categorical distribution with $L$ categories to model each of the categorical features, that is, $x_j \sim \operatorname{Cat}\left( \operatorname{Softmax} (\w_{\btheta,j}(\z, \xs, \sbb)) \right)$. Note that even though we used one-hot encoding in the encoders, the sampled categorical features will be on the original categorical form and \textit{not} in the one-hot encoded representation. Furthermore, note that the components of $\xsb$ are conditionally independent given $\z, \xs$, and $\sbb$. 

We now give a walk-through of the \texttt{VAEAC} model illustrated in \Cref{fig:VAEAC} for $M=3$-dimensional continuous data. The notation and representation of the network architecture is simplified for ease of understanding. The complete implementation details are found in \Cref{app:subsec:VAEAC_implementaion_length}. The aim is to sample $\x_{\thickbar{\mathcal{S}}}^{(k)} \sim p_{\bpsi, \btheta}(\boldsymbol{x}_{\thickbar{\mathcal{S}}} | \boldsymbol{x}_{\mathcal{S}} = \boldsymbol{x}_{\mathcal{S}}^*)$, for $k = 1,...,K$, where $K$ is the number of Monte Carlo samples in \eqref{eq:KerSHAPConditionalFunction}. For the $M=3$-dimensional setting, there are $2^M=8$ possible coalitions $\s$, that is, $\emptyset, \{1\}, \{2\}, \{3\}, \{1, 2\}, \{1, 3\}, \{2,3\}, \{1,2,3\}$. However, the edge-cases are treated differently in the \textsc{R}-package \texttt{shapr} \parencite{shapr}, which we use to compute the Shapley values. For $\s = \M = \{1,2,3\}$, we have that $\xs = \x$ and $v(\mathcal{M}) = f(\x)$ by definition.
For the other edge-case, $\s = \emptyset$, we have by definition that $\phi_0 = v(\emptyset) = \E[f(\x)]$, which \texttt{shapr} lets the user specify. A common estimate is the average training response \parencite{aas2019explaining}, which we also use in this article. In Figure 2, we illustrate the \texttt{VAEAC} method when given coalition $\s = \{1,3\}$, that is, mask $\thickbar{\mathcal{S}} = \{2\}$, but the procedure is identical for the other coalitions.

Let $\x = \{x_1, x_2, x_3\}$ denote the full observation, where the associated observed and unobserved feature vectors are $\xs = \{x_1, x_3\}$ and $\xsb = \{x_2\}$, respectively. We start by sending $\xs$ and $\sbb$ through the masked encoder, which yields the latent probabilistic representation $p_{\boldsymbol{\psi}}(\boldsymbol{z} | \boldsymbol{x}_{\mathcal{S}}, \thickbar{\mathcal{S}}) = \mathcal{N}_d\big(\boldsymbol{z} | \boldsymbol{\mu}_{\boldsymbol{\psi}}(\boldsymbol{x}_\mathcal{S}, \thickbar{\mathcal{S}}), \operatorname{diag}[\boldsymbol{\sigma}_{\boldsymbol{\psi}}^2(\boldsymbol{x}_\mathcal{S}, \thickbar{\mathcal{S}})] \big)$, where $d=2$. We sample $K$ latent representations $\z^{(k)} \sim p_{\boldsymbol{\psi}}(\boldsymbol{z} | \boldsymbol{x}_{\mathcal{S}}, \thickbar{\mathcal{S}})$, which are then sent through the decoder. This yields $K$ different distrbutions $p_\btheta(\xsb | \z^{(k)}, \xs, \sbb) = \mathcal{N}\big(\boldsymbol{x}_{\thickbar{\mathcal{S}}} | \boldsymbol{\mu}_{\boldsymbol{\theta}}(\z^{(k)}, \xs, \sbb), \operatorname{diag}[\boldsymbol{\sigma}_{\boldsymbol{\theta}}^2(\z^{(k)}, \xs, \sbb)] \big)$, as the means $\boldsymbol{\mu}_{\boldsymbol{\theta}}(\z^{(k)}, \xs, \sbb)$ and variances $\boldsymbol{\sigma}_{\boldsymbol{\theta}}^2(\z^{(k)}, \xs, \sbb)$ are functions of $\z^{(k)}$. When $K \to \infty$, we obtain an infinite mixture of Gaussian distributions, see \Cref{Appendix:sec:UniversalApproximator}. We sample \textit{one} observation from each inferred distribution, that is, $\x_{\thickbar{\mathcal{S}}}^{(k)} \sim p_\btheta(\xsb | \z^{(k)}, \xs, \sbb)$. Sampling more values is also possible. \textcite{ivanov_variational_2018} call this process of repeatedly generating the missing values $\xsb$ based on $\xs$ for \textit{missing features multiple imputation}.

\subsection{The Training Phase of the \texttt{VAEAC} Method}
\label{subsub:Training}
The model parameters $\bpsi$ and $\btheta$ are not easily tuned by maximizing the log-likelihood \eqref{eq:loglikVAEAC}, as this optimization problem is challenging due to intractable posterior inference. \textcite[Ch.\ 2]{Kingma2019AnIT} discuss the same issue for the \texttt{VAE}. The optimization problem is bypassed by \textit{variational inference}\footnote{Also called \textit{variational Bayes} and \texttt{VAE}s belong to the family of \textit{variational Bayesian methods}.} \parencite{Blei_2017} where we rather optimize the \textit{variational lower bound} (VLB). The VLB is also called the \textit{evidence lower bound} (ELBO) in variational inference. The main contribution of \textcite{kingma2014autoencoding} is their reparameterization trick of \texttt{VAE}'s latent variables. This allows for efficient fitting of \texttt{VAE}'s model parameters and optimization of the VLB by using backpropagation and stochastic gradient descent \parencite[Ch.\ 6 \& 8]{Goodfellow-et-al-2016}. The same type of techniques are used to optimize \texttt{VAEAC}'s model parameters.

Like for \texttt{VAE}s, we can derive a variational lower bound for \texttt{VAEAC}:
\begin{align}
    \label{eq:VLBVAEAC}
    \begin{split}
    \log p_{\bpsi, \btheta} (\xsb|\xs, \sbb)
    &=
    \E_{p_\bphi(\z|\x,\sbb)} \sqbr{\log \frac{p_{\bpsi, \btheta}(\xsb, \z|\xs, \sbb)}{p_\bphi(\z|\x,\sbb)}} 
    + 
    \E_{p_\bphi(\z|\x,\sbb)} \sqbr{\log \frac{p_\bphi(\z|\x,\sbb)}{p_{\bpsi, \btheta}(\z|\x, \sbb)}} \\
    &\ge
    \E_{p_\bphi(\z|\x,\sbb)} \sqbr{\log p_{\btheta}(\xsb| \z,\xs, \sbb)} - 
    D_\text{KL} \big(p_\bphi(\z|\x,\sbb) \;\|\; p_{\bpsi}(\z|\xs, \sbb)\big) \\
    &= 
    \mathcal{L}_{\texttt{VAEAC}} (\x, \sbb | \btheta, \bpsi, \bphi), 
    \end{split}
\end{align}
where the Kullback–Leibler divergence, $D_\text{KL}$, for two continuous probability distributions $p$ and $q$, is given by $D_\text{KL}\big(p \;\|\; q\big) = \int_{-\infty}^{\infty} p(\z) \log\tfrac{p(\z)}{q(\z)} \diff \z = \E_{p(\z)} \sqbr{\log \tfrac{p(\z)}{q(\z)}} \ge 0$.
We obtain a \textit{tight} variational lower bound if $p_\bphi(\z|\x,\sbb)$ is close to $p_{\bpsi, \btheta}(\z|\x, \sbb)$ with respect to the Kullback–Leibler divergence. The full derivation of \eqref{eq:VLBVAEAC} is presented in \Cref{Appendix:VLB}.

Instead of maximizing the log-likelihood in \eqref{eq:loglikVAEAC}, we rather have the following variational lower bound optimization problem for \texttt{VAEAC}: 
\begin{align}
    \label{eq:OptimazationProblemVAEAC}
    \max_{\btheta, \bpsi, \bphi} \E_{p(\x)}\E_{p(\sbb)} \left[ \mathcal{L}_{\texttt{VAEAC}} (\x, \sbb | \btheta, \bpsi, \bphi) \right]
    = 
    \min_{\btheta, \bpsi, \bphi} \E_{p(\x)}\E_{p(\sbb)} \left[-\mathcal{L}_{\texttt{VAEAC}} (\x, \sbb | \btheta, \bpsi, \bphi) \right]\!,
\end{align}
which is optimized similarly as \texttt{VAE}s. Since we assumed $p_\bphi$ and $p_{\bpsi}$ to be multivariate Gaussians, we have a closed-form expression for the Kullback–Leibler divergence in \eqref{eq:VLBVAEAC}. The parameters $\bpsi$ and $\btheta$ can be directly optimized using Monte Carlo estimates, backpropagation, and stochastic gradient descent.

The gradient of $\mathcal{L}_{\texttt{VAEAC}} (\x, \sbb | \btheta, \bpsi, \bphi)$ with respect to the variational parameter $\bphi$ is more complicated to compute, as the expectation in \eqref{eq:VLBVAEAC} is taken with respect to $p_\bphi(\z|\x,\sbb)$, which is a function of $\bphi$. However, the gradient can be efficiently estimated using the reparameterization trick of \textcite{kingma2014autoencoding}. The trick is to rewrite the latent variable vector as $\z = \boldsymbol{\mu}_{\boldsymbol{\varphi}}(\boldsymbol{x}, \thickbar{\mathcal{S}}) + \boldsymbol{\epsilon} \boldsymbol{\sigma}_{\boldsymbol{\varphi}}(\boldsymbol{x}, \thickbar{\mathcal{S}})$, where $\boldsymbol{\epsilon}\sim \mathcal{N}_d(0, I)$ and $\boldsymbol{\mu}_{\boldsymbol{\varphi}}$ and $\boldsymbol{\sigma}_{\boldsymbol{\varphi}}$ are deterministic functions parameterized by the NN of the full encoder. That is, the randomness in $\z$ has been \textit{externalized} to $\boldsymbol{\epsilon}$ by reparameterizing $\z$ as a deterministic and differentiable function of $\boldsymbol{\varphi}$, $\x$, $\thickbar{\mathcal{S}}$, and the random variable $\boldsymbol{\epsilon}$. Thus, the gradient is
\begin{align*}
    \frac{\partial \mathcal{L}_{\texttt{VAEAC}} (\x, \sbb | \btheta, \bpsi, \bphi)}{\partial \bphi} 
    &=
    \E_{\boldsymbol{\epsilon}\sim \mathcal{N}_d(0, I)} 
    \sqbr{\frac{\partial}{\partial \bphi} \log p_{\btheta}(\xsb| \boldsymbol{\mu}_{\boldsymbol{\varphi}}(\boldsymbol{x}, \thickbar{\mathcal{S}}) + \boldsymbol{\epsilon} \boldsymbol{\sigma}_{\boldsymbol{\varphi}}(\boldsymbol{x}, \thickbar{\mathcal{S}}),\xs, \sbb)} \\
    &
    \phantom{LBO} -
    \frac{\partial}{\partial \bphi}D_\text{KL} \big(p_\bphi(\z|\x,\sbb) \;\|\; p_{\bpsi}(\z|\xs, \sbb)\big),
\end{align*}
where the first term is estimated using Monte Carlo sampling, while we use the closed-form expression mentioned above for the second term.

\subsubsection{Prior in Latent Space}
\label{subsubsec:PriorInLatentSpace}

When optimizing the variational lower bound in \eqref{eq:OptimazationProblemVAEAC}, there are no restrictions on the size of the estimated mean $\boldsymbol\mu_\bpsi$ and standard deviation $\boldsymbol\sigma_\bpsi$ in the masked encoder. That is, they can grow to infinity and cause numerical instabilities. We follow \textcite{ivanov_variational_2018} and solve this problem by adding a normal prior on $\boldsymbol\mu_\bpsi$ and gamma prior on $\boldsymbol\sigma_\bpsi$ to prevent divergence. That is, we introduce a restricted version of $p_\bpsi(\z, \boldsymbol{\mu}_\bpsi, \boldsymbol{\sigma}_\bpsi| \xs, \sbb)$: 
\begin{equation*}
\resizebox{\textwidth}{!}{$p_\bpsi^{\text{res}}(\z, \boldsymbol{\mu}_\bpsi, \boldsymbol{\sigma}_\bpsi| \xs, \sbb) 
    =
    \mathcal{N}_d\big(\z | \boldsymbol{\mu}_\bpsi(\xs, \sbb), \operatorname{diag}[\boldsymbol{\sigma}_\bpsi^2(\xs, \sbb)] \big) \mathcal{N}\big(\boldsymbol{\mu}_{\bpsi} | 0, \sigma_{\mu}^2 \big)
    \Gamma(\boldsymbol{\sigma}_{\bpsi}| 1+\sigma_{\sigma}^{-1}, \sigma_{\sigma}^{-1}).$}
\end{equation*}
Here $\Gamma$ is the gamma distribution with shape parameter $1+\sigma_{\sigma}^{-1}$ and rate parameter $\sigma_{\sigma}^{-1}$, that is, the mean is $1+\sigma_\sigma$ and the variance is $\sigma_\sigma(1+\sigma_\sigma)$. The hyperparameters ${\sigma}_{\mu}$ and $\sigma_{\sigma}$ are set to be large, so as not to affect the learning process significantly.

When re-deriving the variational lower bound \eqref{eq:VLBVAEAC} with $p_\bpsi^{\text{res}}$ instead of $p_\bpsi$, we get two additional regularizing terms in the VLB, after removing fixed constants. Thus, we maximize the following VLB in \eqref{eq:OptimazationProblemVAEAC}:
\begin{align*}
    \begin{split}
    \mathcal{L}_{\texttt{VAEAC}}^\text{res} (\x, \sbb | \btheta, \bpsi, \bphi)
    &= 
    \E_{p_\bphi(\z|\x,\sbb)} \sqbr{\log p_{\btheta}(\xsb| \z,\xs, \sbb)} - 
    D_\text{KL} \big(p_\bphi(\z|\x,\sbb) \;\|\; p_{\bpsi}(\z|\xs, \sbb)\big) \\
    &\phantom{=} \,\,-\frac{\boldsymbol{\mu}_{\boldsymbol{\psi}}^2}{2\sigma_{\mu}^2} + \frac{1}{\sigma_{\sigma}}(\log (\boldsymbol{\sigma}_\bpsi) - \boldsymbol{\sigma}_\bpsi).
    \end{split}
\end{align*}
Large values of $\boldsymbol{\mu}_{\boldsymbol{\psi}}^2$ and $\boldsymbol{\sigma}_\bpsi$ decrease the new VLB through the last two terms, but their magnitude is controlled by the hyperparameters ${\sigma}_{\mu}$ and $\sigma_{\sigma}$. \textcite[Eq.\ 8]{ivanov_variational_2018} obtain identical regularizers, but there is a misprint in their shape parameter of the gamma prior.

\subsection{Masking Schemes}
\label{subseq:MaskingSchemes} 

The default version of the \texttt{VAEAC} method uses the uniform masking scheme, i.e., $p(\sbb) = 2^{-M}$. This means that we can use a $\operatorname{Bernoulli}(0.5)$ to model whether feature $j$ is in $\s$ or not. As $p(\sbb)$ has full support over $\pow(\mathcal{M})$, the \texttt{VAEAC} method can evaluate arbitrary conditioning. However, we can make \texttt{VAEAC} focus more of its estimation efforts on certain coalitions by defining a non-uniform masking scheme. Recall that the mask $\sbb$ and the corresponding coalition $\s$ are uniquely related by $\sbb  = \M \backslash \s$. 

In high-dimensional settings, it is common to compute Shapley values based on a set $\mathcal{C}$ of coalitions sampled from $\pow(\M)$ with replacements \parencite{lundberg2017unified, williamson20a, aas2019explaining, frye_shapley-based_2020, mitchell2022sampling}. In these settings, we can specify a masking scheme based on the sampling frequency of the coalitions in $\mathcal{C}$. More precisely, we let $p(\sbb) = p(\mathcal{M}\backslash\mathcal{S}) = \tfrac{1}{|\mathcal{C}|}\sum_{c \in \mathcal{C}} \mathbf{1} (c = \mathcal{S})$, where $\mathbf{1}$ is the indicator function, and we denote the corresponding \texttt{VAEAC} method by $\texttt{VAEAC}_\mathcal{C}$. Thus, the $\texttt{VAEAC}_\mathcal{C}$ approach will, in contrast to the regular $\texttt{VAEAC}$ approach, \emph{not} try to learn the conditional distributions $p(\xsb|\xs)$ for all $\s \in \mathcal{P}(\mathcal{M})$, but rather only for the coalitions in $\mathcal{C}$. If coalition $\s$ has been sampled twice as often as $\s'$, then the training samples will be twice as often artificially masked by $\sbb$ than $\sbb'$ during the training phase of the $\texttt{VAEAC}_\mathcal{C}$ method. That is, the $\texttt{VAEAC}_\mathcal{C}$ approach trains twice as much on learning $p(\xsb | \xs)$ compared to $p(\x_{\thickbar{\s'}} | \x_{\s'})$. The hyperparameters of $\texttt{VAEAC}_\mathcal{C}$ are the same as those described for $\texttt{VAEAC}$ in \Cref{app:subsec:VAEAC_implementaion_hyper}. In \Cref{sec:SimulationStudy}, we will see that this new non-uniform masking scheme significantly improves the $\texttt{VAEAC}$ methodology in the high-dimensional simulation studies.

\subsection{Missing Feature Values}
\label{subseq:MissingFeatures}
Incomplete data is common for real-world data sets, that is, some feature values for some observations are missing, denoted by \texttt{NA}. Many classical statistical models rely on complete data in their training phase, e.g., linear models and GAMs, while some modern ML models handles missing data, e.g., XGBoost \parencite{chen2015xgboost}. Consider a model of the latter type trained on incomplete data. In the Shapley value explanation framework, the approach used to estimate the feature dependencies should also handle incomplete data in the training phase to fully utilize all available data. In contrast to the competing conditional approaches used in \Cref{sec:SimulationStudy}, the \texttt{VAEAC} approach directly supports incomplete training data. The original purpose of the \texttt{VAEAC} method was to generate multiple imputations for missing entries in incomplete data sets, while simultaneously preserving the feature dependencies. 

We do not investigate \texttt{VAEAC}'s handling of missing data in the current article, as we want a fair comparison with the competing methods, which lack direct support for missing data. However, we give a brief outline of how \texttt{VAEAC} handles missing feature values in the training phase below. Additionally, in \Cref{sec:Conclusion}, we discuss how the \texttt{VAEAC} method can be extended to also support incomplete data in the employment phase. 

We refer to \textcite{ivanov_variational_2018} for the fine details and evaluation of \texttt{VAEAC}'s performance when applied on incomplete training data. However, the main points are the following. First, \texttt{VAEAC} ensures that the missing feature values are always unobserved in the training phase by masking them on an individual basis. Hence, the mask distribution must be conditioned on $\x$, that is, $p(\sbb)$ turns into $p(\sbb|\x)$ in \eqref{eq:loglikVAEAC} and \eqref{eq:OptimazationProblemVAEAC}. Second, the \texttt{NA} values of $\xsb$ must be excluded when computing the $\log p_{\btheta}(\xsb| \z,\xs, \sbb)$ term in \eqref{eq:VLBVAEAC}. This is solved by marginalizing out the missing features, that is, $\log p_{\btheta}(\xsb| \z,\xs, \sbb) = \sum_{i: i\in\sbb, x_i \neq \texttt{NA}} \log p_{\btheta}(\x_i| \z,\xs, \sbb)$. Finally, to ensure that the full encoder can distinguish between missing and masked features during training, \textcite{ivanov_variational_2018} add an additional \textit{missing feature mask} to the full encoder. That is, the full encoder receives the missing feature mask, the unobserved/masked feature mask $\sbb$, and the training instance $\x$ where \texttt{NA} values are set to zero.


\subsection{Skip Connections}
\label{subsec:SkipConnections}

Our \texttt{VAEAC} model includes skip connections with concatenation between the layers of the masked encoder and the decoder, as seen in \Cref{fig:VAEAC}. A problem with early deep neural networks was that they suffered from the \textit{vanish gradient problem} in the backpropagation step. This means that the model parameters of the first layers would not be updated. Skip connections is one of several possible architectural tools (other tools are activation functions such as LeakyReLU) used to ensure effective learning of all model parameters \parencite{ronneberger2015u, mao2016image}. 

By using skip connections, we allow for information to pass directly from the masked encoder to the decoder without going through the latent space. \textcite{ivanov_variational_2018} report that they saw an improvement in the log-likelihood when they included skip connections in the \texttt{VAEAC} model. They describe the latent variables as being responsible for the global information, while skip connections pass through local information.


\section{Simulation Studies}
\label{sec:SimulationStudy}

A major problem of evaluating explanation methods is that there is in general no ground truth for authentic real-world data. In this section, we simulate data for which we can compute the true Shapley values \eqref{eq:ShapleyValuesDef}. We then compare how close the estimated Shapley values are to the corresponding true ones. However, this evaluation methodology is intractable in high-dimensional settings due to the exponential number of terms in the Shapley value formula. Furthermore, the expectations in \eqref{eq:ContributionFunc} rely on solving integrals of increasing dimensions which requires more Monte Carlo samples in \eqref{eq:KerSHAPConditionalFunction} to maintain the precision. In high-dimensional settings, we rather compare how close a sampled set of estimated contribution functions \eqref{eq:KerSHAPConditionalFunction} are to the true counterparts \eqref{eq:ContributionFunc}.

We conduct and discuss two types of simulation studies: one for continuous data (\Cref{subsection:simstudy:cont}) and one with mixed data (\Cref{subsection:simstudy:mixed}). For the sake of completeness, a simulation study limited to categorical features has also been investigated, but as real-world data sets are seldom restricted to only categorical features, we rather report the corresponding results in \Cref{subsection:simstudy:cat}. In both simulation studies, we first look at exact low-dimensional settings before progressing to sampled high-dimensional settings.

For the continuous data setting, we compare our \texttt{VAEAC} approach to Shapley value estimation with the \texttt{independence} method from \textcite{lundberg2017unified}, the \texttt{empirical}, \texttt{Gaussian}, and \texttt{copula} approaches from \textcite{aas2019explaining}, and the \texttt{ctree} method from \textcite{redelmeier:2020}. These methods are briefly described in \Cref{Appendix:AlternativeApproaches}. For the mixed data setting, with continuous and categorical features, the methods of \textcite{aas2019explaining} are dismissed as they lack support for categorical data. Use of feature encodings, such as one-hot encoding, is possible, but drastically increases the computational time, due to the detour of estimating Shapley values for each dummy feature \parencite[q.v.\ Table.\ 4 \& 6]{redelmeier:2020}, and were thus deemed infeasible. This \textit{external} use of one-hot encoding must not be confused with \texttt{VAEAC}'s \textit{internal} use of one-hot encoding for handling categorical data, which was described in \Cref{subsub:ModelDescriptionVAEAC}. The internal use does \emph{not} result in computing excess Shapley values for dummy features, as \texttt{VAEAC} can receive and generate categorical data. In the high-dimensional simulation studies, we also include the novel $\texttt{VAEAC}_\mathcal{C}$ approach described in \Cref{subseq:MaskingSchemes}.

All mentioned methodologies are implemented in the software package \texttt{shapr} \parencite{shapr} (version 0.2.0) in the \textsc{R} programming language \parencite{R_language}, with default hyperparameters. See \Cref{Appendix:Architecture} for the choice of \texttt{VAEAC}'s hyperparameters. \textcite{ivanov_variational_2018} have implemented \texttt{VAEAC} in Python \parencite{Python_language}, see \url{https://github.com/tigvarts/vaeac}. We have made changes to their implementation and used \texttt{reticulate} \parencite{reticulate} to run Python code in \textsc{R}, such that our \texttt{VAEAC} and $\texttt{VAEAC}_\mathcal{C}$ approaches run on top of the \texttt{shapr}-package. Our version of \texttt{VAEAC} and $\texttt{VAEAC}_\mathcal{C}$ can be accessed on \url{https://github.com/LHBO/ShapleyValuesVAEAC}.

\subsection{Evaluation Criteria}
\label{subsec:simulations:Evaluation_method}
We consider three types of evaluation criteria which we describe in this section. The first, EC1, is the mean absolute error (MAE), averaged over all test observations and features:
\begin{align}
    \label{eq:MAE}
    \operatorname{EC1} 
    = 
    \operatorname{MAE}_{\phi}(\text{method } \texttt{q}) 
    =
    \frac{1}{N_\text{test}} \sum_{i=1}^{N_\text{test}} \frac{1}{M} \sum_{j=1}^M |\phi_{j, \texttt{true}}(\boldsymbol{x}^{[i]}) - \hat{\phi}_{j, \texttt{q}}(\boldsymbol{x}^{[i]})|.  
\end{align}
Here, $\boldsymbol{\phi}_{\texttt{true}}$ and $\hat{\boldsymbol{\phi}}_{\texttt{q}}$ are the true Shapley values and the Shapley values computed using method \texttt{q}, respectively. The true Shapley values are not known in general. However, in our simulation studies, the conditional distribution function $p_\texttt{true}$ is analytically known and samplable. This means that we can compute the true contribution function $v_\texttt{true}(\s)$ in \eqref{eq:ContributionFunc} by sampling $\boldsymbol{x}_{\thickbar{\mathcal{S}}, \texttt{true}}^{(k)} \sim p_\texttt{true}(\xsb|\xs)$ and using \eqref{eq:KerSHAPConditionalFunction}, for $\s \in \pow(\M)$. The true Shapley values are then obtained by inserting the $v_\texttt{true}(\s)$ quantities into the Shapley formula \eqref{eq:ShapleyValuesDef}. The $v_\texttt{true}(\s)$ can be arbitrarily precise by choosing a sufficiently large number of Monte Carlo samples $K$. The Shapley values for method \texttt{q} are computed in the same way, but using the conditional distributions estimated using approach \texttt{q} instead of the true ones. 

The second criterion, EC2, measures the mean squared-error between $v_\texttt{true}$ and $v_\texttt{q}$, i.e.,
\begin{align}
    \label{eq:MAE_contribution_function}
    \operatorname{EC2} 
    = 
    \operatorname{MSE}_{v}(\text{method } \texttt{q}) 
    =
    \frac{1}{N_\text{test}} \sum_{i=1}^{N_\text{test}} \frac{1}{N_\s} \sum_{\s \in \pow^*(\mathcal{M})} \left( v_{\texttt{true}}(\s, \boldsymbol{x}^{[i]}) - \hat{v}_{\texttt{q}}(\s, \boldsymbol{x}^{[i]})\right)^2.  
\end{align}
Here $\pow^*(\mathcal{M}) = \pow(\mathcal{M}) \backslash \{\emptyset, \M\}$, that is, we skip $v(\emptyset)$ and $v(\M)$ as they are the same regardless of which method \texttt{q} that is used. Thus, $N_\s = |\pow^*(\mathcal{M})| = 2^M-2$.

The final criterion, EC3, is given by
\begin{align}
    \label{eq:EC3}
    \operatorname{EC3} 
    = 
    \operatorname{EPE}_{v}(\text{method } \texttt{q}) 
    =
    \frac{1}{N_\text{test}} \sum_{i=1}^{N_\text{test}} \frac{1}{N_\mathcal{S}} \sum_{\s \in \pow^*(\mathcal{M})} \left( f(\boldsymbol{x}^{[i]}) - \hat{\v}_{\texttt{q}}(\s, \boldsymbol{x}^{[i]})\right)^2\!,
\end{align}
which is based on \eqref{eq:MAE_contribution_function} being an estimator of 
\begin{align}
    \label{eq:expectation_decomposition}
    \E_\s\E_{\x} (v_{\texttt{true}}(\s, \boldsymbol{x})- \hat{v}_{\texttt{q}}(\s, \boldsymbol{x}))^2 
    =
    \E_\s\E_{\x} (f(\x) - \hat{v}_{\texttt{q}}(\s, \boldsymbol{x}))^2 - \E_\s\E_{\x} (f(\x)-v_{\texttt{true}}(\s, \boldsymbol{x}))^2,
\end{align}
where the decomposition follows from \textcite[Appendix A]{covert2020understanding}. The first term on the right hand side of \eqref{eq:expectation_decomposition} corresponds to \eqref{eq:EC3}, while the second term is a fixed (unknown) constant not influenced by approach \texttt{q}.

An advantage of the EC3 is that $v_\texttt{true}$ is not involved, which means that it also may be used for real-world data. However, this criterion has two main drawbacks. First, it can only be used to rank different approaches and not assess how close they are to the optimal method. This is because we do not know the minimum value of this criterion. The second drawback, which it has in common with the EC2, is that it evaluates contribution functions and not Shapley values. The estimates for $v(\s)$ can overshoot for some coalitions and undershoot for others, and such errors may cancel each other out in the Shapley formula.

In high-dimensional settings, it is intractable to consider all $2^M$ coalitions $\s \in \pow(\M)$ in \eqref{eq:MAE}--\eqref{eq:EC3}. In such cases, we sample $N_\s$ coalitions $\s$ from $\pow(\M)$ with replacement and weights equal to $\tfrac{|\s|!(M-|\s|-1)!}{M!}$, that is, the weights in the Shapley value formula \eqref{eq:ShapleyValuesDef}. We use this subset of coalitions, denoted by $\mathcal{C}$, instead of $\pow(\M)$ when computing the three evaluation criteria.

\subsection{Simulation Study: Continuous Data}
\label{subsection:simstudy:cont}
For the continuous simulation study, we generate dependent features from the multivariate Burr distribution, which is briefly explained in \Cref{Appendix:sub:GenerateDataContinuous}. The Burr distribution is skewed, strictly positive, and has known conditional distributions. Thus, we can compute the true Shapley values with arbitrary precision by using Monte Carlo integration. This simulation study is an extension of the one conducted in \textcite{aas2021explaining}.

The full description of the data generating process is given in \Cref{Appendix:sub:GenerateDataContinuous}, but the essentials are presented here. We sample 
$N_\text{train}$ training and $N_\text{test}$ test observations from a $\operatorname{Burr}(\kappa, \boldsymbol{b}, \boldsymbol{r})$ distribution. The scale parameter $\kappa$ is set to $2$, which yields an average pair-wise Kendall's $\tau$ coefficient of $0.20$, that is, moderate feature dependence. We sample $\boldsymbol{b}, \boldsymbol{r}\in \mathbb{R}^M$ from plausible values to make the setup work for any $M$ divisible by five. We generate the response $y$ according to an extension of the non-linear and heteroscedastic function used in \textcite{aas2021explaining}, which was inspired by \textcite{chang2019prediction}. That is,
\begin{align}
    \label{eq:response_high_dim}
    y = \sum_{k=0}^{(M-5)/5}\left[\sin(\pi c_{3k+1}u_{5k+1}u_{5k+2}) + c_{3k+2}u_{5k+3}\exp\{c_{3k+3}u_{5k+4}u_{5k+5}\}\right] + \varepsilon,
\end{align}
where $\varepsilon$ is added noise, $c_l$ is a sampled coefficient for $l = 1, \dots, \tfrac{3M}{5}$, and $u_j = F_j(x_j)$, for $j = 1,2,\dots, M$. Here $\boldsymbol{x}$ is multivariate Burr distributed and $F_j$ is the true parametric (cumulative) distribution function for the $j$th feature $x_j$. Thus, each $u_j$ is uniformly distributed between $0$ and $1$, but the dependence structure between the features is kept. 

We let the predictive function $f$ be a random forest model with $500$ trees fitted using the \textsc{R}-package \texttt{ranger} \parencite{ranger} with default parameters. We use the different approaches to compute the three evaluation criteria. We repeat this setup $R$ times and report the average EC values, i.e., the quality of the approaches is evaluated based on a total of $R\times N_\text{test}$ test observations. For each approach, we use $K=250$ Monte Carlo samples to estimate the contribution functions in \eqref{eq:KerSHAPConditionalFunction}. We experimented with larger values of $K$ and saw a marginal increase in accuracy, but at the cost of increased computation time, as reported in \Cref{Appendix:sec:NumMonteCarlo}. We found $K=250$ to be a suitable trade-off between precision and computation time. However, we use $K = 5000$ to compute the true counterparts.

\subsubsection{Continuous Data: Low-Dimensional}
\label{subsection:simstudy:cont:lowdimensional}
In the low-dimensional setting, that is, $M\in\{5,10\}$, we use all coalitions $\s \in \mathcal{C} = \pow(\mathcal{M})$. We consider three different training sample sizes $N_\text{train} \in \{100, 1000, 5000\}$, while $N_\text{test} = 100$. The smallest training size is chosen to demonstrate that \texttt{VAEAC} works well even for limited data, in contrast to the common belief that neural networks need large amounts of data to be properly fitted. We repeat each experiment $R = 20$ times for each approach and report the average evaluation criteria over those repetitions.

\begin{table}[!t]
\centering
\small
\begin{adjustbox}{max width=1\textwidth}
\begin{tabular}{@{}ccccccccccc@{}}
\toprule
&&&&\multicolumn{7}{c}{Methods}  \\
\cmidrule{5-11}
$M$ & $N_\s$ & $N_\text{train}$ & Crit. & \texttt{True} & \texttt{Indep}. & \texttt{Empir}. & \texttt{Gauss}. & \texttt{Copula} & \texttt{Ctree} & \texttt{VAEAC} \\
\midrule
& & \multirow{3}{*}{100} &  EC1 &  --- & 0.0967 & 0.1051 & 0.0534 & 0.0568 & 0.0829 & \B0.0492   \\
&&& EC2 & --- &  0.1082 & 0.1242 & 0.0248 & 0.0274 & 0.0684 & \B0.0184 \\
&&&  EC3 & 0.3041 & 0.4107 & 0.4333 & 0.3305 & 0.3310 & 0.3624 & \B0.3199  \\
\cmidrule{3-11}
&& \multirow{3}{*}{1000} &  EC1 &  --- & 0.1095 & 0.0479 & 0.0424 & 0.0472 & 0.0480 & \B0.0264  \\
5&$2^5$&& EC2 & --- &  0.1579 & 0.0499 & 0.0159 & 0.0257 & 0.0325 & \B0.0092  \\
&&&  EC3 &  0.4285 & 0.5821 & 0.4864 & 0.4490 & 0.4575 & 0.4615 & \B0.4402  \\
\cmidrule{3-11}
&&\multirow{3}{*}{5000}&  EC1 &  --- &  0.1157 & 0.0279 & 0.0426 & 0.0473 & 0.0338 & \B0.0239  \\
&&& EC2 & --- &  0.1915 & 0.0183 & 0.0171 & 0.0245 & 0.0182 & \B0.0071  \\
&&&  EC3 & 0.5222 & 0.7049 & 0.5254 & 0.5358 & 0.5425 & 0.5321 & \B0.5230  \\
\midrule
&&\multirow{3}{*}{100} &  EC1 &  --- & 0.0938 & 0.0820 & 0.0464 & 0.0518 & 0.0746 & \B0.0461   \\
&&& EC2 & --- & 0.2005 & 0.1751 & \B0.0266 & 0.0383 & 0.1258 & 0.0303 \\
&&&  EC3 & 0.2468 & 0.4474 & 0.4261 & \B0.2666 & 0.2798 & 0.3773 & 0.2728  \\
\cmidrule{3-11}
&& \multirow{3}{*}{1000}&  EC1 & --- & 0.1082 & 0.0434 & 0.0384 & 0.0456 & 0.0419 & \B0.0288  \\
10& $2^{10}$ &&EC2 & --- &   0.2832 & 0.0952 & 0.0185 & 0.0286 & 0.0687 & \B0.0152  \\
&&&  EC3 & 0.3609 & 0.6680 & 0.4720 & \B0.3728 & 0.3938 & 0.4390 & 0.3795  \\
\cmidrule{3-11}
&&\multirow{3}{*}{5000} &  EC1 &  --- & 0.1160 & 0.0316 & 0.0384 & 0.0459 & 0.0293 & \B0.0218\\
&&& EC2 & --- &  0.3291 & 0.0592 & 0.0183 & 0.0293 & 0.0442 & \B0.0089 \\
&&&  EC3 & 0.4548 & 0.8140 & 0.5294 & 0.4656 & 0.4879 & 0.5041 & \B0.4621\\
\bottomrule
\end{tabular}
\end{adjustbox}
\caption{{\small The average evaluation criteria over the $R=20$ repetitions of the low-dimensional continuous data simulation studies in \Cref{subsection:simstudy:cont:lowdimensional}, based on all coalitions $\s \in \pow(\mathcal{M})$. \vspace{-2ex}
}}
\label{tab:SimStudyHighDimPart1}
\end{table}

The results of the experiments are shown in \Cref{tab:SimStudyHighDimPart1}, where the best performing method is indicated with bold font for each experiment and evaluation criterion. The \texttt{true} approach is only included for EC3, as it is the reference value for the other methods in EC1 and EC2. As previously mentioned, we consider EC1 to be the most informative criterion. We see that \texttt{VAEAC} is the overall most accurate method with respect to all criteria. It obtains the lowest EC1 in all experiments, and often with a significant margin, while EC2 and EC3 indicates that the \texttt{Gaussian} approach is marginally better when $M = 10$ and $N_\text{train} \le 1000$. 

In general, most of the methods become more accurate when we increase $N_\text{train}$, except the \texttt{independence} approach which becomes less precise. This tendency is most evident for the \texttt{empirical} method, but it is also clear for the \texttt{ctree} and \texttt{VAEAC} approaches. The \texttt{Gaussian} and (Gaussian) \texttt{copula} methods do not benefit from increasing $N_\text{train}$ from $1000$ to $5000$. That is, the estimates of the parameters in these methods are stable, and the approaches cannot achieve better evaluation scores as they are based on an incorrect parametric assumption about the data distribution.

The \texttt{VAEAC} model used in these simulations consists of $18724$ parameters, see \Cref{app:subsec:VAEAC_implementaion_hyper}, which might sound enormous considering the small training set. In contrast, the \texttt{Gaussian} and \texttt{copula} models have both $M(M + 3)/2$ parameters each, while the \texttt{independence} approach has zero parameters as it randomly samples the training data. However, recall from \Cref{sec:VAEAC} that each observation can be masked in $2^M$ different ways, hence, \texttt{VAEAC}'s parameters are fitted based on $2^M\times N_\text{train}$ possible different observations. 

There are some inconsistency between the evaluation criteria. In the $M = 10$ and $N_\text{train} = 1000$ setting, the EC3 indicates that the \texttt{Gaussian} approach is the best at estimating the contribution functions, with the \texttt{VAEAC} method a close second. Furthermore, they are both close to the optimal value. However, when considering the EC1, the \texttt{Gaussian} method yields $33\%$ less accurate Shapley value estimates than the \texttt{VAEAC} approach, with respect to the mean absolute error in EC1. The inconsistency between EC1 and EC3 can be linked to two things. First, the EC3 evaluates the average precision of the contribution functions, while the EC1 measures the precision of the Shapley values. Second, the estimates of $v(\s)$ can overshoot for some coalitions and undershoot for others, and such errors may cancel each other out in the Shapley value formula. Overall, there seems to be a moderate coherency between the criteria, at least in the ranking of the approaches. For real-world data, only the EC3 can be used as the true quantities are unknown. However, one should bear in mind that this criterion can only be used to rank the different approaches and not assess the precision of the estimated Shapley values.

\subsubsection{Continuous Data: High-Dimensional}
\label{subsection:simstudy:cont:highimensional}

\begin{table}[!t]
\small
\centering
\begin{adjustbox}{max width=1\textwidth}
\begin{tabular}{@{}cccccccccccc@{}}
\toprule
&&&&\multicolumn{8}{c}{Methods}  \\
\cmidrule{5-12}
$M$ & $N_\s$ & $N_\text{train}$ & Crit & \texttt{True} & \texttt{Indep}. & \texttt{Empir}. & \texttt{Gauss}. & \texttt{Copula} & \texttt{Ctree} & \texttt{VAEAC} & $\texttt{VAEAC}_\mathcal{C}$ \\
\midrule
&&\multirow{3}{*}{100} &  EC1 &  --- & 0.0796 & 0.0626 & 0.0523 & 0.0536 & 0.0777 & \B0.0398 & 0.0419\\
&&& EC2 & --- &   0.4991 & 0.3124 & \B0.0643 & 0.0718 & 0.3519 & 0.1001 & 0.0663\\
&&&  EC3 & 0.2336 & 0.5947 & 0.4692 & 0.2841 & 0.2913 & 0.4966 & 0.3077 & \B0.2835\\
\cmidrule{3-12}
&&\multirow{3}{*}{1000} &  EC1 &  --- & 0.0981 & 0.0630 & 0.0368 & 0.0384 & 0.0556 & \B0.0321 & 0.0346 \\
25&1000&& EC2 & --- &  0.7402 & 0.3010 & \B0.0468 & 0.0570 & 0.2212 & 0.0800 & 0.0496\\
&&&  EC3 & 0.3592 & 0.9022 & 0.5829 & \B0.3880 & 0.4022 & 0.5247 & 0.4164 & 0.3966\\
\cmidrule{3-12}
&&\multirow{3}{*}{5000} &  EC1 &  --- & 0.1087 & 0.0667 & 0.0357 & 0.0373 & --- & \B0.0272 & 0.0297\\
&&& EC2 & --- &  0.8724 & 0.3101 & 0.0489 & 0.0580 & --- & 0.0600 & \B0.0365\\
&&&  EC3 & 0.4459 & 1.0775 & 0.6834 & 0.4755 & 0.4886 & --- & 0.4903 & \B0.4732\\
\midrule
\multirow{6}{*}[-0.5\dimexpr \aboverulesep + \belowrulesep + \cmidrulewidth]{50} & \multirow{6}{*}[-0.5\dimexpr \aboverulesep + \belowrulesep + \cmidrulewidth]{1000} & \multirow{3}{*}{5000} &  EC1 & --- & 0.1669 & 0.1289 & \B0.0458 & 0.0493 & --- & 0.0531 & 0.0484 \\
&&& EC2 & --- &  3.4396 & 2.1031 & \B0.1296 & 0.1774 & --- & 0.4643 & 0.1644\\
&&&  EC3 &  0.7974 & 3.7147 & 2.5857 & \B0.9087 & 0.9585 & --- & 1.2041 & 0.9396 \\
\cmidrule{3-12}
&&\multirow{3}{*}{10000} &  EC1 &  --- & 0.1710 & 0.1308 & 0.0452 & 0.0495 & --- & 0.0510 & \B0.0444 \\
&&& EC2 & --- & 3.4776 & 2.0407 & 0.1261 & 0.1782 & --- & 0.4281 & \B0.1203\\
&&&  EC3 & 0.8362 & 3.8901 & 2.6502 & 0.9432 & 1.0033 & --- & 1.2185 & \B0.9407\\
\midrule
\multirow{6}{*}[-0.5\dimexpr \aboverulesep + \belowrulesep + \cmidrulewidth]{100} & \multirow{6}{*}[-0.5\dimexpr \aboverulesep + \belowrulesep + \cmidrulewidth]{1000}& \multirow{3}{*}{5000} &  EC1 &  --- & 0.2436 & 0.2281 & \B0.0555 & 0.0566 & --- &  0.1024 & 0.0594\\
&&& EC2 & ---  & 6.4815 & 5.8814 & \B0.2337 & 0.2536 & --- & 1.4657 & 0.2823\\
&&&  EC3 &  0.9556 & 6.6358 & 6.0560 & \B1.1604 & 1.1662 & --- &  2.2270 & 1.2092\\
\cmidrule{3-12}
&&\multirow{3}{*}{10000}&  EC1 &  --- & 0.2479 & 0.2359 & 0.0558 & 0.0571 & --- &  0.1013 & \B0.0504  \\
&&& EC2 & --- & 6.7383 & 6.1740 & 0.2428 & 0.2653 & --- &  1.4989 & \B0.1916\\
&&&  EC3 & 1.0333 & 6.9940 & 6.4447 & 1.2458 & 1.2606 & --- &  2.3644 & \B1.2097\\
\midrule
\multirow{6}{*}[-0.5\dimexpr \aboverulesep + \belowrulesep + \cmidrulewidth]{250} & \multirow{6}{*}[-0.5\dimexpr \aboverulesep + \belowrulesep + \cmidrulewidth]{1000} & \multirow{3}{*}{10000} &  EC1 &  --- & \phantom{0}0.8597 & \phantom{0}0.9286 & 0.1612 & \B0.1467 & --- & \phantom{0}0.5175 & 0.1842 \\
&&& EC2 & --- & 36.4480 & 43.3528 & 1.1056 & \B0.9021 & --- & 11.9938 & 1.6597 \\
&&&  EC3 & 2.5101 & 37.8259 & 44.4095 & 3.6158 & \B3.3988 & --- & 14.1812 & 4.1448 \\
\cmidrule{3-12}
&&\multirow{3}{*}{25000} &  EC1 &  --- &  \phantom{0}0.8774 & \phantom{0}1.0091 & 0.1658 & 0.1544 & --- & \phantom{0}0.6891 & \B0.1474\\
&&& EC2 & --- & 38.2186 & 48.2050 & 1.2072 & 1.0198 & --- & 21.4286 & \B0.9415\\
&&&  EC3 & 2.8566 & 39.8862 & 49.4167 & 4.1068 & 3.8659 & --- & 23.8203 & \B3.7863\\
\bottomrule
\end{tabular}
\end{adjustbox}
\caption{{\small The average evaluation criteria over the $R=10$ repetitions of the high-dimensional continuous data simulation studies in \Cref{subsection:simstudy:cont:highimensional}, based on the coalitions $\s \in \mathcal{C}$. 
}}
\label{tab:SimStudyHighDimPart2}
\end{table}

We now let the number of features to be $M \in \set{25, 50, 100, 250}$. In these high-dimensional settings, we sample $N_\mathcal{S} = 1000$ coalitions from $\pow(\M)$ since it is computationally intractable to consider all $2^M$ coalitions, as described in \Cref{subsec:simulations:Evaluation_method}. Furthermore, we decrease $R$ from $20$ to $10$ due to increased computational times, see \Cref{subsection:simstudy:cont:ComputationTimes}.

The results of the high-dimensional simulation study is presented in \Cref{tab:SimStudyHighDimPart2}, where there are several noteworthy findings. The \texttt{ctree} approach is not applied in the higher dimensions due to increased memory and time consumption. The \texttt{VAEAC} method works well for $M=25$, and is the best approach with regard to EC1 for all training sizes. As expected, when $M \ge 50$, we see that using our $\texttt{VAEAC}_\mathcal{C}$ approach, which focuses on the relevant coalitions, greatly increases the performance of the \texttt{VAEAC} methodology. Furthermore, we see that $\texttt{VAEAC}_\mathcal{C}$ is the overall best approach when $N_\text{train}$ is sufficiently large. For limited training sizes, the \texttt{Gaussian} and \texttt{copula} approaches outperform the $\texttt{VAEAC}_\mathcal{C}$ method. We have used different values of $N_\text{train}$ in the different dimensions to illustrate the threshold where $\texttt{VAEAC}_\mathcal{C}$ consistently outperforms the other methods. Finally, for $M \in \set{50, 100, 250}$ all three evaluation criteria yield the same ranking of the approaches. 

We speculate that the performance of our $\texttt{VAEAC}_\mathcal{C}$ method could be improved by proper hyperparameter tuning. The same can be said about the \texttt{empirical} and \texttt{ctree} approaches. We have used the same network architecture for all $M$, and it might be that, for example, the $d=8$-dimensional latent space is insufficient for $M=250$. However, proper hyperparameter tuning of the \texttt{VAEAC} methods is computationally expensive and left to further work.

\subsection{Simulation Study: Mixed Data}
\label{subsection:simstudy:mixed}

The second type of simulation study concerns mixed data following the setup of \textcite{redelmeier:2020}, to which we refer for technical details on the data generating process. The essentials are presented in \Cref{Appendix:sub:GenerateDataMixed}. 

\subsubsection{Mixed Data: Low-Dimensional}
\label{subsection:simstudy:mixed:low}

In the exact low-dimensional setting, we consider experiments in dimensions $4$ and $6$, as outlined in \Cref{tab:sim_mixed:setup}. Computing the true contribution functions and Shapley values are intractable for larger dimensions due to numerical integration. We investigate several degrees of feature dependencies, and to make the  tables more legible, we only report the EC1 \eqref{eq:MAE}, which we consider to be the most informative criterion.

To generate $M$-dimensional dependent mixed data, we first start by generating $N_\text{train} \in \{100, 1000, 5000\}$ samples from $\mathcal{N}_M(\boldsymbol{0}, \Sigma_{\rho})$, where the covariance matrix $\Sigma_{\rho}$ is the equicorrelation matrix, that is, $1$ on the diagonal and $\rho$ off-diagonal. A large $\rho$ implies high feature dependence. The categorical features are created by categorizing the first $M_\text{cat}$ of the $M$ continuous Gaussian features into $L$ categories at certain cut-off-values, while the remaining $M_\text{cont}$ continuous features are left untouched, see \Cref{tab:sim_mixed:setup}. The response is generated according to
\begin{align}
    \label{eq:yResponseMixedData}
    y
    =
    \alpha + \sum_{j=1}^{M_\text{cat}}\sum_{l=1}^L \beta_{jl}\mathbf{1}(x_{j} = l) 
    + \sum_{j=M_\text{cat}+1}^{M_\text{cat}+M_\text{cont}} \gamma_jx_{j} + \epsilon,
\end{align}
where the coefficients are defined in \Cref{tab:sim_mixed:setup}, $\epsilon \overset{iid}{\sim} \mathcal{N}(0, 1)$, and $\mathbf{1}$ is the indicator function. The reason for the linear form in \eqref{eq:yResponseMixedData}, in contrast to the non-linear form in \eqref{eq:response_high_dim}, is that linearity is necessary for computing the true Shapley values for the mixed data generating process, see \Cref{Appendix:sub:GenerateDataMixed} and \textcite{redelmeier:2020}. We fit a linear regression model of the same form as \eqref{eq:yResponseMixedData}, but without the error term, to act as the predictive model $f$. We randomly sample $N_\text{test} = 500$ test observations from the joint distribution to compute the average EC1 over $R$ repetitions. The results are presented in \Cref{tab:sim_mixed:res}, and we obtain similar results in all experiments.

The \texttt{independence} approach is the best method for independent data ($\rho = 0$) regardless of the number of training observations $N_\text{train}$. For $\rho = 0.1$, the value of $N_\text{train}$ determines which method performs the best. The \texttt{independence} approach is best for small training sets ($N_\text{train} = 100$), \texttt{ctree} performs the best for moderate training sets ($N_\text{train} = 1000$), and \texttt{VAEAC} is the best approach for large training sets ($N_\text{train} = 5000$). For moderate to strong dependence ($\rho \ge 0.3$) \texttt{VAEAC} is generally the best approach and often with a distinct margin. The \texttt{independence} approach performs increasingly worse for increasing dependence, which demonstrates the necessity of methods that can handle dependent data.

\begin{table}[!t]
\centering
\small
\begin{adjustbox}{max width=0.975\textwidth}
\begin{tabular}{@{}ccccccccc@{}}
\toprule
$M$ & $M_\text{cont}$ & $M_\text{cat}$ & $L$ & Categorical cut-off values  & $\boldsymbol{\beta}_{1}$ & $\boldsymbol{\beta_{2}}$ & $\boldsymbol{\gamma}$ \\
\midrule
4 & 2 & 2 & 4 & $(-\infty,\, -0.5,\, 0,\, 1,\, \infty)$ & 1, 0, -1, 0.5 & 2, 3, -1, -0.5 & 1, -1 \\
6 & 4 & 2 & 3 & $(-\infty,\, \phantom{-0.5,\,\,} 0,\, 1,\, \infty)$  & 1, 0, -1 & 2, 3, -0.5 & 1, -1, 2, -0.25 \\
\bottomrule
\end{tabular}
\end{adjustbox}
\caption{{\small Table of the number of continuous and categorical features, along with the number of categories $L$ and the associated cut-off values for the simulation studies in \Cref{subsection:simstudy:mixed:low}. The last three columns display the true model parameters of \eqref{eq:yResponseMixedData}, while $\alpha = 1$. \vspace{-2ex}
}}
\label{tab:sim_mixed:setup}
\end{table}

\begin{table}[t]
\centering
\small
\begin{adjustbox}{max width=0.975\textwidth}
\begin{tabular}{@{}ccccclcccccc@{}}
\toprule
&&&&&&\multicolumn{6}{c}{$\overline{\text{EC1}}$ for each $\rho$} \\
\cmidrule{7-12}
$M_\text{cont}$ & $M_\text{cat}$ & $L$ & $R$ & $N_\text{train}$ & Method & $0.0$ & $0.1$ & $0.3$ & $0.5$ & $0.8$ & $0.9$ \\
\midrule
& & & & \multirow{3}{*}{100} & \texttt{Indep}. & \B0.0761 & \B0.1310 & 0.2364 & 0.3642 & 0.5717 & 0.6540 \\
& & & & & \texttt{Ctree} & 0.0927 & 0.1412 & \B0.2048 & 0.2442 & 0.2459 & 0.2186 \\
& & & & & \texttt{VAEAC} & 0.1338 & 0.1537 & 0.2135 & \B0.1950 & \B0.1818 & \B0.1949 \\
\cmidrule{5-12}
& & & &\multirow{3}{*}{1000} & \texttt{Indep}. & \B 0.0289 & 0.0811 & 0.2203 & 0.3541 & 0.5615 & 0.6478 \\
$2$ & $2$ & $4$ & $25$ & & \texttt{Ctree} & 0.0320 & \B0.0706 & 0.1050 & 0.1150 & 0.1052 & 0.0954 \\
& & & & & \texttt{VAEAC} & 0.0773 & 0.0770 & \B0.0869 & \B0.0792 & \B0.0667 & \B0.0633 \\
\cmidrule{5-12}
& & & & \multirow{3}{*}{5000} & \texttt{Indep}. & \B0.0240 & 0.0800 & 0.2163 & 0.3505 & 0.5601 & 0.6464 \\
& & & & & \texttt{Ctree} & 0.0402 & 0.0603 & 0.0735 & 0.0764 & 0.0685 & 0.0626 \\
& & & & & \texttt{VAEAC} & 0.0523 & \B0.0545 & \B0.0564 & \B0.0555 & \B0.0613 & \B0.0489 \\
\midrule
& & & & \multirow{3}{*}{100} & \texttt{Indep}. & \B0.0804 & \B0.1180 & 0.2649 & 0.4093 & 0.6289 & 0.7196 \\
& & & & & \texttt{Ctree} & 0.0867 & 0.1327 & 0.2097 & 0.2344 & 0.2084 & \B0.1814 \\ 
& & & & & \texttt{VAEAC} & 0.1606 & 0.1604 & \B0.1605 & \B0.1840 & \B0.1976 & 0.1935\\
\cmidrule{5-12}
& & & &\multirow{3}{*}{1000} & \texttt{Indep}. & \B0.0312 & 0.0924 & 0.2521 & 0.4061 & 0.6382 & 0.7159\\
$4$ & $2$ & $3$ & $10$ & & \texttt{Ctree} & 0.0352 & \B0.0801 & 0.1110 & 0.1176 & 0.1036 & 0.0844 \\
& & & & & \texttt{VAEAC} & 0.0829 & 0.0900 & \B0.0821 & \B0.0800 & \B0.0704 & \B0.0559 \\
\cmidrule{5-12}
& & & & \multirow{3}{*}{5000} & \texttt{Indep}. & \B0.0258 & 0.0915 & 0.2551 & 0.4056 & 0.6341 & 0.7074 \\
& & & & & \texttt{Ctree} & 0.0332 & 0.0619 & 0.0786 & 0.0786 & 0.0676 & 0.0559 \\
& & & & & \texttt{VAEAC} & 0.0546 & \B0.0556 & \B0.0602 & \B0.0553 & \B0.0529 & \B0.0495\\
\bottomrule
\end{tabular}
\end{adjustbox}
\caption{{\small The average EC1 values for the low-dimensional mixed data simulation studies outlined in \Cref{tab:sim_mixed:setup} for various dependencies $\rho$ and based on all coalitions $\s \in \pow(\mathcal{M})$. \vspace{-2ex}
}}
\label{tab:sim_mixed:res}
\end{table}

\subsubsection{Mixed Data: High-Dimensional}
\label{subsection:simstudy:mixed:high}

In the high-dimensional mixed data simulation studies, we extend the data generating process described in the low-dimensional setting. We use the EC3 in \eqref{eq:EC3} to rank the approaches, as it is extremely time-consuming to calculate the EC1 and EC2 even in this simple linear simulation setup, see \Cref{subsection:simstudy:cont:ComputationTimes}.

We still generate the data according to a $\mathcal{N}_M(\boldsymbol{0}, \Sigma_{\rho})$ and discretize $M_\text{cat}$ of these variables at the categorical cut-off values given in \Cref{tab:sim_mixed:setup_high_dim}. We also use the same response function \eqref{eq:yResponseMixedData} and regression model as in \Cref{subsection:simstudy:mixed:low}. However, to extend the setup such that it works for any $M$, we sample the model coefficients in \eqref{eq:yResponseMixedData}, while the intercept remains $\alpha = 1$. We sample $\beta_{jl}$ from $\{-1.0, -0.9, \dots 2.9, 3.0\}$, for $j = 1, \dots, M_\text{cat}$ and $l = 1, \dots, L$, and $\gamma_j$ from $\{-1.0, -0.9, \dots 0.9, 1.0\}$, for $j = 1, \dots, M_\text{cont}$. All sampling is done with replacements and equal probability. We only consider $N_\text{train} = 1000$, as the other training sizes gave almost identical ranking of the methods in \Cref{tab:sim_mixed:res}. The number of test observations and repetitions are $N_\text{test} = 500$ and $R=10$, respectively. Finally, we sample $N_\mathcal{S} = 1000$ coalitions in all settings and use $100$ epochs to train both the \texttt{VAEAC} and $\texttt{VAEAC}_\mathcal{C}$ models.

\begin{table}[!t]
\centering
\small
\begin{adjustbox}{max width=0.975\textwidth}
\begin{tabular}{@{}ccccccccc@{}}
\toprule
$M$ & $M_\text{cont}$ & $M_\text{cat}$ & $L$ & Categorical cut-off values \\
\midrule
10 &  5 &  5 & 3 & $(-\infty,\, \phantom{-1,}\,\, 0,\, 0.5,\, \phantom{1,}\,\, \infty)$           \\
25 & 15 & 10 & 5 & $(-\infty,\, -1,\, 0,\, 0.5,\, 1,\, \infty)$ \\
50 & 25 & 25 & 2 & $(-\infty,\, \phantom{-1,}\,\, \phantom{0,}\,\, 0.5,\, \phantom{1,}\,\, \infty)$                 \\
50 & 25 & 25 & 4 & $(-\infty,\, -1,\, 0,\, 0.5,\, \phantom{1,}\,\, \infty)$    \\
100& 50 & 50 & 2 & $(-\infty,\, \phantom{-1,}\,\,  0,\, \phantom{0.5,}\,\, \phantom{1,}\,\, \infty)$   \\
100& 75 & 25 & 3 & $(-\infty,\, -1,\, \phantom{0,}\,\, \phantom{0.5,}\,\, 1,\, \infty)$        \\
\bottomrule
\end{tabular}
\end{adjustbox}
\caption{{\small Table of the number of continuous and categorical features, along with the number of categories $L$, and the associated cut-off values for the  high-dimensional mixed data simulation studies. \vspace{-3ex}
}}
\label{tab:sim_mixed:setup_high_dim}
\end{table}

\begin{table}[!t]
\centering
\small
\begin{adjustbox}{max width=0.975\textwidth}
\begin{tabular}{@{}ccccclcccccc@{}}
\toprule
&&&&&&\multicolumn{6}{c}{$\overline{\text{EC3}}$ for each $\rho$} \\
\cmidrule{7-12}
$M_\text{cont}$ & $M_\text{cat}$ & $L$ & $R$ & $N_\text{train}$ & Method & $0.0$ & $0.1$ & $0.3$ & $0.5$ & $0.8$ & $0.9$\\
\midrule
\multirow{4}{*}{5} & \multirow{4}{*}{5} & \multirow{4}{*}{3} & \multirow{4}{*}{10} &\multirow{4}{*}{1000} & \texttt{Indep}. & \B1.9834 & 1.9706 & 1.9526 & 1.9940 & 2.0130 & 2.0680 \\
& & & & & \texttt{Ctree} & 1.9853 & 1.9764 & 1.8568 & 1.6723 & 1.1497  & 0.9079 \\
& & & & & \texttt{VAEAC} & 2.0010 & 1.9641 & \B1.7882 & 1.5804 & 1.2869 & 0.9093 \\
& & & & & $\texttt{VAEAC}_\mathcal{C}$ & 2.0032 & \B1.9610 & 1.7893 & \B1.5721 & \B1.2490 & \B0.8865  \\
\midrule
\multirow{4}{*}{15} & \multirow{4}{*}{10} & \multirow{4}{*}{5} & \multirow{4}{*}{10} &\multirow{4}{*}{1000} & \texttt{Indep}. & \B4.0698  & 4.1725 & 4.3139  & 4.4766  & 4.6167 & 4.7836  \\
& & & & & \texttt{Ctree} & 4.0740 & 4.1788 & 4.0014  & 3.7119  & 2.6228 & 2.0605 \\
& & & & & \texttt{VAEAC} &  4.1396 & \B4.1592  & 3.7626  & 3.4643  & 2.5199 & 2.0980 \\
& & & & & $\texttt{VAEAC}_\mathcal{C}$ & 4.1445 & 4.1886 & \B3.7195  & \B3.4047  & \B2.3611 & \B1.8218  \\
\midrule  
\multirow{8}{*}[-0.5\dimexpr \aboverulesep + \belowrulesep + \cmidrulewidth]{25} & \multirow{8}{*}[-0.5\dimexpr \aboverulesep + \belowrulesep + \cmidrulewidth]{25} & \multirow{4}{*}{2} & \multirow{4}{*}{10} &\multirow{4}{*}{1000} & \texttt{Indep}. & \B5.1431 & 5.5418 & 4.9215  & 4.8469 & 4.6949 & 4.6823  \\
& & & & & \texttt{Ctree} &  5.1486 & 5.5480 & 4.5682  & 3.8533 & 2.2973 & 1.6329\\
& & & & & \texttt{VAEAC} & 5.3110 & 5.5235  & 4.4277  & 3.6303 & 2.2686 & 1.6559 \\
& & & & & $\texttt{VAEAC}_\mathcal{C}$ & 5.3121 & \B5.5137 & \B4.2855  & \B3.4714 & \B2.1214 & \B1.3864 \\
\cmidrule{3-12}  
& & \multirow{4}{*}{4} & \multirow{4}{*}{10} &\multirow{4}{*}{1000} & \texttt{Indep}. & \B7.1610 & 7.2926 & 7.6252  & 7.6523  & 7.9985 & 7.9451\\
& & & & & \texttt{Ctree} & 7.1680 & 7.3943 &  7.1921 & 6.3603 & 4.5871 &  3.4432\\
& & & & & \texttt{VAEAC} & 7.3535 & \B7.2221 & 6.8976  & 6.0281  & 4.4857 & 3.4131\\
& & & & & $\texttt{VAEAC}_\mathcal{C}$ & 7.3556 & 7.2569 & \B6.7564  & \B5.7906  & \B4.0719 & \B3.0303\\
\midrule
\multirow{4}{*}{50} & \multirow{4}{*}{50} & \multirow{4}{*}{2} & \multirow{4}{*}{10} &\multirow{4}{*}{1000} & \texttt{Indep}. & \B10.0998 & 10.3732 & 11.0575 & 11.6383 & 12.7133 & 13.2549   \\
& & & & & \texttt{Ctree} & 10.1195 & 10.4232 & \phantom{0}9.8860 & \phantom{0}8.4248 & \phantom{0}5.1261 & \phantom{0}3.7018 \\
& & & & & \texttt{VAEAC} & 10.5142 & 10.3583 & \phantom{0}9.2926 & \phantom{0}8.0861 & \phantom{0}5.0971 & \phantom{0}4.3668 \\
& & & & & $\texttt{VAEAC}_\mathcal{C}$ & 10.5750 & \B10.3556 & \phantom{0}\B8.8561 & \phantom{0}\B7.1282 & \phantom{0}\B4.2543 & \phantom{0}\B2.8744 \\
\midrule
\multirow{4}{*}{75} & \multirow{4}{*}{25} & \multirow{4}{*}{3} & \multirow{4}{*}{10} &\multirow{4}{*}{1000} & \texttt{Indep}. &  \B8.2279 &  8.2268  & 7.9975 & 7.8198 & 7.5932 & 7.7030  \\
& & & & & \texttt{Ctree} &  8.2329 & \B8.2255 & 7.4081 & 6.1508 & 3.6026 & 2.4320  \\
& & & & & \texttt{VAEAC} & 8.6734 & 8.3551 & 7.3716 & 6.3859 & 4.1005 & 3.0407 \\
& & & & & $\texttt{VAEAC}_\mathcal{C}$ & 8.6284 & 8.2754 & \B6.8487 & \B5.4040 & \B2.9895 & \B1.9742  \\
\bottomrule
\end{tabular}
\end{adjustbox}
\caption{{\small 
The average EC3 values for the high-dimensional mixed data simulation studies outlined in \Cref{tab:sim_mixed:setup_high_dim} for various dependencies $\rho$ and based on the coalitions $\s \in \mathcal{C}$. \vspace{-3ex}
}}
\label{tab:sim_mixed:res_high_dim}
\end{table}

The results of the high-dimensional mixed data studies are presented in \Cref{tab:sim_mixed:res_high_dim}, and they tell a very similar story to that of the low-dimensional simulation studies. For independent data ($\rho = 0)$, the \texttt{independence} approach is the best method with \texttt{ctree} a close second, while the \texttt{VAEAC} and $\texttt{VAEAC}_\mathcal{C}$ methods falls somewhat behind. For moderate to strong dependence ($\rho \ge 0.3$),  $\texttt{VAEAC}_\mathcal{C}$ is undeniably the best approach, as it significantly outperforms the other methods in all but one setting where it is marginally beaten by \texttt{VAEAC}. The margin by which the $\texttt{VAEAC}_\mathcal{C}$ outperforms the other methods increases when the dimension and/or the dependence increases. The former follows from $\texttt{VAEAC}_\mathcal{C}$'s simple but effective masking scheme, which enables it to focus on the relevant coalitions in $\mathcal{C}$.

\subsection{Computation Times}
\label{subsection:simstudy:cont:ComputationTimes}

The evaluation of the methods should not be limited to their accuracy, but also include their computational complexity. The CPU times of the different methods will vary significantly depending on operating system and hardware. We ran the simulations on a shared computer server running Red Hat Enterprise Linux 8.5 with two Intel(R) Xeon(R) Gold 6226R CPU @ 2.90GHz (16 cores, 32 threads each) and 768GB DDR4 RAM. The $\texttt{VAEAC}$ and $\texttt{VAEAC}_\mathcal{C}$ approaches are identical, except in the masking scheme, and therefore obtain nearly identical running times. We only report the time of $\texttt{VAEAC}_\mathcal{C}$ and rather include the training time of the model in parenthesis.

\begin{table}
\centering
\small
\begin{adjustbox}{max width=1\textwidth, max height=0.4\textheight}
\begin{tabular}{@{}cccccccccc@{}}
\toprule
&&&\multicolumn{6}{c}{Methods}  \\
\cmidrule{4-9}
$M$ & $N_\s$ & $N_\text{train}$ & \texttt{Indep}. & \texttt{Empir}. & \texttt{Gauss}. & \texttt{Copula} & \texttt{Ctree} & $\texttt{VAEAC}_{\mathcal{C}}$ \\
\midrule
\multirow{3}{*}{5} & \multirow{3}{*}{$2^5$} & 100 & \phantom{00}0.4 & \phantom{00}0.4 & \phantom{0}0.6 & \phantom{00}0.9 & \phantom{00}0.3 &  \phantom{00}1.1 \phantom{00}(0.2) \\
&& 1000 & \phantom{00}1.1 & \phantom{00}1.1 & \phantom{0}0.9 & \phantom{00}1.2 & \phantom{00}0.6 & \phantom{00}2.1 \phantom{00}(1.0) \\
&& 5000 & \phantom{00}1.5 & \phantom{00}1.5 & \phantom{0}1.1 & \phantom{00}1.5 & \phantom{00}1.0 & \phantom{00}4.2 \phantom{00}(2.9) \\ 
\midrule
\multirow{3}{*}{10} & \multirow{3}{*}{$2^{10}$} & 100 & \phantom{0}11.5 & \phantom{0}11.5 & 25.9 & \phantom{0}44.8 & \phantom{0}23.0 & \phantom{0}24.2 \phantom{00}(0.3) \\
&& 1000 & \phantom{0}34.8 & \phantom{0}35.0 & 36.4 & \phantom{0}65.4 &  \phantom{0}45.8 &  \phantom{0}39.7 \phantom{00}(1.3) \\ 
&& 5000 & \phantom{0}43.5 & \phantom{0}44.1 & 49.0 & \phantom{0}88.3 & 121.2 & \phantom{0}39.7 \phantom{00}(4.3) \\ 
\midrule
\multirow{3}{*}{25} & \multirow{3}{*}{1000} & 100 & \phantom{00}8.5 & \phantom{00}8.5 & 17.4 & \phantom{0}38.3 & \phantom{0}18.4 & \phantom{0}28.8  \phantom{00}(2.4)\\
&& 1000 & \phantom{0}27.0 & \phantom{0}27.0 & 21.6 & \phantom{0}45.7 & \phantom{0}57.6 & \phantom{0}41.8 \phantom{0}(11.0)\\ 
&& 5000 & \phantom{0}38.2 & \phantom{0}36.8 & 24.5 & \phantom{0}56.2 & --- & \phantom{0}65.2 \phantom{0}(32.2)\\ 
\midrule

\multirow{2}{*}{50} & \multirow{2}{*}{1000} & 5000 & \phantom{0}65.1 & \phantom{0}62.4 & 32.1 & \phantom{0}76.9 & --- &  101.6  \phantom{0}(54.7) \\
&& 10000  & 122.3 & 121.8 & 69.4 & \phantom{0}96.1 & --- & 165.4 \phantom{0}(78.4)  \\ 
\midrule

\multirow{2}{*}{100} & \multirow{2}{*}{1000} & 5000  & 171.5 & 167.2 & 48.4 & 112.2 & --- & 167.2  \phantom{0}(92.5)\\ 
&& 10000  & 210.1 & 208.7 & 86.0 & 197.0 & --- & 428.3 (326.7)\\ 
\midrule
\multirow{2}{*}{250} & \multirow{2}{*}{1000} & 10000 & 380.0 & 466.6 & 97.8 & 192.1 & --- & 433.2 (311.9) \\ 
&& 25000 & 901.8 & 992.1 & 79.6 & 232.2 & --- & 614.7 (487.8) \\ 
\bottomrule
\end{tabular}
\end{adjustbox}
\caption{{\small 
Average CPU times in minutes needed to compute the Shapley values using the different methods for $N_\text{test} = 100$ observations in the continuous simulation studies in \Cref{subsection:simstudy:cont}.}}
\label{tab:SimStudyHighDimTime}
\end{table}

In \Cref{tab:SimStudyHighDimTime}, we display the average CPU times in minutes needed for each approach to explain $N_\text{test} = 100$ test observations, averaged over the $R$ repetitions, for the continuous data simulation study in \Cref{subsection:simstudy:cont}. We see that the competing methods are generally faster than the $\texttt{VAEAC}_\mathcal{C}$ approach, except for the \texttt{ctree} method for larger training sizes. When $N_\text{train} = 100$, the \texttt{independence} and \texttt{empirical} approaches are the fastest methods. For larger training sizes ($N_\text{train} \ge 1000$) and higher dimensions ($M\ge25$), the \texttt{Gaussian} approach is the fastest method. The CPU times of the \texttt{independence} approach are higher than expected for large $M$. This is probably due to the fact that the \texttt{independence} approach is implemented as a special case of the \texttt{empirical} approach in version $0.2.0$ of the \texttt{shapr}-package. 

When $N_\text{train}$ increases, most methods yield larger computational times. $\texttt{VAEAC}_\mathcal{C}$'s time increase is mainly caused by the training phase and not by the employment phase in the continuous data simulation studies. When we train the NNs in $\texttt{VAEAC}_\mathcal{C}$, we have used a conservative fixed number of epochs, see \Cref{app:subsec:VAEAC_implementaion_hyper}. Thus, when $N_\text{train}$ increases, each epoch takes longer time to run. For example, when $M=250$ and $N_\text{train} = 25000$, the training phase of the $\texttt{VAEAC}_\mathcal{C}$ method constitutes $79\%$ of the total running time. Optimal choice of hyperparameters (epochs, learning rate, NN architecture) and use of early stopping regimes might decrease the training time without loss of accuracy. However, the training time is a fixed time-cost, which would be near negligible if we were to explain, for example, $N_\text{test} = 10^6$ observations instead of only $N_\text{test} = 100$.

In \Cref{tab:SimStudyMixedTime}, we display the average CPU times in minutes needed for each approach to explain $N_\text{test} = 500$ test observations for the mixed data simulation studies in \Cref{subsection:simstudy:mixed}. We see that the \texttt{independence} approach is the fastest method in all settings. However, the $\texttt{VAEAC}_\mathcal{C}$ approach is much faster than the \texttt{ctree} method for $M = M_\text{cat} + M_\text{cont} \ge 10$. As $N_\text{train} = 1000$ in all high-dimensional settings ($M \ge 10$), the increase in CPU time for $\texttt{VAEAC}_\mathcal{C}$ is mainly caused by increased sampling time in the employment phase and not the training time in the training phase, which slowly increases with $M$.

The true Shapley values took $51.93$ minutes to compute in the $M=4$-dimensional experiments and $30.33$ hours to compute in the $M=6$ setting, on average, for one repetition of one value of $\rho$. Thus, in total it took $243.70$ CPU days to compute the true Shapley values needed to compute the results in \Cref{tab:sim_mixed:res}. This illustrates why computing EC1 and EC2 is computationally infeasible for the high-dimensional mixed data simulation studies, hence, why we only used the EC3 criterion to rank the approaches in \Cref{tab:sim_mixed:res_high_dim}.

\begin{table}[t]
\centering
\begin{adjustbox}{max width=1\textwidth}
\begin{tabular}{@{}cccccccc@{}}
\toprule
&&&&&\multicolumn{3}{c}{Methods}  \\
\cmidrule{6-8}
$M_\text{cont}$ & $M_\text{cat}$ & $L$ & $N_\s$ & $N_\text{train}$ & \texttt{Indep}. & \texttt{Ctree} & $\texttt{VAEAC}_\mathcal{C}$ \\
\midrule
\multirow{3}{*}{2} & \multirow{3}{*}{2} & \multirow{3}{*}{4} & \multirow{3}{*}{$2^4$} & $100$ & \phantom{0}0.1 & \phantom{000}0.6 & \phantom{00}1.8 \phantom{0}(0.1)\\
&&&& 1000 & \phantom{0}0.4 & \phantom{000}0.6 & \phantom{00}2.3 \phantom{0}(0.5)\\
&&&& 5000 & \phantom{0}0.5 & \phantom{000}0.7 & \phantom{00}4.1 \phantom{0}(2.4)\\ 
\midrule
\multirow{3}{*}{4} & \multirow{3}{*}{2} & \multirow{3}{*}{3} & \multirow{3}{*}{$2^6$} & $100$ & \phantom{0}0.2 & \phantom{000}2.9 & \phantom{00}2.2 \phantom{0}(0.1)\\
&&&& 1000 & \phantom{0}1.2 & \phantom{000}2.9 & \phantom{00}2.7 \phantom{0}(0.6)\\
&&&& 5000 & \phantom{0}1.6 & \phantom{000}3.0 & \phantom{00}5.1 \phantom{0}(3.0)\\ 
\midrule
5 & 5 & 3 & 1000 & 1000 & \phantom{0}4.0 & \phantom{00}30.8 & \phantom{0}18.9 \phantom{0}(0.9)\\
\midrule
15 & 10 & 5 & 1000 & 1000 & 13.5 & \phantom{0}258.3 & \phantom{0}73.8 \phantom{0}(3.6)\\
\midrule
\multirow{2}{*}[-0.5\dimexpr \aboverulesep + \belowrulesep + \cmidrulewidth]{25} & \multirow{2}{*}[-0.5\dimexpr \aboverulesep + \belowrulesep + \cmidrulewidth]{25} & 2 & 1000 & 1000 & 37.8 & \phantom{0}273.2 & 156.8 \phantom{0}(5.4)\\
\cmidrule{3-8}
&& 4 & 1000 & 1000 & 39.7 & 1326.4 & 141.2 \phantom{0}(6.4)\\
\midrule
50 & 50 & 2 & 1000 & 1000 & 98.8 & 2385.2 & 276.4 (10.1)\\
\midrule
75 & 25 & 3 & 1000 & 1000 & 68.1 & 2453.8 & 248.8 (10.2) \\
\bottomrule
\end{tabular}
\end{adjustbox}
\caption{{\small 
Average CPU times in minutes needed to compute the Shapley values using the different methods for $N_\text{test} = 500$ observations in the mixed data simulation studies in \Cref{subsection:simstudy:mixed}.}}
\label{tab:SimStudyMixedTime}
\end{table}

\section{Application on Real Data Example}
\label{sec:RealDataExample}

In this section, we explain predictions made by a random forest model fitted to the classical Abalone data set with mixed features. The data set originates from a study by the Tasmanian Aquaculture and Fisheries Institute \parencite{nash1994population}. It has been used in several XAI papers \parencite{vilone2020comparative, aas2021explaining, frye_shapley-based_2020} and other machine-learning studies \parencite{sahin2018abalone, smith2018neural, mohammed2020predicting} as it is freely available from the UCI Machine Learning Repository \url{http://archive.ics.uci.edu/ml/datasets/Abalone}.

An abalone is an edible marine snail, and the harvest of abalones is subject to quotas that are partly based on the age distribution. It is a time-consuming task to determine the age of abalones, as one has to cut the shell through its cone, stain it, and manually count the number of rings through a microscope. The goal is therefore to predict the age of abalones based on other easier obtainable physical measurements. The Abalone data set consists of $4177$ samples with $9$ features each: \texttt{Rings} (+1.5 gives the age in years), \texttt{Length} (longest shell measurement), \texttt{Diameter} (perpendicular to length), \texttt{Height} (with meat in shell), \texttt{WholeWeight} (whole abalone), \texttt{ShuckedWeight} (weight of meat), \texttt{VisceraWeight} (gut-weight after bleeding), \texttt{ShellWeight} (after being dried), and \texttt{Sex} (female, infant, male). The distances are given in millimeters and weights in grams. \texttt{Rings} is an integer, \texttt{Sex} is categorical with $3$ categories, and the rest of the features are continuous. Hence, \texttt{VAEAC} is a suitable method for this mixed data set, in addition to the other methods in \Cref{subsection:simstudy:mixed}. 

\begin{figure}[p!]
    \centering
    \includegraphics[width=1\textwidth]{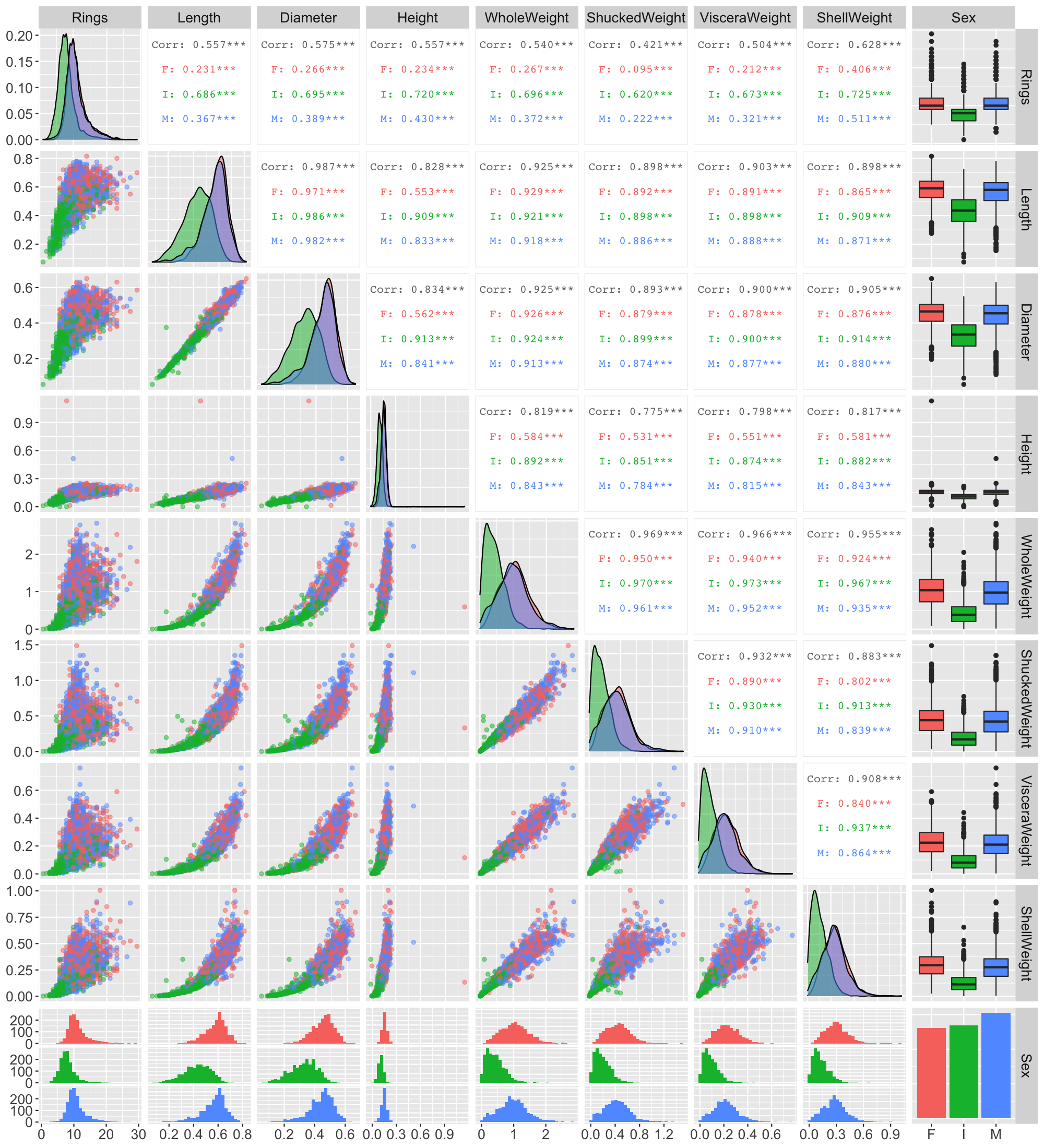}
    \caption{{\small Pairwise scatter plots, marginal density functions, and pairwise Pearson correlation coefficients for the response variable \texttt{Rings} and the features of the Abalone data set. The figure is grouped by \texttt{Sex}, where the females are red, infants are green, and males are blue. The correlations reported in black correspond to all observations, while the colored correlations are grouped based on \texttt{Sex}. The data is highly correlated and display a clear distinction between infants and females/males. 
    }}
    \label{fig:RealData:Pairs}
\end{figure}

An overview of the Abalone data set can be seen in \Cref{fig:RealData:Pairs}, where we show the pairwise scatter plots, marginal density functions, and pairwise Pearson correlation coefficients (for continuous features). There is a clear non-linearity and heteroscedasticity among the pairs of features, and there is significant pairwise correlation between the features. All continuous features have a pairwise correlation above $0.775$, or $0.531$ when grouped by \texttt{Sex}.

\textcite{aas2021explaining} also used the Abalone data set when demonstrating their method for estimating Shapley values. However, their method lacks support for categorical features. Hence, they discarded the categorical feature \texttt{Sex}. We consider this to be a critical methodological limitation, as we will show that \texttt{Sex} is the most important feature for some test observations. This is not surprising, as \Cref{fig:RealData:Pairs} displays a clear distinction between infants and females/males. Hence, it is important to include \texttt{Sex} in the prediction problem.

We treat the \texttt{Rings}/age prediction as a regression problem. We split the data set at random into a test and training set consisting of $100$ and $4077$ samples, respectively. We use a random forest model as it can detect any potential non-linear relationships between the response and the features. The model was fitted to the training data using the \textsc{R}-package \texttt{ranger} \parencite{ranger} with 500 trees and default parameter settings. In a real use case, one would typically not directly use an off-the-shelf ML model, but rather consider, for example, possible feature transformations and conduct model selection and hyper-parameter tuning before using the model. However, as this section aims to illustrate the proposed \texttt{VAEAC} approach and the local Shapley value methodology on a practical example with real-world data, we have taken the liberty to use the default version of random forest in \texttt{ranger}.

\begin{figure}[!t]
    \centering
    \includegraphics[width=1\textwidth]{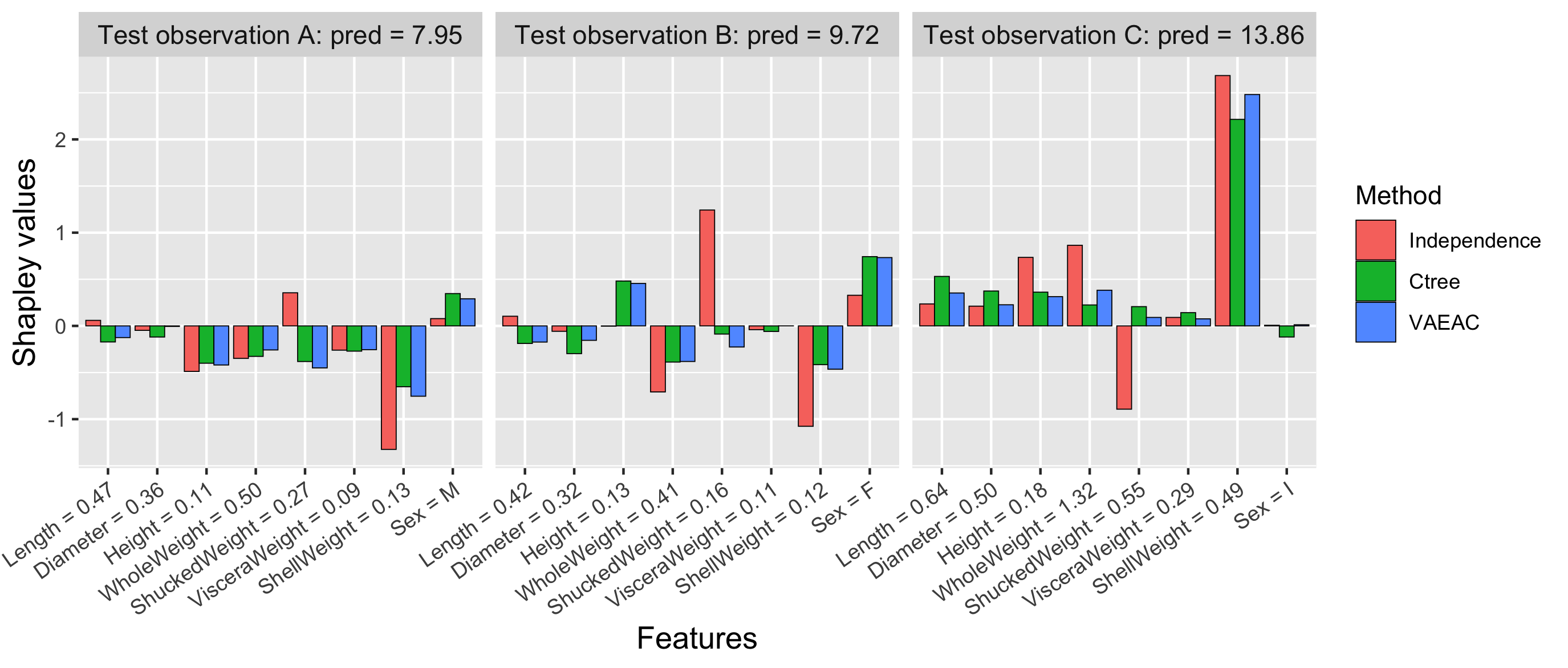}
    \caption{{\small Shapley values for three test observations, where the predicted response is included in the header. The \texttt{VAEAC} and \texttt{ctree} approaches yield similar estimates, while the \texttt{independence} approach often differ in both magnitude and direction. 
    }}
    \label{fig:RealData:ShapleyValues}
\end{figure}

\Cref{fig:RealData:ShapleyValues} shows the estimated local Shapley values for three test observations, based on the \texttt{independence}, \texttt{ctree}, and \texttt{VAEAC} approaches. Recall from \Cref{subsec:ShapleyValuesExplainability} that the Shapley values for a test observation $\boldsymbol{x}^*$ explain the difference between the prediction $f(\boldsymbol{x}^*)$ and the global average prediction $\phi_0 = \E[f(\x)]$. We estimate $\phi_0$ by the mean of the response in the training set, which is $\phi_0=9.93$ in our training set. The three displayed test observations are chosen such that their predicted responses are, respectively, below, similar to, and above $\phi_0$. Moreover, they have different values of the feature \texttt{Sex}. This entails that observation A has mostly negative Shapley values, while observation B has a mixture of positive and negative Shapley values, and observation C has primarily positive Shapley values.

The Shapley values based on the \texttt{ctree} and \texttt{VAEAC} approaches are comparable. For both, \texttt{Sex} gets the largest absolute Shapley value for test observation B. That is, this categorical feature is regarded to be the most important feature for explaining the prediction for this test observation. The importance of \texttt{Sex} is similar to that of other features for test observation A, while it is negligible for test observation C. For test observations A and C, \texttt{ShellWeight} is the most important feature.

The \texttt{independence} approach often yields distinctly different results than the \texttt{ctree} and \texttt{VAEAC} methods, both in magnitude and direction. We do not trust the Shapley values of the \texttt{independence} approach, as there is strong feature dependence in our data. Based on the poor performance of the \texttt{independence} approach for dependent data in the simulation studies, we strongly believe the associated explanations to be incorrect and that they should be discarded. In the next section, we further justify the use of the \texttt{ctree} and \texttt{VAEAC} approaches.

\subsection{Evaluation}

For all approaches treated in this paper, the Shapley value in \eqref{eq:ShapleyValuesDef} is a weighted sum of differences $v(\mathcal{S} \cup \{j\}) - v(\mathcal{S})$ over possible subsets $\mathcal{S}$. However, the approaches differ in how $v(\mathcal{S})$, or more specifically, the conditional distribution $p(\x_{\thickbar{\mathcal{S}}} | \xs = \x^*_{\mathcal{S}})$, is estimated. We now illustrate that the conditional samples generated using the \texttt{VAEAC} and \texttt{ctree} approaches are more representative than the samples generated using the \texttt{independence} method. Thus, it is likely that the estimated Shapley values using the former two methods also are more accurate for the Abalone example.

\begin{figure}[!t]
    \centering
    \includegraphics[width=1\textwidth]{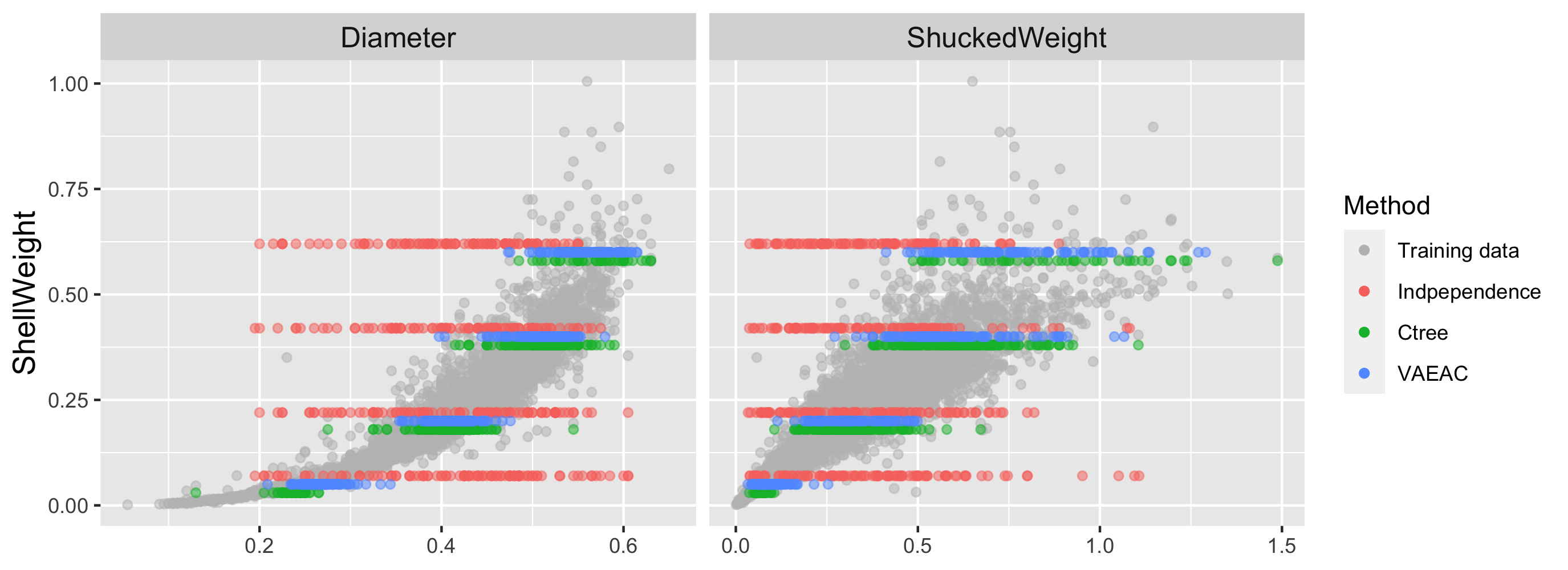}
    \caption{{\small Sampling from the estimated conditional distributions using the \texttt{independence} (red), \texttt{ctree} (green), and \texttt{VAEAC} (blue) approaches. The gray dots are the training data. The samples are generated conditioned on \texttt{ShellWeight} $ \in \{0.05, 0.2, 0.4, 0.6\}$. Note that the red and green dots have been slightly displaced vertically to improve visibility of the figure. The \texttt{ctree} and \texttt{VAEAC} approaches yield on-distribution data, while the \texttt{independence} method generates samples off-distribution.
    }}
    \label{fig:RealData:CondDistSamples}
\end{figure}

Since there are $2^M = 2^8 = 256$ conditional distributions involved in the Shapley formula, it is impossible to show all here. However, we have included some examples to illustrate that the \texttt{VAEAC} and \texttt{ctree} approaches give more correct approximations to the true conditional distributions than the \texttt{independence} approach. \Cref{fig:RealData:CondDistSamples} shows pairplots of \texttt{ShellWeight} against both \texttt{Diameter} and \texttt{ShuckedWeight}, where the grey dots are the training data. The red, green, and blue dots are generated samples from the estimated conditional distribution of the variable on the x-axis given $\texttt{ShellWeight} \in \{0.05, 0.2, 0.4, 0.6\}$, using the \texttt{independence}, \texttt{ctree}, and \texttt{VAEAC} approaches, respectively. 

The \texttt{independence} approach generates unrealistic samples far outside the range of observed values in the training data, in contrast to the \texttt{VAEAC} and \texttt{ctree} methods. It is well known that evaluation of predictive machine learning models far from the domain at which they have been trained, can lead to spurious predictions \parencite{nguyen2015deep, goodfellow2014explaining}, which is related to the \textit{garbage-in-garbage-out} problem. Thus, it is important that the explanation methods are evaluating the predictive model at appropriate feature combinations. 

We can also use the EC3, defined in \eqref{eq:EC3}, to rank the approaches. Recall that the EC3 can be used when the true data generating process is unknown. However, we can only rank the approaches and not state how close they are the optimal approach. Based on the discussion above and \Cref{fig:RealData:CondDistSamples}, it is natural to expect the \texttt{independence} approach to obtain a high value relative to the \texttt{ctree} and \texttt{VAEAC} approaches, which should obtain similar low values. The \texttt{independence} approach produces a value of $3.57$, while the \texttt{ctree} and \texttt{VAEAC} methods obtain $1.27$ and $1.18$, respectively. Thus, according to the EC3, the most accurate approach is \texttt{VAEAC}.

\section{Summary and Conclusion}
\label{sec:Conclusion}

Shapley values originated in cooperative game theory as a solution concept of how to fairly divide the payout of a game onto the players and have a solid theoretical foundation therein. It gained momentum as a model-agnostic explanation framework for explaining individual predictions following the work of \textcite{lundberg2017unified}. However, their work relied on the assumption of independent features. Thus, their methodology may lead to incorrect explanations when the features are dependent, which is often the case for real-world data. There have been several proposed methods for appropriately modeling the dependence between the features, but this is computationally expensive as the number of feature combinations to model grows exponentially with the number of features.

We introduce the variational autoencoder with arbitrary conditioning \texttt{VAEAC} \parencite{ivanov_variational_2018} as a tool to estimate the conditional expectations in the Shapley value explanation framework. The \texttt{VAEAC} approach uses a single variational autoencoder to simultaneously estimate the dependence structure between all feature combinations. This is in contrast to the state-of-the-art dependence-aware methods available in the \textsc{R}-package \texttt{shapr} \parencite{shapr}, which fits a new model to each feature combination. Furthermore, \texttt{VAEAC} supports a mixture of continuous and categorical features. The \texttt{ctree} approach \parencite{redelmeier:2020} is the only other method that directly supports dependent mixed features. The other approaches could be extended to support categorical features by using one-hot encoding, but this is in general infeasible as it drastically increases the computational complexity \parencite[q.v.\ Table.\ 4 \& 6]{redelmeier:2020}, in particular for categorical features with many categories.

Through a series of low-dimensional simulation studies, we demonstrated that our \texttt{VAEAC} approach to Shapley value estimation outperforms or matches the state-of-the-art dependence-aware approaches, even for very small training sets. In the high-dimensional simulation studies, we used efficient sampling of coalitions and evaluated the approaches based on the sampled coalitions. Evaluating all coalitions is infeasible in high dimensions as the number of feature dependencies to estimate grows exponentially with the number of features. We introduced our $\texttt{VAEAC}_\mathcal{C}$ approach, which employs a simple but effective non-uniform masking scheme that allows the $\texttt{VAEAC}_\mathcal{C}$ approach to focus on the relevant coalitions. This significantly improved the accuracy of the \texttt{VAEAC} methodology in the high-dimensional simulation studies, as our novel $\texttt{VAEAC}_\mathcal{C}$ approach outperformed all other methods by a significant margin when trained on sufficiently large training sets. Furthermore, our $\texttt{VAEAC}_\mathcal{C}$ approach was also computationally competitive with the other methods. Further work on proper hyper-parameter tuning of the $\texttt{VAEAC}_\mathcal{C}$ method might decrease computation times.

Another possible application of masking schemes is in the \textit{asymmetric Shapley value} framework of \textcite{frye2020asymmetric}, a special case of the methodology of \textcite{heskes2020causal}. The asymmetric Shapley value incorporates prior knowledge of the data's casual structure (should such be known) into the explanations of the model's predictions. In their framework, for \textit{distal} causes, it is natural to consider a masking scheme that solely focuses on coalitions where the \textit{ancestor} features are known, and the \textit{descendant} features are unknown. For \textit{proximate} causes, the reverse masking scheme would be suitable.

The \texttt{VAEAC} approach was also used to explain predictions made by a random forest model designed to predict the age of an abalone (marine snail). In this case, the true Shapley values are unknown, but we saw a strong coherency between the explanations from the \texttt{VAEAC} and \texttt{ctree} approaches. The \texttt{independence} approach gave different Shapley values both in direction and magnitude. We provided results that indicate that the former two approaches provide more sensible Shapley value estimates than the latter. Furthermore, our \texttt{VAEAC} approach was the best method according to the evaluation criterion. 

The focus of this article has been on modeling the feature dependencies in the \textit{local} Shapley value framework. However, the used approaches can also be used to model feature dependencies in \textit{global} Shapley value frameworks. For example, both \textcite{covert2020understanding} and \textcite{anonymous2022joint} currently assume feature independence when estimating their global Shapley values. Thus, their frameworks would benefit from using, e.g., our \texttt{VAEAC} approach.

In the current implementation of \texttt{VAEAC}, we use the Gaussian distribution to model the continuous features. This can generate inappropriate values if the feature values are, for example, strictly positive or restricted to an interval $(a,b)$. In such settings \texttt{VAEAC} can generate negative samples or samples outside the interval, respectively. A naive approach is to set the negative values to zero and the values outside the interval to the nearest endpoint value. However, a better approach would be to transform the data to an unbounded form before using the \texttt{VAEAC} approach. That is, strictly positive values can be log-transformed, while interval values can first be mapped to $(0,1)$ and then further to the full real line by, for example, the logit function. A more complicated but undoubtedly interesting approach is to replace the Gaussian distribution in the decoder with a distribution that has support on the same range as the features, like the gamma distribution for strictly positive values and the beta distribution for values in $(0,1)$. 

The \texttt{VAEAC} approach currently supports incomplete data in the training phase, but the methodology can also be extended to support incomplete data in the employment phase. This could be done by including a missing feature mask in the masked encoder, similar to that we explained for the full encoder in \Cref{subseq:MissingFeatures}. This would allow the \texttt{VAEAC} method to condition on missing values. That is, in the employment phase, the masked encoder would receive the observed features $\xs$ of the test sample $\x$, with missing values set to zero, the unobserved feature mask $\sbb$, and the missing feature mask indicating the missing features. The missing feature mask is needed to distinguish between actual zeros in $\xs$ and zeros induced by us setting missing features to zero. We discuss a small example in \Cref{app:subsec:IncompleteDataEmplymentPhase}.

\subsection*{Acknowledgments}
This work was supported by The Norwegian Research Council 237718 through the Big Insight Center for research-driven innovation. We want to thank Annabelle Redelmeier for her advice on setting up the low-dimensional mixed data simulation study. Furthermore, we are very grateful to the anonymous reviewers for their suggestions on improving the presentation of this work.

\newpage
\appendix
\section*{Appendix}
\label{Appendix}

We start the appendix by giving a brief description of the alternative approaches to \texttt{VAEAC} in \Cref{Appendix:AlternativeApproaches}. In \Cref{Appendix:VLB}, we give the full derivation of the variational lower bound. The implementation details of \texttt{VAEAC} are discussed in \Cref{Appendix:Architecture}. In \Cref{subsection:simstudy:cat}, we conduct a simulation study with only categorical features. The data generating process of the different simulation studies are elaborated in \Cref{Appendix:GenerateData}. In \Cref{Appendix:sec:NumMonteCarlo}, we experiment with different number of Monte Carlo samples on the MAE. Finally, \Cref{Appendix:sec:UniversalApproximator} illustrates \texttt{VAEAC} as a universal approximator on the Abalone data set.

\section{Alternative Approaches}
\label{Appendix:AlternativeApproaches}

In this section, we give a short description of the other approaches available in the \textsc{R}-package \texttt{shapr} \parencite{shapr}, that is, the \texttt{independence}, \texttt{empirical}, \texttt{Gaussian}, \texttt{copula}, and \texttt{ctree} approaches. We use the default hyperparameters implemented in the \texttt{shapr}-package when using these methods. 

\begin{table}[ht!]\centering
\begin{tabular}{p{2.45cm}p{4.16cm}p{6.9cm}}
Method & Citation & Description \\
\toprule
\texttt{Independence} & \textcite{lundberg2017unified} & Assume the features are independent. Estimate \eqref{eq:ContributionFunc} by
\eqref{eq:KerSHAPConditionalFunction} where $\boldsymbol{x}_{\thickbar{\mathcal{S}}}^{(k)}$ are sub-samples from the training data. \\
\midrule
\texttt{Empirical} & \textcite{aas2019explaining} &  Calculate the Mahalanobis distance between the observation being explained and every training instance. Use this distance to calculate a weight for each training instance using $\eta = 1$. Approximate \eqref{eq:ContributionFunc} using a function of these weights. \\
\midrule
\texttt{Gaussian} & \textcite{aas2019explaining} & Assume the features are jointly Gaussian. Sample $K$ times from the corresponding conditional distribution. Estimate \eqref{eq:ContributionFunc} with \eqref{eq:KerSHAPConditionalFunction} using these samples. \\
\midrule
\texttt{Copula} & \textcite{aas2019explaining} &  Assume the dependence structure of the features can
be approximated by a Gaussian copula. Sample $K$ times from the corresponding conditional distribution. Estimate \eqref{eq:ContributionFunc} with \eqref{eq:KerSHAPConditionalFunction} using these samples. \\
\midrule
\texttt{Ctree} & \textcite{redelmeier:2020} & Fit conditional inference trees for each coalition. Sample $K$ times from the corresponding conditional distribution (with replacements). Estimate \eqref{eq:ContributionFunc} with \eqref{eq:KerSHAPConditionalFunction} using these samples.  \\
\bottomrule
\end{tabular}
\caption{{\small A short description of the approaches used to estimate \eqref{eq:ContributionFunc} in the simulation studies.}}
\label{tabel:summaryApproaches}
\end{table}

\section{Variational Lower Bound for \texttt{VAEAC}}
\label{Appendix:VLB}

Here we give the full derivation of the variational lower bound in \eqref{eq:VLBVAEAC}:
\begin{align*}
    \begin{split}
    \log p_{\bpsi, \btheta} (\xsb|\xs, \sbb)
    &=
    \int p_\bphi(\z|\x,\sbb) \log p_{\bpsi, \btheta}(\xsb|\xs, \sbb) \diff \z \\
    &=
    \E_{p_\bphi(\z|\x,\sbb)} \sqbr{\log p_{\bpsi, \btheta}(\xsb|\xs, \sbb)} \\
    &= 
    \E_{p_\bphi(\z|\x,\sbb)} \bigg[\log  \frac{p_{\bpsi, \btheta}(\xsb, \z|\xs, \sbb)}{p_{\bpsi, \btheta}(\z|\underbrace{\xs, \xsb}_{\x}, \sbb)}\bigg] \\
    &= 
    \E_{p_\bphi(\z|\x,\sbb)} \sqbr{\log \set{\frac{p_{\bpsi, \btheta}(\xsb, \z|\xs, \sbb)}{p_\bphi(\z|\x,\sbb)} \frac{p_\bphi(\z|\x,\sbb)}{p_{\bpsi, \btheta}(\z|\x, \sbb)}}} \\
    &= 
    \E_{p_\bphi(\z|\x,\sbb)} \sqbr{\log \frac{p_{\bpsi, \btheta}(\xsb, \z|\xs, \sbb)}{p_\bphi(\z|\x,\sbb)}} 
    + 
    \E_{p_\bphi(\z|\x,\sbb)} \sqbr{\log \frac{p_\bphi(\z|\x,\sbb)}{p_{\bpsi, \btheta}(\z|\x, \sbb)}} \\
    &= 
    \E_{p_\bphi(\z|\x,\sbb)} \sqbr{\log \frac{p_{\bpsi, \btheta}(\xsb, \z|\xs, \sbb)}{p_\bphi(\z|\x,\sbb)}} 
    + 
    \underbrace{D_\text{KL}\big( p_\bphi(\z|\x,\sbb) \;\|\; p_{\bpsi, \btheta}(\z|\x, \sbb)\big)}_{\ge 0} \\
    &\ge
    \E_{p_\bphi(\z|\x,\sbb)} \sqbr{\log \frac{p_{\bpsi, \btheta}(\xsb, \z|\xs, \sbb)}{p_\bphi(\z|\x,\sbb)}} \\
    &=
    \E_{p_\bphi(\z|\x,\sbb)} \sqbr{\log \frac{p_{\btheta}(\xsb| \z,\xs, \sbb) p_{\bpsi}(\z|\xs, \sbb)}{p_\bphi(\z|\x,\sbb)}} \\
    &=
    \E_{p_\bphi(\z|\x,\sbb)} \sqbr{\log p_{\btheta}(\xsb| \z,\xs, \sbb)} + 
    \E_{p_\bphi(\z|\x,\sbb)} \sqbr{\log \frac{p_{\bpsi}(\z|\xs, \sbb)}{p_\bphi(\z|\x,\sbb)}} \\
    &=
    \E_{p_\bphi(\z|\x,\sbb)} \sqbr{\log p_{\btheta}(\xsb| \z,\xs, \sbb)} - 
    \E_{p_\bphi(\z|\x,\sbb)} \sqbr{\log \frac{p_\bphi(\z|\x,\sbb)}{p_{\bpsi}(\z|\xs, \sbb)}} \\
    &=
    \E_{p_\bphi(\z|\x,\sbb)} \sqbr{\log p_{\btheta}(\xsb| \z,\xs, \sbb)} - 
    D_\text{KL} \big(p_\bphi(\z|\x,\sbb) \;\|\; p_{\bpsi}(\z|\xs, \sbb)\big) \\
    &= 
    \mathcal{L}_{\texttt{VAEAC}} (\x, \sbb | \btheta, \bpsi, \bphi).
    \end{split}
\end{align*}

\section{Implementation Details of \texttt{VAEAC}}
\label{Appendix:Architecture}
In this section, we describe different implementation details, masking schemes, and possible implementation improvements for the \texttt{VAEAC} approach.

\begin{figure}[!ht]
    \centering
    \includegraphics[width=1\textwidth]{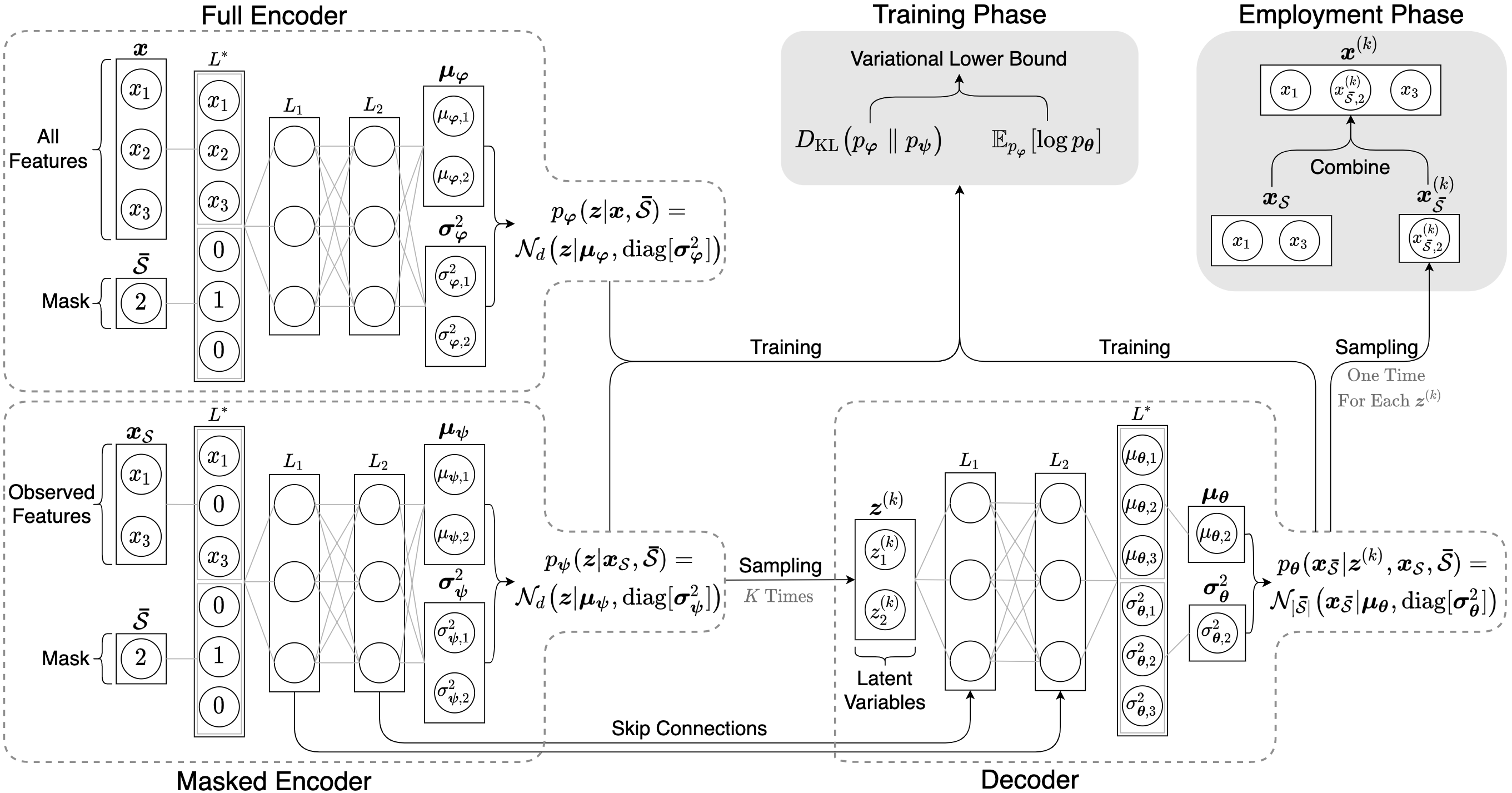}
    \caption{{\small An extended version of \Cref{fig:VAEAC} where we include the fixed-length implementation details introduced in \Cref{app:subsec:VAEAC_implementaion_length}.\vspace{-2ex}}
    }
    \label{fig:VAEAC_appendix}
\end{figure}

\subsection{Fixed-Length Input and Output}
\label{app:subsec:VAEAC_implementaion_length}

In \Cref{sec:VAEAC}, we used a simplified representation to focus the spotlight on the
essential aspects of the \texttt{VAEAC} approach. We now provide the full implementation details.

The \texttt{VAEAC} method is implemented in the auto-grad framework PyTorch \parencite{pytorch}. In the implementation, we need to take into consideration that the number of unobserved features $|\thickbar{\mathcal{S}}|$ varies depending on the coalition $\mathcal{S}$. Hence, the input of the full encoder and masked encoder will be of varying length, which is inconsistent with NNs' assumption of fixed-length input. \Cref{fig:VAEAC_appendix} is an extended version of \Cref{fig:VAEAC} where we introduce a zeroth layer $L^*$ in the encoders which enforces a fixed-length input to the first layer $L_1$ of the NNs. To ensure masks $\thickbar{\mathcal{S}}$ of equal length, we introduce $I(\thickbar{\mathcal{S}}) = \left\{\mathbf{1}(j \in \thickbar{\mathcal{S}}): j = 1, \dots, M\right\} \in \{0, 1\}^M$, which is an $M$-dimensional binary vector where the $j$th element $I(\thickbar{\mathcal{S}})_j$ is $1$ if the $j$th feature is unobserved and $0$ otherwise. For a fixed-length observed feature vector $\xs$, we define $\tilde{\x}_{\mathcal{S}} = \x \circ I(\mathcal{S})$, where $\circ$ is the element-wise product. Similarly, the decoder also has a fixed-length output. Thus, we also introduce a last layer $L^*$ in the decoder which extracts the relevant values of $\boldsymbol{\mu}_\btheta$ and $\boldsymbol{\sigma}_\btheta^2$ given mask $\thickbar{\mathcal{S}}$. 

If we did not provide the full encoder with the mask $\sbb$, it would map the input $\x$ to the same $\boldsymbol{\mu}_\bphi$ and $\boldsymbol{\sigma}_\bphi^2$ for all $\s$. By also providing $\sbb$, the full encoder can find different latent representations of $\x$ for each $\sbb$. The conceptual aim of the full encoder is to guide the masked encoder, in the training phase, to find relevant latent representations of $\xs$ and $\sbb$. Thus, including the mask $\sbb$ in the full encoder will aid this procedure.

We now elaborate the example given in \Cref{subsub:ModelDescriptionVAEAC} with $\x = \{x_1, x_2, x_3\}$, $\mathcal{S} = \{1,3\}$ and $\thickbar{\mathcal{S}} = \{2\}$. The observed and unobserved feature vectors are $\xs = \{x_1, x_3\}$ and $\xsb = \{x_2\}$, respectively. The associated fixed-length versions computed in layer $L^*$ are $I(\thickbar{\mathcal{S}}) = \{0, 1, 0\}$ and $\tilde{\x}_{\mathcal{S}} = \{x_1, x_2, x_3\} \circ \{1, 0, 1\} =  \{x_1, 0, x_3\}$. Note that \texttt{VAEAC} does not have issues with differentiating actual zeros from zeros induced by the fixed-length transformation as it has access to $I(\thickbar{\mathcal{S}})$. We consider these implementation details to be a part of the NN architecture, hence, the tilde-notation and the $I$ function are not included in the mathematical derivations. The final complete sample is therefore $\x^{(k)} = \x_{\thickbar{\mathcal{S}}}^{(k)} \circ I(\sbb) + \x \circ I(\s)$.

\begin{figure}[!t]
    \centering
    \includegraphics[width=1\textwidth]{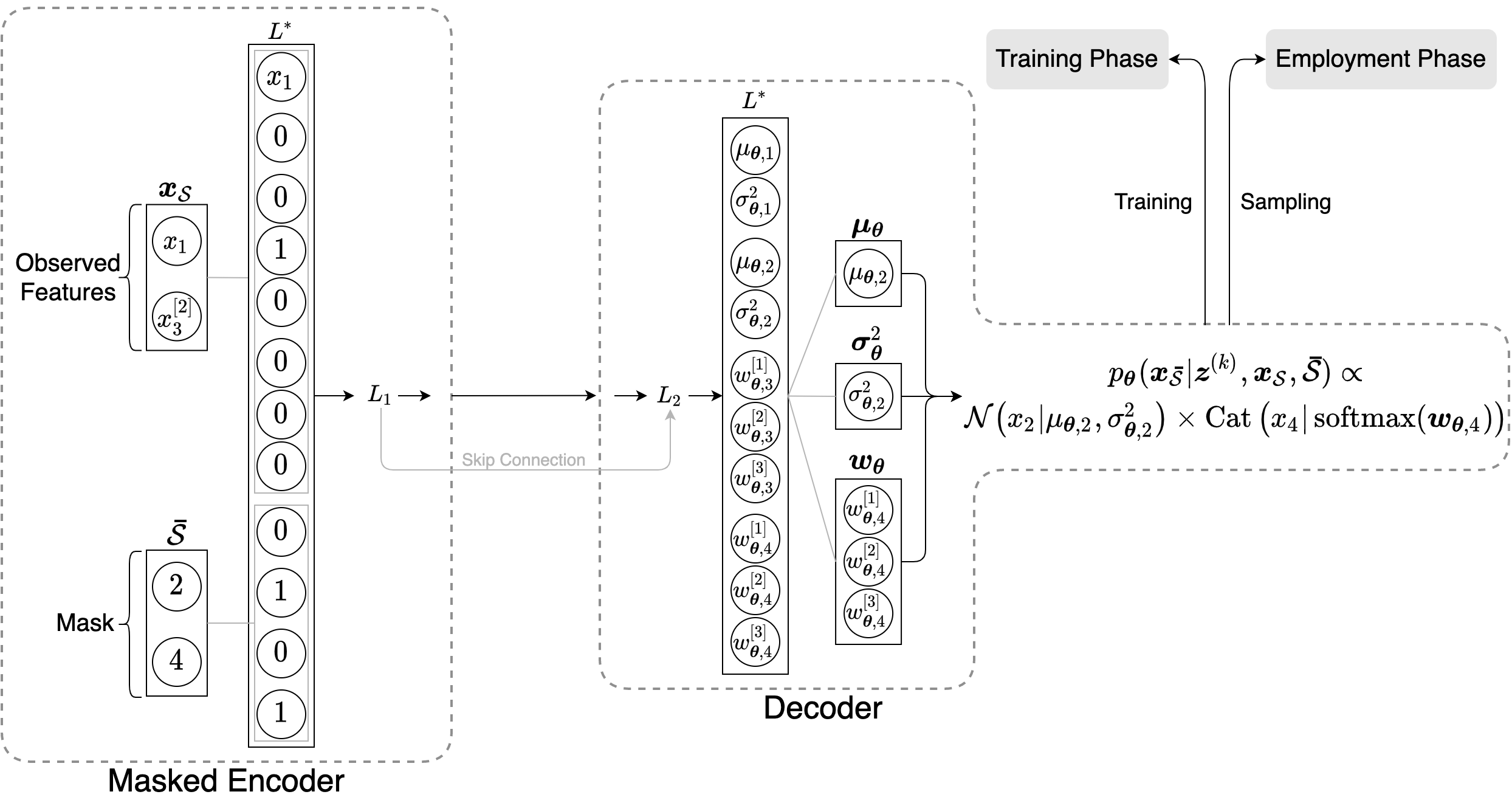}
    \caption{Illustration of \texttt{VAEAC} with mixed features and fixed-length input. Features $x_1$ and $x_2$ are continuous, while $x_3$ and $x_4$ are categorical with $3$ categories each. Here $\xs = \{x_1, x_3\}$ and $\xsb = \{x_2, x_4\}$, i.e., $\sbb = \{2,4\}$. Furthermore, $x_3$ takes on the second category indicated by the superscript in $x_3^{[2]}$. The $\w_\btheta$ are the logits of the different categories. The structure of the full encoder is identical to the masked encoder, except that it receives the full input during the training phase.\vspace{-2ex}}
    \label{fig:VAEAC_cat}
\end{figure}

\Cref{fig:VAEAC_cat} illustrates the \texttt{VAEAC} method's \textit{internal} use of one-hot encoding in an $M=4$-dimensional mixed data setting. Features $x_1$ and $x_2$ are continuous, while $x_3$ and $x_4$ are categorical with $3$ categories each. The figure illustrates the setting where $\sbb = \{2,4\}$, which means that $\xsb = \{x_2, x_4\}$ are the unobserved features to be estimated conditioned on the observed features $\xs = \{x_1, x_3\}$. Let $x_3$ belong to the second category, which we indicate by superscript, that is, $x_3^{[2]}$. Thus, the one-hot encoding of $x_3^{[2]}$ is $\{0,1,0\}$, while $x_4$ becomes $\{0,0,0\}$ in the one-hot fixed-length input notation. The decoder output the estimated means $\boldsymbol{\mu}_\btheta$ and variances $\boldsymbol{\sigma}^2_\btheta$ for the continuous variables and logits $\boldsymbol{w}_\btheta$ for each category of the categorical variables.

\subsection{Hyperparameters}
\label{app:subsec:VAEAC_implementaion_hyper}

Several hyperparameters must be set before training a \texttt{VAEAC} model, both architecturally and training-wise. The NNs of the encoder, masked encoder, and decoder can be of different size, but throughout the paper we use shared parameters and set $\texttt{depth} = 3$, $\texttt{width} = 32$, $d=\texttt{latent\_dim} = 8$ based on experience. This gives a total of $18724$ model parameters, which are tuned using the Adam optimization algorithm \parencite{kingma2014adam} with learning rate $\texttt{lr} = 0.001$. Smaller networks can be used, but we found them to perform sub-optimal for the simulation studies in this paper.

We use the same values for the hyperparameters ${\sigma}_{\mu}$ and ${\sigma}_{\sigma}$ of the prior in latent space as \textcite{ivanov_variational_2018}, that is, $10^4$ for both of them. This value correspond to a very weak, almost disappearing, regularization. That is, the distribution is close to uniform near zero and does not affect the learning process significantly. The activation function used is LeakyReLU, with the default parameters set in the PyTorch implementation. We use $\texttt{batch\_size} = 64$ and generate $K=250$ Monte Carlo samples from each conditional distribution for each test observation. The masking scheme $p(\sbb)$ is set to have equal probability for all coalitions $\sbb$ for \texttt{VAEAC}, while the masking scheme for $\texttt{VAEAC}_\mathcal{C}$ is the sampling frequencies of the coalitions $\s$ in $\mathcal{C}$, see \Cref{subseq:MaskingSchemes}.

In all simulation studies, we let the number of epochs be $\texttt{epochs} = 200$ if $N_\text{train} \le 1000$ and $\texttt{epochs} = 100$ if $N_\text{train} \ge 5000$, unless otherwise specified. These are conservative numbers. We use the \texttt{VAEAC} model at the epoch with the lowest \textit{importance sampling} estimate, denoted by IWAE \parencite{CVAE_Sohn}. The IWAE estimates the log-likelihood by
\begin{align}
    \label{eq:importanceSamplingVAEAC}
    \operatorname{IWAE} 
    = 
    \log \frac{1}{V} \sum_{i=1}^{V} \frac{p_{\boldsymbol{\psi}}(\boldsymbol{z}_i | \boldsymbol{x}_{\mathcal{S}}, \thickbar{\mathcal{S}})p_{\boldsymbol{\theta}}(\boldsymbol{x}_{\thickbar{\mathcal{S}}} | \boldsymbol{z}_i, \boldsymbol{x}_{{\mathcal{S}}}, \thickbar{\mathcal{S}})}{p_{\boldsymbol{\phi}}(\boldsymbol{z}_i | \boldsymbol{x}_{\mathcal{S}}, \thickbar{\mathcal{S}})}
    \approx
    \log p_{\bpsi, \btheta}(\xsb | \xs, \sbb),
\end{align}
where $V$ is the number of IWAE validation samples $\boldsymbol{z}_i \sim p_{\boldsymbol{\phi}}(\boldsymbol{z}_i | \boldsymbol{x}_{\mathcal{S}}, \thickbar{\mathcal{S}})$ generated for each $\xs$. We set $V = 40$. The IWAE is computed on a hold out validation set which is set to be $25\%$ of the training data. That is, if $N_{\text{train}} = 100$, we actually only use $75$ observations to train the \texttt{VAEAC} model and the rest to validate the trained model. In our experience, the VLB and IWAE stabilize long before the last epoch, hence, we have chose conservative values of \texttt{epochs} which could be optimzed to decrease the training time. 

To stabilize the results, we initiate $15$ \texttt{VAEAC} models, let them run for $5$ epochs, and then only continue training the \texttt{VAEAC} with lowest VLB. This is to reduce the effect of the randomly initiated model parameters, which can put \texttt{VAEAC} in a sub-optimal part of the parameter space which it cannot escape. One could also use other types of adaptive learning rates to solve this problem, see \textcite[Ch.\ 8]{Goodfellow-et-al-2016}.

\subsection{Improvements in Implementation}
\label{app:subsec:VAEAC_implementaion_improvments}
The current implementation of \texttt{VAEAC} can be improved in several ways. First, we have used a conservative number of epochs during training, but the IWAE \eqref{eq:importanceSamplingVAEAC} often stabilizes before the reaching the last epoch. Hence, the fitting time of \texttt{VAEAC} could be reduced by applying some type of early stopping regime if no improvement is seen in the IWAE over several epochs. Second, the sampling process of \texttt{VAEAC} is not parallelized. However, as the test observations are independent, the associated conditional samples could be generated simultaneously on separate CPUs or, even better, GPUs. We did not have access to GPUs when we ran the simulations.

\subsection{Incomplete Data in the Employment Phase}
\label{app:subsec:IncompleteDataEmplymentPhase}
The \texttt{VAEAC} method can be extended to also support incomplete data in the employment phase, and we now sketch how this can be done. This was not discussed by \textcite{ivanov_variational_2018}, as conditioning on missing features could not occur in the original application of \texttt{VAEAC}. They used the \texttt{VAEAC} framework to impute missing feature values in a data set conditioned on the observed feature values. While in the Shapley value framework, we consider all feature combinations and can therefore end up in a setting where we need to condition on missing features. 

Assume that we are in the employment phase and consider the same setup as in \Cref{fig:VAEAC_appendix}, but where the third feature is now missing, that is, $x_3 = \texttt{NA}$. The current version of \texttt{VAEAC} would essentially model $p(x_2 | x_1, x_3)$ by $p(x_2 | x_1)$. This naive approach is also applicable for the competing conditional methods in \Cref{sec:SimulationStudy}, but it can lead to incorrect inference when the data is not \textit{missing completely at random} (MCAR). MCAR is a strong assumption to make about real-world incomplete data \parencite{van2018flexible}. A weaker assumption is \textit{missing at random} (MAR), that is, the missingness can be fully accounted for by the observed features. In the MAR setting, knowing that $x_3 = \texttt{NA}$ could be related to the value of $x_2$. Thus, when modelling $x_2$, we need to condition on $x_3 = \texttt{NA}$ as the missingness can contain information. For example, if $x_2$ is often negative when $x_3$ is missing, but positive  when $x_3$ is not missing, then knowing that $x_3 = \texttt{NA}$ should influence the estimation of $x_2$.

We propose to extend the \texttt{VAEAC} framework by also including a missing feature mask $\mathcal{I}$ in the masked encoder, similar to what \textcite{ivanov_variational_2018} did for the full encoder, see \Cref{subseq:MissingFeatures}. The masked encoder would then receive the observed features $\xs = \{x_1, 0\}$ with $x_3 = \texttt{NA}$ set to zero, the unobserved feature mask $\sbb = \{2\}$, and the missing feature mask $\mathcal{I} = \{3\}$. To ensure missing feature masks of equal length, we would apply the same approach we used for the fixed-length unobserved feature masks $\sbb$ above. Thus, the fixed-length input to the masked encoder would be $\{x_1, 0, 0, 0, 1, 0, 0, 0, 1\}$. Here, the first three elements correspond to the fixed-length notation of $\xs$, the next three corresponds to $\sbb$, and the last three elements corresponds to $\mathcal{I}$.

\section{Simulation Study: Categorical Data}
\label{subsection:simstudy:cat}

The categorical simulation study follows the setup of \textcite{redelmeier:2020}. Their results are based on a single run, which can lead to incorrect conclusions as the results vary significantly between each repetition, hence, we conduct repeated simulations. Similarly to the mixed data studies in \Cref{subsection:simstudy:mixed}, we compare the \texttt{VAEAC} approach with the \texttt{independence} and \texttt{ctree} methods.

Let $\{\boldsymbol{x}^{(i)}\}_{i = 1, \dots, N_\text{test}}$ be the set of all unique $M$-dimensional categorical observations where each feature has $L$ categories. Thus, $N_\text{test} = L^M$, which is manageable for small dimensional settings. We replace $N_\text{test}$ in the denominator of EC1 \eqref{eq:MAE} by $p(\boldsymbol{x}^{(i)})$, the corresponding probability mass function, and move it inside the summation. In larger dimensional settings, we use a subset of the $N_\text{test} = 2000$ most likely feature combinations and scale the probabilities such that they sum to $1$ over those combinations.

The categorical data generating process and related Shapley value computations are elaborated in \Cref{Appendix:sub:GenerateDataCategorical}. The main point there is that the categorical data is generated by sampling from a multivariate Gaussian $\mathcal{N}_M(\boldsymbol{\mu}, \Sigma_{\rho})$ before categorizing each of the $M$ variables into $L$ categories at certain cut-off-values, defined in \Cref{tab:sim_cat:setup}. The covariance matrix $\Sigma_{\rho}$, which reflects the feature dependence, is $1$ on the diagonal and $\rho$ off-diagonal. 

We consider six different setups which are described in \Cref{tab:sim_cat:setup}. We use $N_\text{train} = 1000$ training observations to fit the predictive model, while $\rho \in \{0.0, 0.1, 0.3, 0.5, 0.8, 0.9\}$ and  $\boldsymbol{\mu} = \boldsymbol{0}$. The response is generated according to 
\begin{align}
\label{eq:cat_data_response_func}
    y_i = \alpha + \sum_{j=1}^M\sum_{l=1}^L \beta_{jl} \mathbf{1}(x_{ij} = l) + \epsilon_i,
\end{align}
where $x_{ij}$ is the $j$th feature value of the $i$th training observation and $\epsilon_i \sim \mathcal{N}(0, 0.1^2)$, for $i = 1, \dots, N_\text{train}$. The parameters $\alpha$, $\beta_{jl}$ are sampled from $\mathcal{N}(0, 1)$, for $j=1,\dots,M$ and $l=1,\dots,L$. The predictive model $f$ takes the same form as \eqref{eq:cat_data_response_func}, but without the noise term. The model parameters are estimated using standard linear regression. 

The results of the categorical experiments are shown in \Cref{tab:sim_cat:results}. For all dimensions, we see that the \texttt{independence} approach is the best for independent data ($\rho = 0$). However, both \texttt{ctree} and \texttt{VAEAC} significantly outperform \texttt{independence} for moderate to strong correlation ($\rho \ge 0.3$), which is common in real-world data. For the smaller experiments ($M \le 4$), we see that \texttt{ctree} performs better than \texttt{VAEAC} in most settings, except for $\rho = 0.5$ in two of the experiments. We have no intuitive explanation for why \texttt{VAEAC} outperforms \texttt{ctree} for this particular correlation. For the larger experiments with moderate to strong dependence, we see that \texttt{VAEAC} more often outperforms \texttt{ctree} than for the smaller experiments, especially for $M=7$. However, for independent data, \texttt{VAEAC} is outperformed by \texttt{independence} and \texttt{ctree} in all experiments.

\begin{table}
\small
\centering
\begin{tabular}{@{}cccc@{}}
\toprule
$M$ & $L$ & $N_\text{test}$ & Categorical cut-off values \\
\midrule
 3 & 3 & 27   & $(-\infty, \phantom{-1.5,\,} \phantom{-1,\,} \phantom{-0.5,\,} 0, \phantom{0.5,\,} 1, \infty)$\\
 3 & 4 & 64   & $(-\infty, \phantom{-1.5,} \phantom{-1,}\! -0.5, 0, \phantom{0.5,\,}  1, \infty)$\\
 4 & 3 & 81   & $(-\infty, \phantom{-1.5,\,} \phantom{-1,\,} \phantom{-0.5,\,} 0, \phantom{0.5,\,}  1, \infty)$ \\
 5 & 6 & 2000 & $(-\infty, -1.5, -1,  \phantom{-0.5,\,} 0, 0.5, 1, \infty)$ \\
 7 & 4 & 2000 & $(-\infty, \phantom{-1.5,} \phantom{-1,} -0.5, \phantom{0,} 0.5, 1, \infty)$ \\
10 & 3 & 2000 & $(-\infty, \phantom{-1.5,} \phantom{-1,} -0.5,  \phantom{0,} \phantom{0.5,\,} 1, \infty)$ \\
\bottomrule
\end{tabular}
\caption{ {\small Outline of the setup of the categorical simulation studies in \Cref{subsection:simstudy:cat}. The fourth column describes the cut-off values between the different categories. \vspace{-2ex}}}
\label{tab:sim_cat:setup}
\end{table}

\begin{table}
\centering
\small
\begin{tabular}{@{}cccccccccc@{}}
\toprule
&&&&\multicolumn{6}{c}{$\overline{\text{EC1}}$ for each $\rho$} \\
\cmidrule{5-10}
$M$ & $L$ & $R$ & Method & $0.0$ & $0.1$ & $0.3$ & $0.5$ & $0.8$ & $0.9$ \\
\midrule
\multirow{3}{*}{3} & \multirow{3}{*}{3} & \multirow{3}{*}{20} & \texttt{Indep}. & \textbf{0.0170} & \textbf{0.0357} & 0.0946 & 0.1488 & 0.2584 & 0.3623 \\
& & & \texttt{Ctree} & 0.0229 & 0.0378 & \textbf{0.0428} & \textbf{0.0353} & \textbf{0.0298} & \textbf{0.0277}  \\
& & & \texttt{VAEAC} & 0.0395 & 0.0423 & 0.0483 & 0.0414 & 0.0425 & 0.0428 \\
\midrule
\multirow{3}{*}{3} & \multirow{3}{*}{4} & \multirow{3}{*}{20} & \texttt{Indep}.  & \textbf{0.0177} & \textbf{0.0390} & 0.1038 & 0.1482 & 0.2905 & 0.3733 \\
& & & \texttt{Ctree} & 0.0231 & 0.0393 & \textbf{0.0529} & 0.0512 & \textbf{0.0485} & \textbf{0.0401} \\
& & & \texttt{VAEAC} & 0.0396 & 0.0479 & 0.0533 & \textbf{0.0501} & 0.0508 & 0.0536\\
\midrule
\multirow{3}{*}{4} & \multirow{3}{*}{3} & \multirow{3}{*}{20} & \texttt{Indep}. & \textbf{0.0182} & \textbf{0.0412} & 0.1259 & 0.1822 & 0.2610 & 0.3590 \\ 
& & & \texttt{Ctree} & 0.0213 & 0.0432 & \textbf{0.0488} & 0.0480 & \textbf{0.0417} & \textbf{0.0334} \\
& & & \texttt{VAEAC} & 0.0387 & 0.0475 & 0.0514 & \textbf{0.0474} & 0.0446 & 0.0389 \\
\midrule
\multirow{3}{*}{5} & \multirow{3}{*}{6} & \multirow{3}{*}{10} & \texttt{Indep}. &  \B 0.0263 & 0.0519 & 0.1244 & 0.1908 & 0.3299 & 0.4176\\
& & & \texttt{Ctree} & 0.0266 & \B0.0404 & \B0.0636 & 0.0743 & \B 0.0841 & 0.0980\\
& & & \texttt{VAEAC} & 0.0620 & 0.0714 & 0.0725 & \B0.0720 & 0.0859 & \B 0.0748 \\
\midrule
\multirow{3}{*}{7} & \multirow{3}{*}{4} & \multirow{3}{*}{10} & \texttt{Indep}. & \B0.0189 & 0.0449 & 0.1210 & 0.2020 & 0.3410 & 0.4005 \\
& & & \texttt{Ctree} & 0.0206 & \B0.0446 & 0.0652 & 0.0706 & 0.0714 & 0.0736\\
& & & \texttt{VAEAC} & 0.0513 & 0.0584 & \B0.0631 & \B0.0655 & \B0.0642 & \B0.0647\\
\midrule
\multirow{3}{*}{10} & \multirow{3}{*}{3} & \multirow{3}{*}{10} & \texttt{Indep}. & \B0.0163 & 0.0551 & 0.1252 & 0.2260 & 0.4028 & 0.3995\\
& & & \texttt{Ctree} & 0.0169 & \B0.0439 & 0.0579 & 0.0618 & \B0.0613 & 0.0578 \\
& & & \texttt{VAEAC} & 0.0498 & 0.0490 & \B0.0514 & \B0.0566 & 0.0662 & \B0.0561\\
\bottomrule
\end{tabular}
\caption{{\small Table presenting the average EC1 for the different methods applied on the categorical data simulations with correlation $\rho$. The bolded numbers denote the smallest average MAE per experiment and $\rho$, that is, the best approach. \vspace{-2ex}}}
\label{tab:sim_cat:results}
\end{table}

\section{Data Generating Processes of Dependent Data}
\label{Appendix:GenerateData}
Here we present how the data are generated for the three types of simulation studies investigated in this article. In addition, we explain how we calculate the associated true Shapley values.

\subsection{Continuous Data}
\label{Appendix:sub:GenerateDataContinuous}
The continuous multivariate Burr distribution is chosen as its conditional distributions have known analytical expressions and are samplable. This allows us to compute the true contribution functions and Shapley values with arbitrary precision. The Burr distribution allows for heavy-tailed and skewed marginals and nonlinear dependencies, which can be found in real-world data sets. 

Other multivariate distributions with closed-form conditional distributions exist, for example, the multivariate Gaussian and the generalized hyperbolic (GH), which are fairly similar. Therefore, both give an unfair advantage to the \texttt{Gaussian} and \texttt{copula} approaches which assume Gaussian data. However, in a similar experiment to that of \Cref{subsection:simstudy:cont:lowdimensional}, not reported here, we generated data according to a GH distribution instead. The \texttt{VAEAC} approach was still able to perform on par with the \texttt{Gaussian} and \texttt{copula} methods, and much better than the \texttt{independence}, \texttt{empirical}, and \texttt{ctree} approaches.

The density of an $M$-dimensional Burr distribution is
\begin{align*}
    p_M(\boldsymbol{x}) = \frac{\Gamma(\kappa + M)}{\Gamma(\kappa)} \left(\prod_{m=1}^M b_mr_m\right) \frac{\prod_{m=1}^M x_m^{b_m-1}}{\left(1+\sum_{m=1}^M r_mx_m^{b_m}\right)^{\kappa+M}},
\end{align*}
for $x_m > 0$ \parencite{takahasi_note_1965}. The $M$-dimensional Burr distribution has $2M+1$ parameters, namely, $\kappa$, $b_1, \dots b_M$, and $r_1, \dots, r_M$. Furthermore, the Burr distribution is a compound Weibull distribution with the gamma distribution as compounder \parencite{takahasi_note_1965}, and it can also be seen as a special case of the Pareto IV distribution \parencite{Yari2006InformationAC}.

Any conditional distribution of the Burr distribution is in itself a Burr distribution \parencite{takahasi_note_1965}. Without loss of generality, assume that the first $S < M$ features are the unobserved features, then the conditional density $p(x_1,\dots,x_S | x_{S+1} = x^*_{S+1}, \dots, x_M = x_M^*)$, where $\boldsymbol{x}^*$ indicates the conditional values, is an $S$-dimensional Burr density. The associated parameters are then $\tilde{\kappa}, \tilde{b}_1, \dots, \tilde{b}_S$, and $\tilde{r}_1, \dots, \tilde{r}_S$, where $\tilde{\kappa} = \kappa + M - S$, while $\tilde{b}_j = b_j$ and $\tilde{r}_j = \frac{r_j}{1+\sum_{m=S+1}^M r_m(x_m^*)^{b_m}}$, for all $j=1,2,\dots,S$. These conditional distributions are then used to compute the true contribution functions $v_\texttt{true}(\s)$, for $\s \in \mathcal{C}$, and the Shapley values $\boldsymbol{\phi}_\texttt{true}$, as described in \Cref{subsec:simulations:Evaluation_method,subsection:simstudy:cont}.

An outline of one repetition of the continuous simulation study for a fixed $M$ is given below. All sampling is done with replacements and equal probability, and are identical for the different values of $N_{\text{train}}$ such that we can see the effect of the training size. This is repeated $R$ times and the average evaluation criteria are presented in \Cref{tab:SimStudyHighDimPart1,tab:SimStudyHighDimPart2}.
\begin{enumerate}
\item Generate training and test Burr data according to $\operatorname{Burr}(\kappa, \boldsymbol{r}, \boldsymbol{b})$. Where $r_j$ and $b_j$ are sampled from $\{1, 1.25, \dots, 5\}$ and $\{2, 2.25, \dots, 6\}$, respectively, for $j = 1,2,\dots, M$. Transform each feature by $u_j = F_j(x_j)$, where $F_j$ is the true parametric (cumulative) distribution function for the $j$th feature $x_j$.

\item Generate the response $y$ according to
\begin{align}
    \label{eq:response_high_dim_v2}
    y = \sum_{k=1}^{M/5}\left[\sin(\pi c_{k+1}u_{k+1}u_{k+2}) + c_{k+2}u_{k+3}\exp\{c_{k+3}u_{k+4}u_{k+5}\}\right],
\end{align}
where $c_k$ is sampled from $\{0.1, 0.2, \dots, 2\}$. Add noise to \eqref{eq:response_high_dim} by sampling $M/5$ features $u^{(1)},\dots, u^{(M/5)}$ and creating the noise according to $\frac{\epsilon}{M/5}\sum_{k=1}^{M/5} u^{(k)}$, where $\epsilon \sim \mathcal{N}(0,1)$.

\item Fit a random forest with $500$ trees using the \textsc{R}-package \texttt{ranger} \parencite{ranger} with default parameter settings to the training data.

\item Sample $N_\s=1000$ coalitions from $\mathcal{P}(\mathcal{M})$ to constitute $\mathcal{C}$. If $N_\s \geq 2^M$, then we set $\mathcal{C} = \mathcal{P}(\mathcal{M})$. Use the test data and $\mathcal{C}$ to compute the evaluation criteria for the different approaches as described in \Cref{subsec:simulations:Evaluation_method}. 
\end{enumerate}

\begin{figure}[t]
    \centering
    \includegraphics[width=1\textwidth]{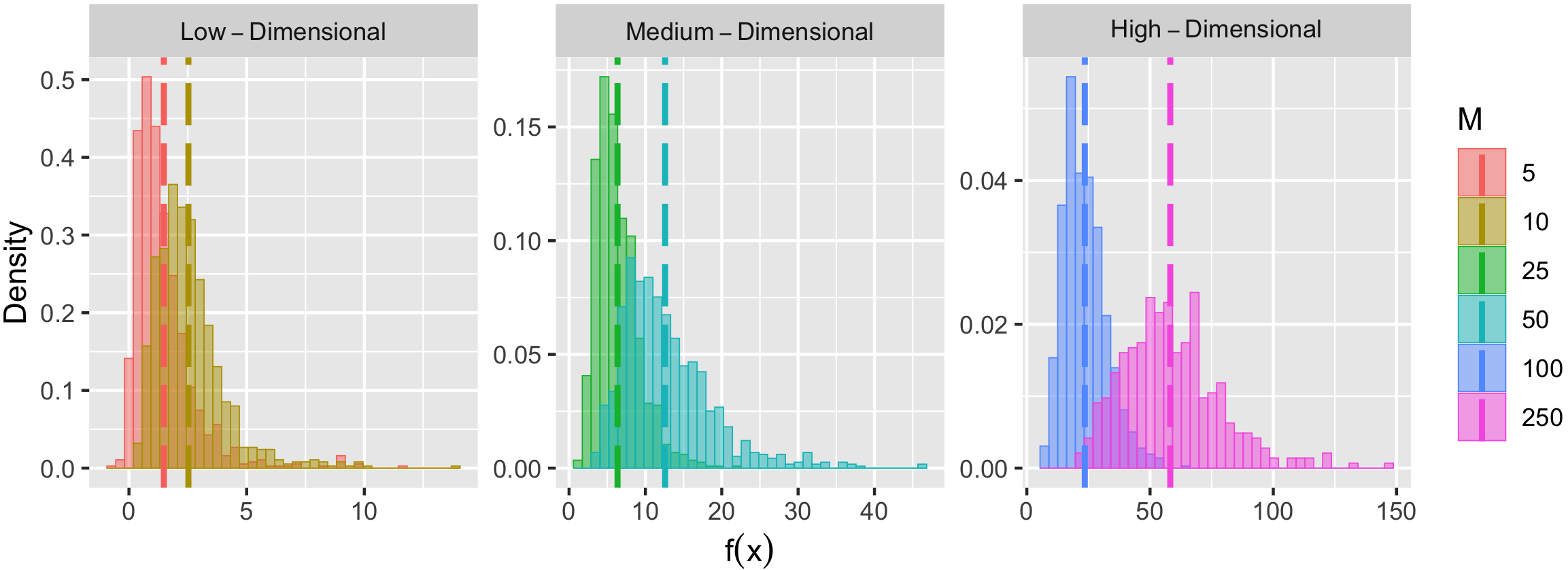}
    \caption{{\small Empirical distributions of the predicted responses and average $\phi_{0}$ over all repetitions.}}
    \label{fig:cont_pred_dist}
\end{figure}

In \Cref{tab:SimStudyHighDimPart1,tab:SimStudyHighDimPart2}, we see that the EC1 \eqref{eq:MAE} increases in line with $M$, which is partially linked to larger responses in \eqref{eq:response_high_dim_v2} for higher dimensions. The Shapley values for a test observation $\x$ are distributed such that $f(\x) = \phi_{0} + \phi_{\texttt{q}, 1} + \dots + \phi_{\texttt{q}, M}$, where $\phi_0$ is the expected prediction without any features, which we have set equal to $\bar{y}_\text{train}$ \parencite{aas2019explaining}. Intuitively, $\phi_j$, for $j=1,\dots,M$, are steps taken from $\phi_0 = \bar{y}_\text{train}$ to the predicted value $f(\x)$. The EC1 tells us the average absolute error we make in each step compared to the true Shapley value decomposition $f(\x) = \phi_{0} + \phi_{\texttt{true}, 1} + \dots + \phi_{\texttt{true}, M}$. Thus, the magnitude of EC1 depends on the values of $M$, $f(\x)$, and $\phi_{0}$. If the latter two are close, which they are in the low-dimensional settings, see \Cref{fig:cont_pred_dist}, we expect the EC1 to be reasonably low. In the high-dimensional settings, they are quite different and the $\phi_{\texttt{q},j}$ values have to cover a larger distance. Furthermore, the decomposition consist of more terms to precisely estimate. \Cref{fig:cont_pred_dist} displays the empirical distributions of the predicted test values $f(\x)$, across all repetitions, and the dashed lines represent the average $\phi_{0}$ for the different dimensions $M$.

\subsection{Mixed Data}
\label{Appendix:sub:GenerateDataMixed}
We now present a shortened description of the dependent mixed data generating process and related Shapley value computations of \textcite{redelmeier:2020}. For simplifying the true Shapley value computations, they use a linear predictive function of the form
\begin{align}
    \label{def:eq:sim_lin_pred_func}
    f(\boldsymbol{x}) = \alpha + \sum_{ j \in \mathcal{C}_{\text{cat}}} \sum_{l=1}^L \beta_{jl}\mathbf{1}(x_j = l) + \sum_{j \in \mathcal{C}_{\text{cont}}} \gamma_j x_j,
\end{align}
where $\mathcal{C}_{\text{cat}}$ and $\mathcal{C}_{\text{cont}}$ denote the set of categorical and continuous features, respectively, $L$ is the number of categories for each of the categorical features, and $\mathbf{1}(x_j = l)$ is the indicator function taking the value $1$ if $x_j=l$ and $0$ otherwise. Furthermore, $\alpha$, $\beta_{jl}$ for $j \in \mathcal{C}_{\text{cat}}$, $l=1,\dots,L$, and $\gamma_j$ for $j \in \mathcal{C}_{\text{cont}}$ are the parameters of the linear model. The dimension of the model is $M = |\mathcal{C}_{\text{cat}}| + |\mathcal{C}_{\text{cont}}|$, where $|\mathcal{C}_{\text{cat}}|$ and $|\mathcal{C}_{\text{cont}}|$ denote the number of categorical and continuous features, respectively. 

To generate $M$-dimensional dependent mixed data, we sample from an $M$-dimensional Gaussian distribution $\mathcal{N}_M(\boldsymbol{\mu}, \Sigma_\rho)$, but discretize $|\mathcal{C}_{\text{cat}}|$ of the features into $L$ categories each. Here $\boldsymbol{\mu}$ is the mean and $\Sigma_\rho$ is the covariance matrix, which reflects the feature dependence, and it is $1$ on the diagonal and $\rho$ off-diagonal. That is, we first sample a random variable
\begin{align*}
    (\tilde{x}_1, \dots, \tilde{x}_M) \sim  \mathcal{N}_M(\boldsymbol{\mu}, \Sigma_\rho),
\end{align*}
and then transform the features $\tilde{x}_j$, for $j\in\mathcal{C}_{\text{cat}}$, to categorical features $x_j$ using the following transformation: 
\begin{align*}
    x_j = l, \text{ if } v_j < \tilde{x}_j \le v_{l+1}, \text{ for } l = 1, \dots, L \text{ and } j\in\mathcal{C}_{\text{cat}},
\end{align*}
where $v_1,\dots, v_{L+1}$ is an increasing and ordered set of cut-off values defining the categories with $v_1 = -\infty$ and $v_{L+1} = +\infty$. We redo this $N_\text{train}$ times to create the training data set of $M$ dependent mixed features. The strength of the dependencies between the features is controlled by the correlation $\rho$ specified in $\Sigma_\rho$. Furthermore, the actual value of $x_j$ is irrelevant and the features are treated as non-ordered categorical features.

Computing the conditional expectation is troublesome for the mixed data setting, but the linear assumption of the predictive function in \eqref{def:eq:sim_lin_pred_func} simplifies the computations. As the predictive function is linear, the conditional expectation reduces to a linear combination of two types of univariate expectations. Let $\mathcal{S}_\text{cat}$ and $\thickbar{\mathcal{S}}_\text{cat}$ refer to the $\mathcal{S}$ and $\thickbar{\mathcal{S}}$ part of the categorical features $\mathcal{C}_\text{cat}$, respectively, with analogous sets $\mathcal{S}_\text{cont}$ and $\thickbar{\mathcal{S}}_\text{cont}$ for the continuous features. We can then write the desired conditional expectation as 
\begin{align*}
    \E \left[f(\boldsymbol{x}) | \boldsymbol{x}_\mathcal{S} = \boldsymbol{x}_\mathcal{S}^* \right] 
    &=
    \E \left[\alpha + \sum_{j \in \mathcal{C}_\text{cat}} \sum_{l=1}^L \beta_{jl}\mathbf{1}(x_j = l) + \sum_{j \in \mathcal{C}_{\text{cont}}} \gamma_j x_j  \,\Bigg|\, \boldsymbol{x}_\mathcal{S} = \boldsymbol{x}_\mathcal{S}^* \right] \\
    &= 
    \alpha + \sum_{j \in \mathcal{C}_\text{cat}} \sum_{l=1}^L \beta_{jl}\E \left[\mathbf{1}(x_j = l)| \boldsymbol{x}_\mathcal{S} = \boldsymbol{x}_\mathcal{S}^* \right] + \sum_{j \in \mathcal{C}_{\text{cont}}} \gamma_j \E \left[x_j| \boldsymbol{x}_\mathcal{S} = \boldsymbol{x}_\mathcal{S}^* \right] \\
    &=
    \alpha 
    +
    \sum_{j \in \thickbar{\mathcal{S}}_\text{cat}} \sum_{l=1}^L \beta_{jl}\E \left[\mathbf{1}(x_j = l)| \boldsymbol{x}_\mathcal{S} = \boldsymbol{x}_\mathcal{S}^* \right] 
    +
    \sum_{j \in \mathcal{S}_\text{cat}} \sum_{l=1}^L \beta_{jl} \mathbf{1}(x_j^* = l)
    \\
    &\phantom{= \alpha} +
    \sum_{j \in \thickbar{\mathcal{S}}_{\text{cont}}} \gamma_j \E \left[x_j| \boldsymbol{x}_\mathcal{S} = \boldsymbol{x}_\mathcal{S}^* \right]
    + 
    \sum_{j \in \mathcal{S}_{\text{cont}}} \gamma_jx_j^*.
\end{align*}

To calculate the two conditional expectations, we use results from \textcite{arellano2006unified} on Gaussian selection distributions in addition to basic probability theory and numerical integration. Specifically, the conditional expectation for the continuous features takes the form
\begin{align}
    \label{eq:expectation_cond_dist_1}
    \E \left[x_j| \boldsymbol{x}_\mathcal{S} = \boldsymbol{x}_\mathcal{S}^* \right]
    =
    \int_{-\infty}^\infty x p(x_j = x | \boldsymbol{x}_\mathcal{S} = \boldsymbol{x}_\mathcal{S}^*) \diff x
    =
    \int_{-\infty}^\infty x p(x)\frac{p(\boldsymbol{x}_\mathcal{S} = \boldsymbol{x}_\mathcal{S}^* | x_j = x)}{p(\boldsymbol{x}_\mathcal{S})}  \diff x,
\end{align}
where $p(x)$ denotes the density of the standard Gaussian distribution, $p(\boldsymbol{x}_\mathcal{S} = \boldsymbol{x}_\mathcal{S}^* | x_j = x)$ is the conditional distribution of $\boldsymbol{x}_\mathcal{S}$ given $x_j = x$, and $p(\boldsymbol{x}_\mathcal{S})$ is the marginal distribution of $\boldsymbol{x}_\mathcal{S}$. The latter two are both Gaussian and can be evaluated at the specific vector $\boldsymbol{x}_\mathcal{S}^*$ using the \textsc{R}-package \texttt{mvtnorm}, see \textcite{genz2009computation}. The integral is solved using numerical integration.

For the conditional expectation of the categorical features, recall that $x_j = l$ corresponds to the original Gaussian variable $\tilde{x}_j$ falling into the interval $(v_l, v_{l+1}]$. Thus, the conditional expectation takes the form
\begin{align*}
    \E \left[\mathbf{1}(x_j = l)| \boldsymbol{x}_\mathcal{S} = \boldsymbol{x}_\mathcal{S}^* \right] 
    =
    P(v_l < \tilde{x}_j \leq v_{l+1} | \boldsymbol{x}_\mathcal{S})
    =
    \int_{v_l}^{v_{l+1}}  p(x)\frac{p(\boldsymbol{x}_\mathcal{S} = \boldsymbol{x}_\mathcal{S}^* | x_j = x)}{p(\boldsymbol{x}_\mathcal{S})} \diff x,
\end{align*}
which can be evaluated similarly to \eqref{eq:expectation_cond_dist_1} and solved with numerical integration. Once we have computed the necessary conditional expectations for each of the $2^M$ feature subsets $\mathcal{S}$, we compute the Shapley values using \eqref{eq:ShapleyValuesDef}. This is done for both the mixed and categorical data simulation studies.

\subsection{Categorical Data}
\label{Appendix:sub:GenerateDataCategorical}
The dependent categorical data is generated similarly to the mixed data in the previous section, except that $|\mathcal{C}_{\text{cat}}| = M$. That is, we discretize all the $M$ features into $L$ categories. 

To calculate the true Shapley values $\phi_{j, \text{true}}(\boldsymbol{x}^{[i]})$, for $j = 1, \dots, M$ and $i = 1,\dots,N_{\text{test}}$, we need the true conditional expectation for all feature subsets $\mathcal{S}$. When all the features are categorical, the conditional expectation can be written as
\begin{align*}
    \E \left[f(\boldsymbol{x}) | \boldsymbol{x}_\mathcal{S} =\boldsymbol{x}_\mathcal{S}^* \right] 
    =
    \sum_{\boldsymbol{x}_{\thickbar{\mathcal{S}}} \in \mathcal{X}_{\thickbar{\mathcal{S}}}} f(\boldsymbol{x}_{\mathcal{S}}^*, \boldsymbol{x}_{\thickbar{\mathcal{S}}}) p(\boldsymbol{x}_{\thickbar{\mathcal{S}}} | \boldsymbol{x}_{\mathcal{S}} = \boldsymbol{x}_{\mathcal{S}}^*),
\end{align*}
where $\mathcal{X}_{\thickbar{\mathcal{S}}}$ denotes the feature space of the feature vector $\boldsymbol{x}_{\thickbar{\mathcal{S}}}$ which contains $L^{|\thickbar{\mathcal{S}}|}$ unique feature combinations.\footnote{\textcite{redelmeier:2020} incorrectly write $|\thickbar{\mathcal{S}}|^L$.} We need the conditional probability $p(\boldsymbol{x}_{\thickbar{\mathcal{S}}} | \boldsymbol{x}_{\mathcal{S}} = \boldsymbol{x}_{\mathcal{S}}^*)$ for each combination of $\boldsymbol{x}_{\thickbar{\mathcal{S}}} \in \mathcal{X}_{\thickbar{\mathcal{S}}}$. These can be written as $ p(\boldsymbol{x}_{\thickbar{\mathcal{S}}} | \boldsymbol{x}_{\mathcal{S}}) = p(\boldsymbol{x}_{\thickbar{\mathcal{S}}}, \boldsymbol{x}_{\mathcal{S}})\big/p(\boldsymbol{x}_{\mathcal{S}})$,
and then evaluated at the desired $\boldsymbol{x}_{\mathcal{S}}^*$.  Since all feature combinations correspond to hyper-rectangular subspaces of Gaussian features, we can compute all joint probabilities exactly using the cut-offs $v_1, \dots, v_{L+1}$:
\begin{align*}
    p(x_1=l_1, \dots, x_M = l_M) 
    =
    P(v_{l_1} < \tilde{x}_1 \le v_{l_1 + 1}, \dots, v_{l_M} < \tilde{x}_M \le v_{l_M + 1}),
\end{align*}
for $l_j = 1, \dots L$ and $j=1, \dots, M$. Here $p$ denotes the joint probability mass function of $\boldsymbol{x}$ while $P$ denotes the joint continuous distribution function of $\tilde{\boldsymbol{x}}$. The probability on the right-hand side is easily computed using the cumulative distribution function of the multivariate Gaussian distribution in the \textsc{R}-package \texttt{mvtnorm}. The marginal and joint probability functions based on only a subset of the features are computed analogously based on a subset of the full Gaussian distribution, which is also Gaussian.

\newpage

\section{Number of Monte Carlo Samples \texorpdfstring{$K$}{TEXT}}
\label{Appendix:sec:NumMonteCarlo}
\begin{figure}[!t]
    \centering
    \includegraphics[width=1\textwidth]{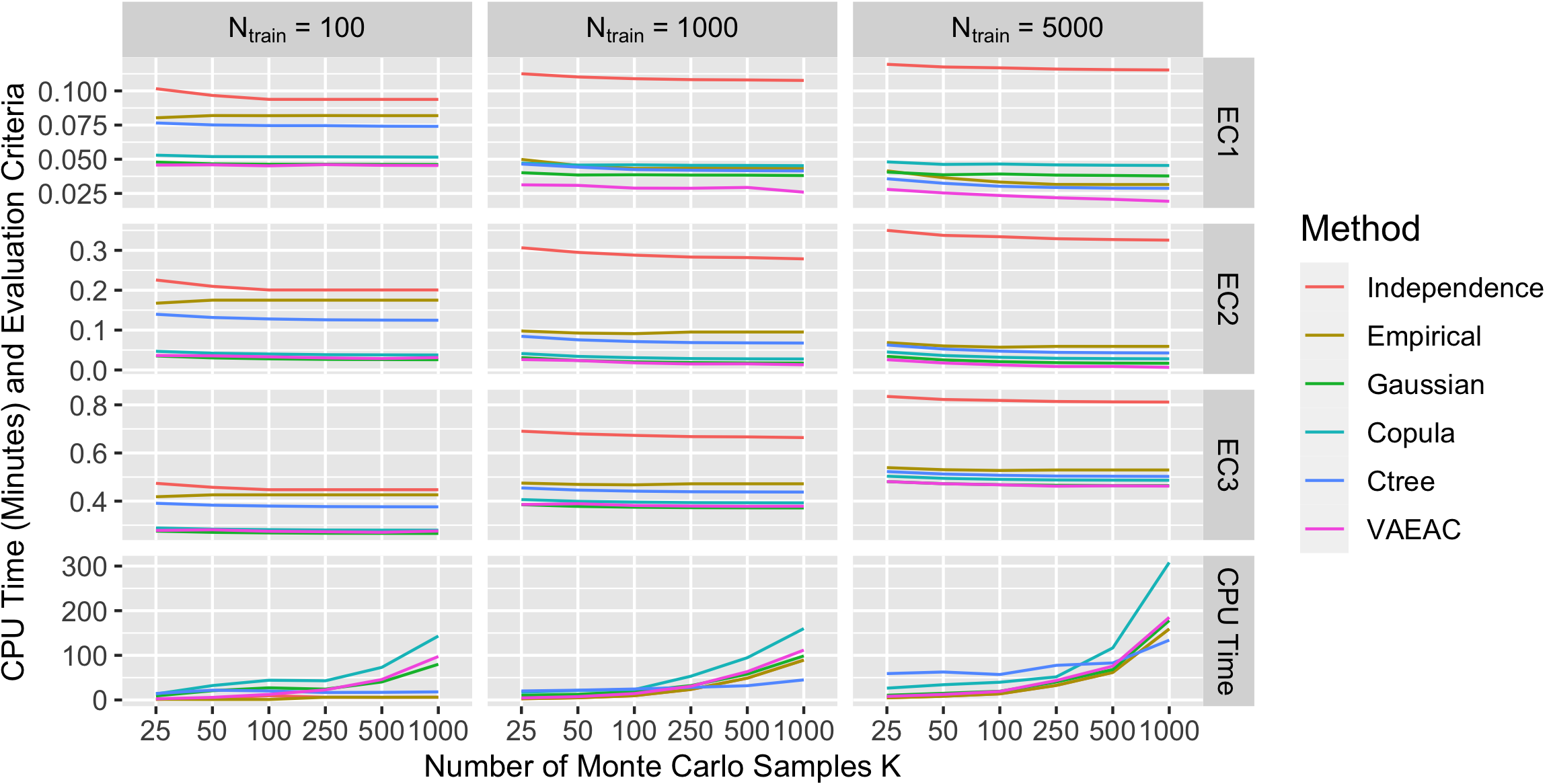}
    \caption{{\small Same set up as in the $M=10$-dimensional setting in \Cref{tab:SimStudyHighDimPart1}, but with different numbers of Monte Carlo samples $K$. The CPU times are given in minutes. \vspace{-2ex}}}
    \label{fig:differentMonteCarloSampleSizes}
\end{figure}

In \Cref{fig:differentMonteCarloSampleSizes}, we see how the number of Monte Carlo samples $K$ effect the evaluation criteria in the $M=10$-dimensional continuous simulation study, see \Cref{subsection:simstudy:cont:lowdimensional}. It is evident that $K > 250$ marginally decreases the evaluation criteria for this setting, but at the cost of higher CPU times. In practice, with limited time to generate the estimated Shapley values, we find $K=250$ to be sufficient. We have used $K=250$ in all experiments in this article.

\newpage
\section{The \texttt{VAEAC} Model as a Universal Approximator}
\label{Appendix:sec:UniversalApproximator}

\begin{figure}[!t]
    \centering
    \includegraphics[width=0.875\textwidth]{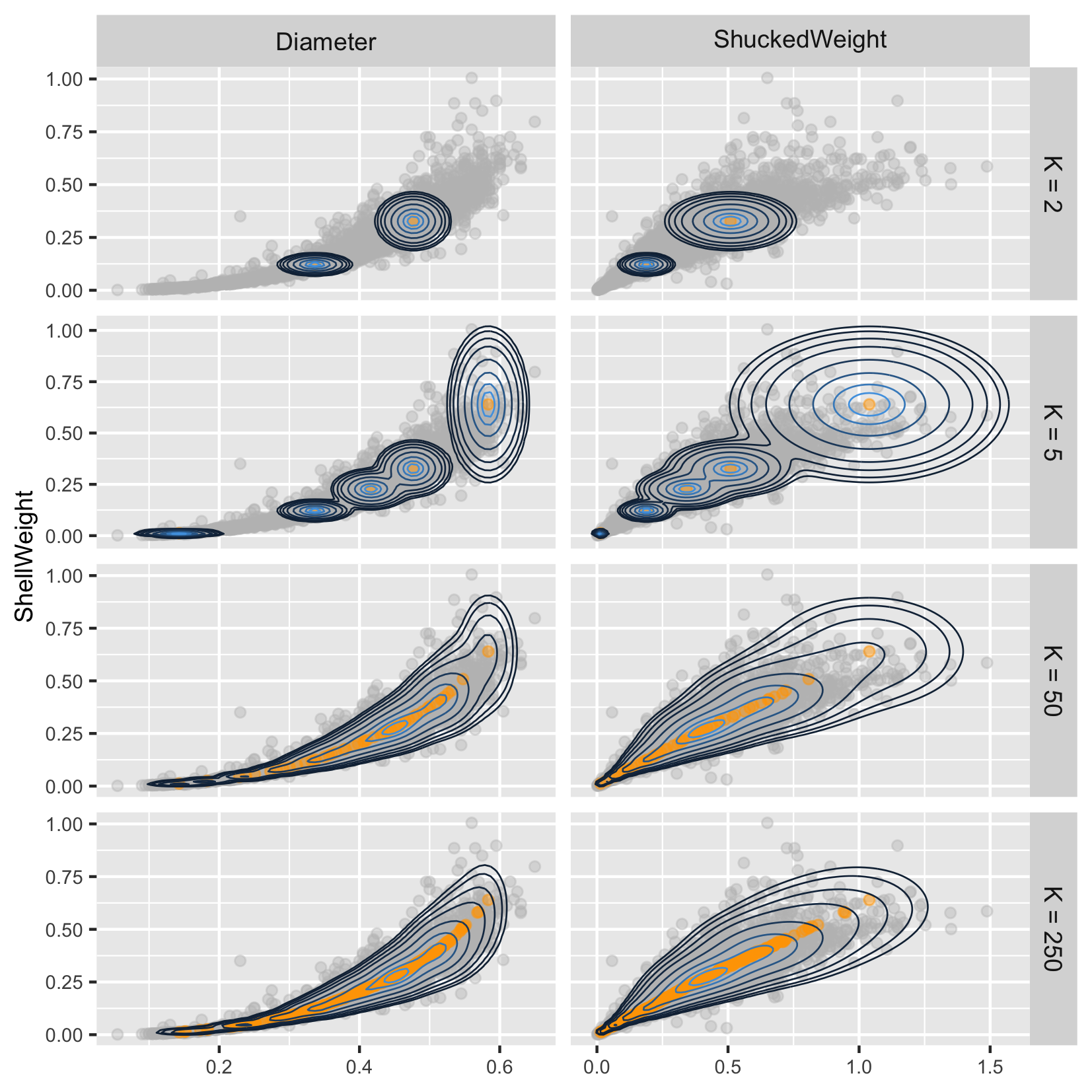}
    \caption{{\small Extension of \Cref{fig:RealData:CondDistSamples} and the Abalone data set from \Cref{sec:RealDataExample}. Illustration of the \texttt{VAEAC} model as a $K$-component Gaussian mixture model.}}
    \label{fig:InfiniteGaussianMixtureModel}
\end{figure}

Here we extend the setting of \Cref{fig:RealData:CondDistSamples}, and illustrate the \texttt{VAEAC} model as a $K$-component Gaussian mixture model, where $K$ is the number of Monte Carlo samples. \Cref{fig:InfiniteGaussianMixtureModel} illustrates, for different value of $K$, the $K$ different conditional distributions $p_\btheta(\xsb | \z^{(k)}, \xs, \sbb) = \mathcal{N}\big(\boldsymbol{x}_{\thickbar{\mathcal{S}}} | \boldsymbol{\mu}_{\boldsymbol{\theta}}(\z^{(k)}, \xs, \sbb), \operatorname{diag}[\boldsymbol{\sigma}_{\boldsymbol{\theta}}^2(\z^{(k)}, \xs, \sbb)] \big)$ inferred by the latent variables $\z^{(k)}$. The gray dots are the training data and the orange dots are the estimated means $\boldsymbol{\mu}_{\boldsymbol{\theta}}(\z^{(k)}, \xs, \sbb)$, while the variances  $\boldsymbol{\sigma}_{\boldsymbol{\theta}}^2(\z^{(k)}, \xs, \sbb)$ are reflected by the contour lines. It is evident that the means follow the center of the training data, while the variance follows the spread of the training data. To better visualize the distributions simultaneously in the same figure, each $p_\btheta(\xsb | \z^{(k)}, \xs, \sbb)$ is scaled to have mode equal to $1$. The contour lines are visually similar to those  obtained when using 2D kernel density estimation (not illustrated here).

\newpage
\printbibliography

\end{document}